\documentclass[10pt,journal,compsoc]{IEEEtran}

\usepackage{amsmath,amssymb,amsfonts}
\usepackage{algorithmic}
\usepackage{graphicx}
\usepackage{textcomp}
\usepackage{xcolor}
\usepackage{booktabs}
\usepackage{multirow}
\usepackage{url}
\usepackage{cite}
\DeclareMathSizes{5.8}{5.8}{5}{5}
\DeclareMathSizes{6}{6}{5}{5}
\DeclareMathSizes{6.2}{6.2}{5}{5}
\DeclareMathSizes{6.3}{6.3}{5}{5}
\DeclareMathSizes{6.4}{6.4}{5}{5}
\DeclareMathSizes{6.5}{6.5}{5}{5}
\DeclareMathSizes{6.6}{6.6}{5}{5}
\DeclareMathSizes{6.8}{6.8}{5}{5}
\DeclareMathSizes{7}{7}{5}{5}
\DeclareMathSizes{7.02625}{7}{5}{5}
\DeclareMathSizes{9.2}{9.2}{6.5}{5}
\DeclareMathSizes{9.25}{9.25}{6.5}{5}
\DeclareMathSizes{9.3}{9.3}{6.5}{5}
\DeclareMathSizes{9.35}{9.35}{6.5}{5}
\DeclareMathSizes{9.4}{9.4}{6.5}{5}
\renewcommand{\dbltopfraction}{0.96}
\renewcommand{\textfraction}{0.03}

\renewcommand{\arraystretch}{0.82}
\renewcommand{\baselinestretch}{0.778}
\begin{document}

\title{Sub-Voxel Displacement and Ipsilateral Territory Decoupling for 3D Bilateral Coronary Artery Tree Reconstruction in Cardiac CT Angiography}

\author{Zeyu~Jia and Xin~Ming%
\thanks{Zeyu Jia is with the School of Biomedical Engineering and Technology, Tianjin Medical University, Tianjin 300070, China, and also with the Medical School, Tianjin University, Tianjin 300072, China (e-mail: jiazeyu@tju.edu.cn). Xin Ming is with the School of Biomedical Engineering and Technology, Tianjin Medical University, Tianjin 300070, China. Corresponding author: Xin Ming.}}

\markboth{IEEE TRANSACTIONS ON MEDICAL IMAGING,~2026}%
{Jia and Ming: Sub-Voxel Displacement and Ipsilateral Territory Decoupling for 3D Coronary Arborescence}

\IEEEtitleabstractindextext{%
\begin{abstract}
Bilateral coronary tree (LCA/RCA) reconstruction from cardiac CT angiography (CCTA) is limited by $64^3$ discretization, cross-territory ambiguity, and ostium drift. Within ground-truth-foreground-conditioned $128^3$ coronary ROIs ($\pm\!15$-voxel margin, $\approx\!9.8\text{--}12.0\text{ mm}$; seed-free inference without ground-truth ostium/centerline seeds inside ROI; whole-volume localization upstream), we present a canonical $(Z_{\text{IS}},Y_{\text{PA}},X_{\text{LR}})$ multi-task network and scalar-potential graph arborescence framework. Volumes, centerlines, and thick-stem ostia map into $(Z_{\text{IS}},Y_{\text{PA}},X_{\text{LR}})$ space, extending a 3D pyramid encoder and tubular decoder with a $128^3\!\to\!64^3$ boundary skip, stop-gradient $128^3$ residual branch, 3D sub-voxel displacement head $\Delta\hat{\mathbf{u}}(\mathbf{x})$, dual-channel ipsilateral head $(\mathbf{P}_{\text{LCA}},\mathbf{P}_{\text{RCA}})$, and $32^3$ ostium detector. Warm-starting from a recall-balanced anchor, $327,385$ active parameters ($398,843$ total) are fine-tuned in one 30-epoch AdamW loop ($\epsilon\!=\!10^{-8}$, 7 groups) under decoupled L2-SP and anchor blending. Our solver couples cylindrical Valsalva-STJ calibration, $128^3$ orifice rooting, contralateral inhibition, heuristic radius-ratio pruning, normal-plane pullback, and coverage-guarded collision reduction. On locked-weight out-of-sample \textbf{OOS-50} (Cases 251--300, $50$ unseen volumes, $100$ trees; Cases 201--250 isolated) versus standalone 3D U-Net (skeletonization/$M_3$--$M_5$ solver) and five proxies, our method achieves $2.7671\!\pm\!1.1385\text{ mm}$ $\text{CD}_{\text{CL}}^{\text{union}}$ ($3\text{-seed SD}\!=\!0.0415\text{ mm}$), $3.2338\!\pm\!1.3557\text{ mm}$ $\text{CD}_{\text{CL}}^{\text{arc1mm}}$, and $3.2926\!\pm\!2.6253\text{ mm}$ $\overline{\text{CD}}_{\text{ipsi}}$ ($\overline{\text{CD}}_{\text{ipsi}}^{\text{arc1mm}}\!=\!3.7460\text{ mm}$, node $p_{\text{Holm}}\!\le\!0.0400$ vs.\ Ext-0b [$\text{arc1mm } p\!=\!0.1629/0.2187$]; $76.33\%$ junction [$23.04\%$ 1-to-1] and $79.65\%$ branch F1). On held-out \textbf{Dev-14} ($14$ volumes, $28$ trees), it achieves $2.1362\!\pm\!0.5041\text{ mm}$ $\text{CD}_{\text{CL}}^{\text{union}}$ ($3\text{-seed SD}\!=\!0.0369\text{ mm}$), $2.3798\!\pm\!0.5725\text{ mm}$ $\text{CD}_{\text{CL}}^{\text{arc1mm}}$, and $2.2480\!\pm\!0.6835\text{ mm}$ $\overline{\text{CD}}_{\text{ipsi}}$ ($\overline{\text{CD}}_{\text{ipsi}}^{\text{arc1mm}}\!=\!2.4951\text{ mm}$, node $\text{CD}_{\text{CL}}^{\text{union}}$ $p_{\text{Holm}}\!=\!0.0494$ vs.\ Ext-0b [$\overline{\text{CD}}_{\text{ipsi}}/\text{arc1mm } p_{\text{Holm}}\!=\!0.1353/0.0676\text{--}0.0906$]; $56.76\%/77.74\%$ $\text{F1}@1/2\text{mm}$; $79.95\%$ junction [$21.06\%$ 1-to-1] and $83.80\%$ branch F1, Med/Mean $4.6880/4.9298\text{ mm}$ ostia), with pooled $N\!=\!64$ $\text{CD}_{\text{CL}}^{\text{union}}\!=\!2.6291\!\pm\!1.0629\text{ mm}$ ($\text{CD}_{\text{CL}}^{\text{arc1mm}}\!=\!3.0470\text{ mm}$, $\overline{\text{CD}}_{\text{ipsi}}\!=\!3.0641\text{ mm}$).
\end{abstract}

}

\maketitle
\IEEEdisplaynontitleabstractindextext
\IEEEpeerreviewmaketitle
\makeatletter
\renewcommand{\normalsize}{\@setfontsize\normalsize{9.35pt}{10.5pt}}
\makeatother
\normalsize
\let\oldthebibliography\thebibliography
\renewcommand{\thebibliography}[1]{\oldthebibliography{#1}\fontsize{5.8}{6.1}\selectfont\def\IEEEbibitemsep{0pt}\setlength{\itemsep}{0pt}\setlength{\parskip}{0pt}}
\setcounter{dbltopnumber}{2}
\setcounter{topnumber}{2}
\setcounter{totalnumber}{4}
\renewcommand{\dbltopfraction}{0.96}
\renewcommand{\textfraction}{0.03}

\vspace{-5pt}
\section{Introduction}
\vspace{-2pt}
\IEEEPARstart{A}{ccurate} three-dimensional (3D) reconstruction of bilateral coronary arterial trees---the left coronary artery (LCA, branching into the left anterior descending and circumflex territories) and the right coronary artery (RCA)---from cardiac computed tomography angiography (CCTA) is a prerequisite for quantitative anatomical assessment and hemodynamic simulation~\cite{taylor2013computational,min2012diagnostic,norgaard2014diagnostic,zeng2023imagecas}. As shown in topology-focused benchmarks~\cite{shit2021cldice,owusu2026same,qiu2025topology}, high voxel-wise overlap (Dice) does not guarantee tree continuity or prevent cross-territory structural errors: dense 3D segmentation networks may sever thin distal vessels ($r < 1.0\text{ mm}$) under partial-volume attenuation or fuse adjacent LCA and RCA branches in the posterior atrioventricular groove, inflating per-tree ipsilateral distance even when pooled union error appears low.

Within coarse-to-fine 3D vascular reconstruction pipelines~\cite{wolterink2019coronary,naeem2025reftr,zhao2025deformcl}, analysis on ImageCAS~\cite{zeng2023imagecas} highlights three bottlenecks on $64^3$ feature lattices:
\begin{enumerate}
    \item \textbf{Coarse-Grid ($64^3$) Spatial Quantization Floor}: Decoding a $128^3$ CCTA crop spanning $160\text{--}200\text{ mm}$ onto a $64^3$ lattice ($\sim 2.5\text{--}3.1\text{ mm}$ spacing) incurs an empirical $0.8\text{--}1.2\text{ mm}$ lattice discretization offset, limiting unrefined sub-millimeter precision ($\text{F1}@1\text{mm}$) to $11\text{--}12\%$.
    \item \textbf{Single-Channel Territory Ambiguity, Root-Interface Separability, and Canonical Axis Alignment}: Once predicted lumen masks merge across the posterior atrioventricular groove, the geometric watershed enters the vascular lumen so local voxel classifiers cannot uniquely recover LCA/RCA territory labels without anchoring each tree at its disjoint aortic sinus-wall root patch and decoupling ipsilateral territory channels. Raw NIfTI arrays in Fortran order $(X_{\text{RL}}, Y_{\text{PA}}, Z_{\text{IS}})$ also require canonical world-axis alignment ($\det = +1$) into $(Z_{\text{IS}}, Y_{\text{PA}}, X_{\text{LR}})$ space while preserving connected LCA/RCA partitions and thick-stem aortic-wall ostia.
    \item \textbf{Recall-Preserving Joint Optimization vs.\ Representation Drift}: Unregularized fine-tuning can erode vessel recall or inflate background drift. Jointly fine-tuning all active parameters from a recall-balanced canonical anchor under decoupled L2-SP anchoring preserves tubular signal-to-noise ratio (validation $\text{SNR} = 8.34\times$, median centerline probability $0.5810$ vs.\ far-field 95th-percentile background $0.0696$). Furthermore, decoupling proximal aortic-wall ostium rooting (Stage $M_3$) from downstream vascular graph extraction (Stages $M_4$--$M_5$) ensures that occasional root-localization outliers ($>10\text{ mm}$) remain confined to the proximal stem rather than propagating topology failures into the distal arborescence.
\end{enumerate}

To address these challenges within standardized ground-truth-foreground-conditioned $128^3$ coronary regions of interest (cropped around the binary coronary mask with a $\pm 15$ raw-voxel margin, $\approx 9.8\text{--}12.0\text{ mm}$, where seed-free inference denotes zero ground-truth ostium or centerline seeds inside the $128^3$ ROI and whole-volume heart/aorta localization prior to ROI cropping is upstream preprocessing outside the evaluated scope), we present a 3D bilateral coronary arborescence framework under canonical axis alignment ($(Z_{\text{IS}}, Y_{\text{PA}}, X_{\text{LR}})$) (Fig.~\ref{fig:architecture_m1_m4}) with four contributions:
\begin{enumerate}
    \item \textbf{$128^3 \to 64^3$ Boundary Skip, Native $128^3$ Residual Tubular Decoder, and Continuous 3D Sub-Voxel Displacement Field}: We extend the 3D tubular decoder with a stride-2 boundary skip ($\mathbf{H}_{128}$), a native $128^3$ residual branch predicting $\mathbf{P}_{\text{lumen}, 128}(\mathbf{x}) \in [0, 1]^{128^3}$ under stop-gradient backbone isolation, and a 3D sub-voxel displacement head (zero-initialized at construction) predicting $\Delta\hat{\mathbf{u}}(\mathbf{x}) \in [-2.0, +2.0]^3\text{ mm}$.
    \item \textbf{Dual-Channel Coordinate-Conditioned Ipsilateral Decoupling}: A residual dual-channel head combines multi-scale 3D features with canonical coordinates $\mathbf{G}_{64} \in [0, 1]^3$ to predict $\hat{A}_{\text{ipsi}}(\mathbf{x}; \mathbf{G}_{64}) = (\mathbf{P}_{\text{LCA}}(\mathbf{x}),\allowbreak \mathbf{P}_{\text{RCA}}(\mathbf{x}))$ under contralateral hard-negative suppression.
    \item \textbf{Two-Stage Recall-Preserving Multi-Task Optimization}: We standardize volumes, reference centerlines, and thick-stem ostia into $(Z_{\text{IS}}, Y_{\text{PA}}, X_{\text{LR}})$ space (axis permutation $(X, Y, Z) \mapsto (Z, Y, X)$ and sagittal reflection, $\det = +1$), warm-start from our recall-balanced reference anchor checkpoint, and fine-tune all $327,385$ active parameters ($81,008$ in the 3D pyramid encoder, $121,183$ in the tubular decoder, and $125,194$ in the 3D ostium detector, out of $398,843$ total parameters; Supplementary Table~S1) in one 30-epoch AdamW loop ($\epsilon = 10^{-8}$, 7 parameter groups) with decoupled L2-SP and per-step anchor blending/shrinkage.
    \item \textbf{Scalar-Potential-Weighted Graph Arborescence with Orifice Contact Rooting, Normal-Plane Pullback, and Ipsilateral-Field-Coupled Collision Reduction}: Within our scalar-potential-weighted graph arborescence solver, we couple cylindrical Valsalva-STJ wall calibration, Stage-$M_3$ native-$128^3$ Valsalva-orifice contact expectation, Stage-$M_4$ native-$128^3$ residual coupling, territory-specific graph tree extraction, heuristic radius-ratio trajectory pruning, normal-plane pullback $(\mathbf{I} - \mathbf{T}\mathbf{T}^\top)\allowbreak\Delta\hat{\mathbf{u}}(\mathbf{x})$, and Stage-$M_5$ coverage-guarded bilateral collision reduction and subtree handover,\footnote{Single-tree connectivity ($\beta_0 = 1$) and acyclicity ($\beta_1 = 0$) from each rooted ostium $\mathbf{o}_c^*$ hold by shortest-path and minimum-spanning-tree graph construction across all $128/128$ evaluated trees (with $0/128$ degenerate fallback stems triggered), rather than representing an unconstrained empirical segmentation outcome.} benchmarking against an independently trained standalone single-task 3D U-Net paired with morphological skeletonization (Ext-0) or our $M_3$--$M_5$ solver (Ext-0b), plus five controlled same-backbone proxies (Ext-1..5).
\end{enumerate}

\vspace{-5pt}
\section{Related Work}
\vspace{-2pt}
\subsection{3D Coronary Artery Segmentation and Centerline Extraction}
\vspace{-2pt}
Deep learning methods for 3D coronary tree extraction follow dense voxel segmentation followed by skeletonization~\cite{ronneberger2015u,isensee2021nnu,shit2021cldice,zeng2023imagecas,gharleghi2022asoca} or explicit centerline graph/trajectory modeling~\cite{wolterink2019coronary,qiu2025topology,zhao2025deformcl,naeem2025reftr,prabhakar2026vesseltok,sinha2024trind}. While clDice~\cite{shit2021cldice} and dynamic snake convolutions~\cite{qi2023dynamic,chen2025dynamic} improve tubular continuity, morphological thinning remains susceptible to fragmentation ($\beta_0 > 1$) and loops ($\beta_1 > 0$) under calcified plaques or low-contrast segments. Hybrid frameworks such as CorSegRec~\cite{qiu2025topology} reconnect broken skeletons via geodesic walks, whereas CNN orientation tracking~\cite{wolterink2019coronary}, RefTr~\cite{naeem2025reftr}, VesselTok~\cite{prabhakar2026vesseltok}, and DeformCL~\cite{zhao2025deformcl} track or deform 3D centerline point clouds. However, local trackers can drift into adjacent veins, and direct 3D offset deformation without normal-plane projection or bilateral territory decoupling remains vulnerable to vertex bunching and cross-territory hijacking.

\vspace{-5pt}
\subsection{Sub-Voxel Distance Regression and Anatomical Priors}
\vspace{-2pt}
Regressing continuous displacement vectors toward anatomical centerlines mitigates voxel discretization in curvilinear structure detection~\cite{wolterink2019coronary}. Physiological scaling principles---such as Murray's law ($r_{\text{parent}}^3 = \sum_k r_{\text{child}, k}^3$)~\cite{murray1926physiological} and Huo--Kassab scaling~\cite{huo2012intraspecific,finet2004fractal}---motivate parent-to-child caliber consistency checks for distinguishing true bifurcations from spurious twigs. Our work combines 3D sub-voxel displacement regression with dual-channel LCA/RCA decoupling, heuristic proximal-to-distal radius tapering and radius-ratio pruning, and coverage-guarded bilateral collision reduction.

\vspace{-6pt}
\section{Methodology}
\label{sec:methods}
\vspace{-2pt}

\begin{figure*}[!t]
\centering
\includegraphics[width=0.80\textwidth]{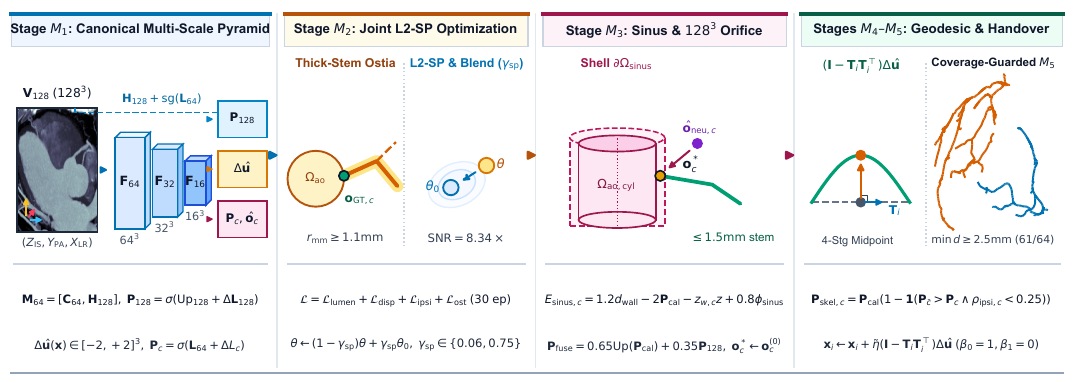}
\vspace{-9pt}
\caption{\fontsize{7.2}{7.8}\selectfont Architectural overview of the canonical axis-aligned ($(Z_{\text{IS}}, Y_{\text{PA}}, X_{\text{LR}})$) 3D bilateral coronary tree reconstruction pipeline across Stages $M_1 \to M_5$: ($M_1$) 3D pyramid encoder, $128^3 \to 64^3$ boundary skip, stop-gradient $128^3$ residual branch ($\mathbf{P}_{\text{lumen}, 128}$), sub-voxel displacement head ($\Delta\hat{\mathbf{u}}(\mathbf{x}) \in [-2, +2]^3\text{ mm}$), dual-channel ipsilateral head $(\mathbf{P}_{\text{LCA}}, \mathbf{P}_{\text{RCA}})$, and $32^3$ dual-sinus ostium detector $(\hat{\mathbf{o}}_{\text{reg}, c}, \hat{\mathbf{o}}_{\text{det}, c})$; ($M_2$) 30-epoch multi-task optimization ($327,385$ active parameters) with decoupled L2-SP, per-step anchor blending, and thick-stem ostium supervision; ($M_3$) cylindrical Valsalva-STJ calibration, sinus-wall rooting, $128^3$ orifice contact expectation, and ostium pullback; and ($M_4$--$M_5$) $128^3$-coupled scalar-potential graph arborescence, contralateral inhibition, heuristic radius-ratio centripetal pruning, 4-stage normal-plane pullback, and coverage-guarded bilateral collision reduction.}
\label{fig:architecture_m1_m4}
\end{figure*}

As shown in Fig.~\ref{fig:architecture_m1_m4}, our pipeline maps an input $128^3$ CCTA volume $\mathbf{V}_{128} \in \mathbb{R}^{1 \times 128 \times 128 \times 128}$ in canonical $(Z_{\text{IS}}, Y_{\text{PA}}, X_{\text{LR}})$ orientation with voxel spacing $\mathbf{s}_{\text{mm}} = (s_z, s_y, s_x) \in \mathbb{R}^3$ to connected, acyclic 3D trees $\mathcal{T}_{\text{LCA}} = (\mathcal{V}_{\text{LCA}}, \mathcal{E}_{\text{LCA}})$ and $\mathcal{T}_{\text{RCA}} = (\mathcal{V}_{\text{RCA}}, \mathcal{E}_{\text{RCA}})$ via Stages $M_1 \to M_5$ (complete algorithmic specifications in Supplementary Algorithms~A1--A5 and parameter/threshold ledgers in Supplementary Tables~S1--S2).

\vspace{-5pt}
\subsection{Stage $M_1$: Multi-Scale 3D Pyramid Encoder and Multi-Branch Tubular Field Network}
\vspace{-2pt}
Given $\mathbf{V}_{128}$, the multi-scale 3D pyramid encoder ($81,008$ parameters; Supplementary Algorithm~A1 and Table~S1) extracts features $\mathbf{F}_{64} \in \mathbb{R}^{16 \times 64^3}$, $\mathbf{F}_{32} \in \mathbb{R}^{32 \times 32^3}$, $\mathbf{F}_{16} \in \mathbb{R}^{64 \times 16^3}$, and global context $\mathbf{g}_{\text{ctx}} \in \mathbb{R}^{64}$. In the 3D tubular decoder ($D_{\text{hid}} = 32$, $D_{\text{mid}} = 16$), $\mathbf{F}_{32}$ is upsampled by a $32^3 \to 64^3$ transposed-convolution block ($8,288$ parameters) to $\mathbf{U}_{32}$, concatenated with $\mathbf{F}_{64}$ as $\mathbf{C}_{64} = [\mathbf{U}_{32}, \mathbf{F}_{64}] \in \mathbb{R}^{48 \times 64^3}$, and decoded by the $64^3$ fusion cascade ($48 \to 32 \to 16 \to 1$, $55,457$ parameters; $63,745$ base decoder parameters) into base logits $\mathbf{L}_{\text{lumen}}$ and $\mathbf{P}_{\text{lumen}}(\mathbf{x}) = \sigma(\mathbf{L}_{\text{lumen}}(\mathbf{x}))$. To overcome $64^3$ discretization and LCA/RCA ambiguity while isolating $128^3$ residual updates from destabilizing the $64^3/32^3$ backbone via stop-gradient detachment ($\text{sg}(\cdot)$), we introduce four branches:
\begin{enumerate}
    \item \textbf{$128^3 \to 64^3$ Boundary Skip and Native $128^3$ Residual Branch}: A stride-2 block ($1 \to 8$, $240$ parameters) extracts $\mathbf{H}_{128} \in \mathbb{R}^{8 \times 64^3}$ from $\mathbf{V}_{128}$ to form $\mathbf{M}_{64} = [\mathbf{C}_{64}, \mathbf{H}_{128}] \in \mathbb{R}^{56 \times 64^3}$. In parallel, a native $128^3$ stem ($240$ parameters) fused with upsampled detached $64^3$ features ($\text{sg}(\mathbf{M}_{64})$, $3,608$ parameters) feeds the $128^3$ residual decoder ($16 \to 8 \to 1$, $3,489$ parameters; $7,337$ total $128^3$ parameters) to predict $\Delta\mathbf{L}_{128} = 1.5\tanh(g_{128}([\mathbf{U}_{128}, \mathbf{S}_{128}]))$, $\mathbf{L}_{\text{lumen}, 128} = \text{Upsample}_{128}(\text{sg}(\mathbf{L}_{\text{lumen}})) + \Delta\mathbf{L}_{128}$, and $\mathbf{P}_{\text{lumen}, 128}(\mathbf{x}) = \sigma(\mathbf{L}_{\text{lumen}, 128}(\mathbf{x}))$.
    \item \textbf{Continuous 3D Sub-Voxel Displacement Field}: A two-layer 3D head ($56 \to 16 \to 3$, $24,291$ parameters, with its final $1^3$ layer zero-initialized at architecture construction prior to Stage-0 pre-training) maps $\mathbf{M}_{64}$ to a bounded 3D vector field ($\delta_{\max} = 2.0\text{ mm}$): $\Delta\hat{\mathbf{u}}(\mathbf{x}) = \delta_{\max} \tanh(g_{\text{disp}}(\mathbf{M}_{64})(\mathbf{x})) \in [-\delta_{\max}, +\delta_{\max}]^3$.
    \item \textbf{Dual-Channel Ipsilateral Decoupling}: Concatenating $\mathbf{M}_{64}$ with coordinates $\mathbf{G}_{64}(\mathbf{x}) = (z, y, x) \in [0, 1]^3$ ($59$ channels), a 3D head ($59 \to 16 \to 2$, $25,570$ parameters, final $1^3$ layer zero-initialized at architecture construction prior to Stage-0 pre-training) predicts residual logits $\Delta\mathbf{L}_{\text{ipsi}}(\mathbf{x}) = (\Delta L_{\text{LCA}}(\mathbf{x}), \Delta L_{\text{RCA}}(\mathbf{x}))$ and $\mathbf{P}_c(\mathbf{x}) = \sigma(\mathbf{L}_{\text{lumen}}(\mathbf{x}) + \Delta L_c(\mathbf{x}))$, $c \in \{\text{LCA}, \text{RCA}\}$.
    \item \textbf{Coarse $32^3$ Coordinate-Masked Dual-Sinus 3D Ostium Detector}: On the $32^3$ crop, a 3D ostium detector ($125,194$ parameters) combines coordinate-masked features, keypoint attention, and DSNT expectation with $\mathbf{g}_{\text{ctx}}$ to predict dual-sinus ostium coordinates $(\hat{\mathbf{o}}_{\text{reg}, c}, \hat{\mathbf{o}}_{\text{det}, c}) \in [0, 1]^3$ and takeoff priors for $c \in \{\text{LCA}, \text{RCA}\}$.
\end{enumerate}

\vspace{-5pt}
\subsection{Stage $M_2$: Reference Target Provenance and Recall-Preserving Multi-Task Optimization}
\vspace{-2pt}
Each raw ImageCAS CCTA volume is cropped to a 3D bounding box around the binary foreground coronary mask padded by $\pm 15$ raw voxels per axis ($\approx 9.8\text{--}12.0\text{ mm}$ margin, resampled to $128^3$ in canonical $(Z_{\text{IS}}, Y_{\text{PA}}, X_{\text{LR}})$ space, where seed-free inference denotes zero ground-truth ostium or centerline seeds inside the $128^3$ ROI and whole-volume localization is upstream preprocessing). Because ImageCAS provides binary voxel lumen masks rather than manually adjudicated centerline graphs or ostia, reference centerlines are extracted deterministically via 3D morphological skeletonization (3D Lee medial-axis thinning), per-node lumen radii $r_{\text{mm}}$ via 3D Euclidean distance transform, connected LCA/RCA subtrees ($\mathcal{G}_{\text{GT}, \text{LCA}}, \mathcal{G}_{\text{GT}, \text{RCA}}$) via connected-component partitioning, and algorithmic thick-stem aortic-wall ostia (proximal trunk nodes with $r_{\text{mm}} \ge \max(1.10\text{ mm}, P_{75}(r_{\text{mm}}))$ and $Z_{\text{IS}} \ge 0.45$, within the top $0.035$ $Z_{\text{IS}}$ band for LCA and $Z_{\text{IS}} \ge 0.52$ for RCA and within $3.5 / 3.0\text{ mm}$ of minimal sinus-wall distance, median $d_{\text{wall}} = 0.7443\text{ mm}$). On the $N_{\text{train}} = 35$ training volumes (\textbf{Train-35}: Cases $1, 10\text{--}14, 100\text{--}128$), we precompute derived continuous supervision targets from these reference structures: $64^3$ distance fields $d_{\text{LCA}}(\mathbf{x}), d_{\text{RCA}}(\mathbf{x})$, Gaussian tubular targets $Y_c(\mathbf{x}) = \exp(-d_c(\mathbf{x})^2 / (2\sigma_L^2))$ ($\sigma_L = 2.0\text{ mm}$), territory masks $m_{\text{pos}, c}, m_{\text{contra}, c}, m_{\text{bg}}$, nearest-centerline 3D offset targets $\Delta\mathbf{u}^*(\mathbf{x}) = \text{clip}(\mathbf{p}_{\text{GT}}(\mathbf{x}) - \mathbf{x}, -2.0, +2.0)$, cylindrical aortic-wall distance $\phi_{\text{ao}}(\mathbf{x}) = d_{\text{mm}}(\mathbf{x}, \partial\Omega_{\text{cyl}})$, and $128^3$ tubular targets $Y_{128}(\mathbf{x}) = \exp(-d_{\text{bi}, 128}(\mathbf{x})^2 / (2\sigma_{128}^2))\mathbb{I}(d_{\text{bi}, 128}(\mathbf{x}) \le 2.5\text{ mm})$ ($\sigma_{128} = 1.65\text{ mm}$). Warm-starting from the pre-trained Stage-0 recall-balanced anchor $\theta_0$ (which already holds non-zero pre-trained $g_{\text{disp}}$ [$48/48$] and $g_{\text{ipsi}}$ [$32/32$] output weights loaded without re-zeroing after checkpoint loading, while $g_{128}$ starts from its constructor zero-init), all $327,385$ active parameters ($398,843$ total, including $71,458$ frozen historical step-flow parameters; Supplementary Algorithm~A2 and Table~S1) are fine-tuned in one 30-epoch loop ($\mathcal{L}_{\text{total}} = \mathcal{L}_{\text{lumen}} + \mathcal{L}_{\text{reg-gt}} + \mathcal{L}_{\text{takeoff}}$) under AdamW ($\epsilon = 10^{-8}$, 7 parameter groups with cosine annealing $T_{\max}=30$: \textit{enc-proj} $5\text{e-}4\!\to\!1\text{e-}5$, \textit{enc-stem} $2\text{e-}5\!\to\!1\text{e-}5$, \textit{lumen-post-gn} $8\text{e-}4\!\to\!6\text{e-}4$, \textit{lumen-pre-gn} $2.5\text{e-}5\!\to\!1\text{e-}5$, \textit{lumen-multitask} $2\text{e-}5\!\to\!1\text{e-}5$, \textit{lumen-hr128} $2.5\text{e-}3\!\to\!2\text{e-}5$, and \textit{ostium} $4\text{e-}5\!\to\!1\text{e-}5$; seeds $\{20261012, 20261013, 20261014\}$ evaluating conditional 30-epoch fine-tuning stability from the shared Stage-0 anchor) with decoupled L2-SP and per-step anchor blending/shrinkage ($\gamma_{\text{sp}} = 0.75$ on base layers, $0.06$ on final $1^3$ layer; Supplementary Table~S2):
\begin{align}
    \mathcal{L}_{\text{lumen}} &= 0.25\,\mathcal{L}_{\text{BCE-vol}}(\mathbf{P}_{\text{lumen}}, \tilde{Y}_{\text{bi}}) + 1.00\,\mathcal{L}_{\text{pos-recall}} \nonumber\\
    &\quad + 85.0\,\mathcal{L}_{\text{far-focal}} + 0.15\,\mathcal{L}_{\text{Tversky}}^{(0.45, 0.55)} + \mathcal{L}_{\text{L2-SP}} \nonumber\\
    &\quad + \mathcal{L}_{\text{disp}} + \mathcal{L}_{\text{ipsi}} + \mathcal{L}_{\text{lumen-128}}, \\
    \mathcal{L}_{\text{disp}} &= 0.15\,\frac{\sum_{\mathbf{x}, k} m_{\text{tube}}(\mathbf{x})\,\text{Smooth}L_1\!\left(\Delta\hat{u}_k - \Delta u_k^*; \beta_d\right)}{3 \sum_{\mathbf{x}} m_{\text{tube}}(\mathbf{x})}, \\
    \mathcal{L}_{\text{ipsi}} &= 0.16\,\Biggl( \sum_{c, \mathbf{x}} w_{\text{ipsi}, c}(\mathbf{x})\,\text{BCE}\!\left(\mathbf{P}_c(\mathbf{x}), Y_c(\mathbf{x})m_{\text{pos}, c}(\mathbf{x})\right) \nonumber\\
    &\quad + 1.8\,\frac{\sum_{c, \mathbf{x}} m_{\text{contra}, c}(\mathbf{x})\,\text{ReLU}(\mathbf{P}_c(\mathbf{x}) - 0.08)^2}{\sum_{c, \mathbf{x}} m_{\text{contra}, c}(\mathbf{x})} \nonumber\\
    &\quad + 0.4\,\frac{\sum_{c, \mathbf{x}} m_{\text{bg}}(\mathbf{x})\,\text{ReLU}(\mathbf{P}_c(\mathbf{x}) - 0.12)^2}{2\sum_{\mathbf{x}} m_{\text{bg}}(\mathbf{x})} \Biggr).
\end{align}
Here $\beta_d = 0.50\text{ mm}$ over $m_{\text{tube}}(\mathbf{x}) = \mathbb{I}(d_{\text{bi}}(\mathbf{x}) \le 3.5\text{ mm})$; $w_{\text{ipsi}, c}(\mathbf{x}) = 0.34\frac{m_{\text{pos}, c}(\mathbf{x})}{\max(1, \sum_{c', \mathbf{x}'} m_{\text{pos}, c'}(\mathbf{x}'))} + 0.44\frac{m_{\text{contra}, c}(\mathbf{x})}{\max(1, \sum_{c', \mathbf{x}'} m_{\text{contra}, c'}(\mathbf{x}'))} + 0.22\frac{m_{\text{bg}}(\mathbf{x})}{\max(1, 2\sum_{\mathbf{x}'} m_{\text{bg}}(\mathbf{x}'))}$ (where each mask category is normalized jointly across both territory channels $c' \in \{\text{LCA}, \text{RCA}\}$ and all spatial voxels $\mathbf{x}' \in \Omega_{64}$ so $\sum_{c, \mathbf{x}} w_{\text{ipsi}, c}(\mathbf{x}) = 1$); $\mathcal{L}_{\text{BCE-vol}}$, $\mathcal{L}_{\text{pos-recall}}$, $\mathcal{L}_{\text{far-focal}}$, and $\mathcal{L}_{\text{Tversky}}^{(0.45, 0.55)}$ sharpen vessel contrast ($\text{Dev-14}$ median centerline probability $0.5810$ vs.\ far-field $P_{95} = 0.0696$, $\text{SNR} = 8.34\times$); $\mathcal{L}_{\text{lumen-128}} = 0.06\,(\mathcal{L}_{\text{BCE}, 128} + 2.0\,\mathcal{L}_{\text{far}, 128} + 0.02\,\text{mean}(\Delta\mathbf{L}_{128}^2))$ supervises thin branches ($r_{\text{mm}} \le 1.1\text{ mm}$) and near-shell hard negatives ($3.0 < d \le 9.0\text{ mm} \wedge \mathbf{V}_{128} > 0.25$) via normalized voxel weights $w_{128}(\mathbf{x}) \propto 0.35\,\mathbb{I}(d_{\text{bi}, 128} \le 2.5)(1 + 1.5\,m_{\text{thin}, 128}) + 0.45\,m_{\text{shell-hn}, 128} + 0.20\,m_{\text{far}, 128}$; and $\mathcal{L}_{\text{reg-gt}} + \mathcal{L}_{\text{takeoff}}$ combines global-context Smooth-$L_1$ ($0.50, \beta=0.05$), mm-scale coordinate Smooth-$L_1$ ($0.12\,[\text{pred}+0.5\,\text{reg}+0.5\,\text{base}], \beta=2.0$), aortic-wall consistency $\mathcal{L}_{\text{ost-wall}}$ within $1.8\text{ mm}$ ($0.05\,[\text{pred}+0.5\,\text{reg}]$ Smooth-$L_1$, $\beta=1.0$), ostium L2-SP ($\lambda_{\text{sp,ost}} = 2.5$), $32^3$ heatmap BCE ($0.08, w_{\text{hm}} = 1 + 15\,H_c^*$), cosine takeoff loss ($0.15(1 - \hat{\mathbf{d}}_{0, c}^\top \mathbf{d}_{0, c}^*)$), and radius MSE ($0.15$; Supplementary Algorithm~A2 and Table~S2).

\vspace{-7pt}
\subsection{Stage $M_3$: Cylindrical Valsalva-STJ Wall Calibration, Sinus-Wall Rooting, and Native-$128^3$ Orifice Contact}
\vspace{-3pt}
For each artery $c \in \{\text{LCA}, \text{RCA}\}$ (Supplementary Algorithm~A3 and Table~S2), the ostium detector forms $\hat{\mathbf{o}}_{\text{neu}, c} = 0.55\,\hat{\mathbf{o}}_{\text{reg}, c} + 0.45\,\hat{\mathbf{o}}_{\text{det}, c}$ alongside aortic root center $\mathbf{c}_{\text{aorta}}$, $128^3$ wall distance $d_{\text{wall}}(\mathbf{x})$, and cylindrical wall distance $d_{\text{cyl}}(\mathbf{x}) = d_{\text{mm}}(\mathbf{x}, \partial\Omega_{\text{cyl}})$ across $[z_{\text{cut}} - 16\text{ mm}, z_{\text{cut}} + 18\text{ mm}]$, calibrating $d_{\text{wall}}(\mathbf{x}) \leftarrow \min(d_{\text{wall}}(\mathbf{x}), d_{\text{cyl}}(\mathbf{x}))$. Screening cranial ipsilateral components ($\rho_{\text{ipsi}, c}(\mathbf{x}) \ge \tau_\rho$) in Valsalva sectors, the solver walks the sinus wall to minimize $E_{\text{sinus}, c}(\mathbf{x}) = 1.20\,d_{\text{wall}}(\mathbf{x}) - 2.00\,\mathbf{P}_{\text{cal}}(\mathbf{x}) - z_{w, c}\,z_{\text{mm}}(\mathbf{x}) + 0.80\,\phi_{\text{sinus}, c}(\mathbf{x} - \mathbf{c}_{\text{aorta}}) + 0.06\,\max(0, \bar{d}_{\text{neu}, c}(\mathbf{x}) - 6.0)$, yielding $\mathbf{x}_c^*$ and blending detached roots ($d_{\text{cyl}}(\mathbf{x}_c^*) > 2.5\text{ mm}$) toward the cylindrical wall projection $\hat{\mathbf{o}}_{\text{wall}, c}$ when $\|\hat{\mathbf{o}}_{\text{reg}, c} - \hat{\mathbf{o}}_{\text{det}, c}\|_{\text{mm}} \le \tau_{\text{rd}}$. Within a $21^3$ window at $128^3$ on the 2-voxel Valsalva shell $\partial\Omega_{\text{sinus}}$, we fuse $\mathbf{P}_{\{128,\text{fuse}\}} = 0.65\,\text{Upsample}_{128}(\mathbf{P}_{\text{cal}}) + 0.35\,\mathbf{P}_{\text{cal}, 128}$, screen the eligible orifice set $\mathcal{Y}_{\text{elig}} = \{\mathbf{y} \in \partial\Omega_{\text{sinus}} : d_{\text{cyl}}(\mathbf{y}) \le 2.2\text{ mm} \wedge \rho_{\text{ipsi}, c}(\mathbf{y}) \ge 0.50 \wedge \|\mathbf{y} - \mathbf{x}_c^*\|_{\text{mm}} \le 6.5\text{ mm} \wedge \mathbf{P}_{\{128,\text{fuse}\}}(\mathbf{y}) \ge 0.34 \wedge \mathbf{V}_{128}(\mathbf{y}) \ge 0.38\}$, and when $|\mathcal{Y}_{\text{elig}}| \ge 3$ compute softmax contact expectation $\hat{\mathbf{o}}_{\text{orif}, c}$ with inverse temperature $\beta_{\text{orif}} = 4.0$ ($w(\mathbf{y}) \propto \exp(4.0(s_{\text{orif}}(\mathbf{y}) - \max s_{\text{orif}}))$, conventional temperature $T = 1/4.0 = 0.25$) over $s_{\text{orif}}(\mathbf{y}) = 2.5\,\mathbf{P}_{\{128,\text{fuse}\}}(\mathbf{y}) + 0.8\,\mathbf{V}_{128}(\mathbf{y}) - 0.55\,d_{\text{cyl}}(\mathbf{y}) - 0.12\,\|\mathbf{y} - \mathbf{x}_c^*\|_{\text{mm}}$ to set $\mathbf{o}_c^{(0)} = 0.35\,\mathbf{x}_c^* + 0.65\,\hat{\mathbf{o}}_{\text{orif}, c}$ (retaining $\mathbf{o}_c^{(0)} = \mathbf{x}_c^*$ when $|\mathcal{Y}_{\text{elig}}| < 3$), then apply monotonic pullback $\mathbf{o}_c^* = \mathbf{o}_c^{(0)} + 0.20\,\eta_{\text{ost}}\,\Delta\hat{\mathbf{u}}(\mathbf{o}_c^{(0)})$ ($\eta_{\text{ost}} = 0.65$) when lumen probability is non-decreasing and $d_{\text{wall}}(\mathbf{o}_c^*) \le d_{\text{wall}}(\mathbf{o}_c^{(0)}) + 0.5\text{ mm}$.

\vspace{-7pt}
\subsection{Stages $M_4$--$M_5$: Scalar-Potential Graph Arborescence, Normal-Plane Pullback, and Coverage-Guarded Collision Reduction}
\vspace{-3pt}
In Stage $M_4$ (Supplementary Algorithm~A4), we construct a scalar-potential-weighted graph arborescence. After cavity gating and coupling $\mathbf{P}_{128, \max} = \text{MaxPool}_{2\times 2\times 2}(\mathbf{P}_{\text{lumen}, 128})$ via the piecewise gated $128^3$ rescue rule $\mathbf{P}_{\text{coupled}}(\mathbf{x}) = 0.85\,\mathbf{P}_{\text{cal}}(\mathbf{x}) + 0.15\,\mathbf{P}_{128, \max}(\mathbf{x})$ on the rescue mask $\rho_{\text{ipsi}, c}(\mathbf{x}) \ge 0.82 \wedge \mathbf{P}_{\text{cal}}(\mathbf{x}) \ge 0.22 \wedge \mathbf{P}_{128, \max}(\mathbf{x}) \ge \max(0.40, \mathbf{P}_{\text{cal}}(\mathbf{x}))$ (with $\mathbf{P}_{128, \max} > \mathbf{P}_{\text{cal}}$ and $\mathbf{P}_{\text{cal}}(\mathbf{x})$ unchanged elsewhere; omitting $\rho_{\text{ipsi}, c} \ge 0.82$ when dual-channel fields are disabled) and post-tree ridge pullback ($0.88\,\mathbf{P}_{\text{smooth}} + 0.12\,\mathcal{G}_{0.35}(\mathbf{P}_{128, \max})$), we apply contralateral exclusion prior to 3D morphological skeletonization ($\tau_{\text{snr}} = 0.25$):
\begin{align}
    \mathbf{P}_{\text{skel}, c}(\mathbf{x}) &= \mathbf{P}_{\text{coupled}}(\mathbf{x})\cdot\Bigl(1 - \mathbb{I}\bigl(\mathbf{P}_{\bar{c}} > \mathbf{P}_c \wedge \mathbf{P}_{\bar{c}} \ge 0.08 \nonumber\\
    &\qquad\qquad\qquad\quad \wedge\, \rho_{\text{ipsi}, c} < 0.25\bigr)\Bigr), \\
    S_{\text{base}}(u, v) &= 0.42\bigl(\mathbf{P}_{\text{cal}}(\mathbf{x}_u) + \mathbf{P}_{\text{cal}}(\mathbf{x}_v)\bigr) \nonumber\\
    &\quad + 0.10\,\text{mean}(\mathbf{P}_{\text{seg}}) + 0.06\,\min(\mathbf{P}_{\text{seg}}).
\end{align}
Over extracted skeleton and local-maximum nodes (where a degenerate 16-point, $5\text{ mm}$ synthetic fallback stem is defined only for empty node sets and is triggered in $0/28$ \textbf{Dev-14} and $0/100$ \textbf{OOS-50} trees), intra-component edges are formed across $k=12$ nearest neighbors ($d_{ij} \le 5.0\text{ mm}$ by default, with a conditional extension to $d_{ij} \le 6.8\text{ mm}$ for sparse LCA skeletons with $M \le 150$ at distal root distances $\min(d_i, d_j) \ge 38.0\text{ mm}, \max(d_i, d_j) \ge 43.0\text{ mm}$ when 15-point interior segment support satisfies $\min_t \mathbf{P}_{\text{cal}} \ge 0.215 \wedge \text{mean}_t \mathbf{P}_{\text{cal}} \ge 0.345$) with potential-weighted cost $w_{ij} = d_{ij}\exp[-\alpha(P_i + P_j)/2]$ ($\alpha = 4.0$). Disconnected components are linked by candidate bridge edges weighted by $w_{\text{bridge}}(u, v) = d_{uv}\exp[-\alpha\,S_{\text{base}}(u, v)]$, where $S_{\text{base}}(u, v)$ in Eq.~(5) evaluates 7 interior segment points ($t \in [0.15, 0.85]$, scaled by $0.85 + 0.15(\rho_{\text{seg}, c}/0.18)$ when $\rho_{\text{seg}, c} < 0.18 \wedge \min(\mathbf{P}_{\text{seg}}) < 0.18$). In both the arborescence solver signature default and active LCA/RCA calls ($d_{\text{br,max}} = 45.0\text{ mm}$; $35.0\text{ mm}$ is only the internal bridging-helper default, with $12.0\text{ mm}$ ceiling for minor components), substantial components are defined by $|\mathcal{C}| \ge 12$, $(|\mathcal{C}| \ge 10 \wedge \bar{\mathbf{P}}_{\text{cal}} \ge 0.50)$, $z_{\min} \le 0.60$, or $d(\mathcal{C}, \mathbf{o}_c^*) \le 20.0\text{ mm}$. For LCA ($c = 0$), we connect substantial components (and medium pairs with $|\mathcal{C}_i|,|\mathcal{C}_j| \ge 6 \wedge \bar{\mathbf{P}}_{\text{cal}} \ge 0.48$) via 1 nearest candidate bridge per qualifying component pair (all-pairs nearest candidate bridges, $d_{uv} \le 45.0\text{ mm}$) and extract the single-source Dijkstra shortest-path tree ($\text{SPT}$) from $v_{\text{root}}$; for RCA ($c = 1$), to prevent cross-cavity jumps across the right atrioventricular groove, we iteratively join the nearest disconnected component to the growing root component via at most 1 minimum-cost bridge per step ($d_{uv} \le 45.0\text{ mm}$ for substantial components, relaxed to $50.0\text{ mm}$ when $|\mathcal{C}| \ge 20 \wedge \bar{\mathbf{P}}_{\text{cal}}(\mathcal{C}) \ge 0.65$ under a 15-point interior segment gate $\min_t \mathbf{P}_{\text{cal}} \ge 0.14 \wedge \text{mean}_t \mathbf{P}_{\text{cal}} \ge 0.24$ [or $\bar{\rho}_{\text{ipsi}} \ge 0.75 \wedge \bar{X}_{\text{LR}} \ge 0.55$]), extract the undirected minimum spanning tree ($\text{MST}$) over the connected subgraph, and orient edges outward from $v_{\text{root}}$ via breadth-first traversal ($\text{BFS}$), rooting at $\mathbf{o}_c^*$ with $\le 1.5\text{ mm}$ stems.

\textbf{Normal-Plane Pullback and Heuristic Radius-Ratio Pruning}: Using BFS from $\mathbf{o}_c^*$, the directed tree is decomposed into root-to-junction and junction-to-leaf trajectories $t = (v_0, \dots, v_K)$. Node radii are assigned via heuristic proximal-to-distal linear tapering ($1.8 \to 1.08\text{ mm}$ along the primary trunk and $1.2 \to 0.72\text{ mm}$ along side branches, with $0.75\,r_{\text{anc}}$ for recovered axial-ridge branches). Terminal trajectories are pruned over four passes using parent-to-child radius ratios $r_{\text{child}}/\max(r_{\text{parent}}, 0.25)$ and survival score $\mathcal{S}_{\text{ipsi}}(t) = 1 - (\mathcal{H}_{\text{tort}}(t) + \mathcal{H}_{\text{terr}}(t) + \mathcal{H}_{\text{cent}}(t))$, combining radius-uniformity-weighted chord-to-arc tortuosity penalty $\mathcal{H}_{\text{tort}}(t)$ ($L_{\text{arc}}(t)/d_{\text{span}}(t) > 1.32$), contralateral territory-drop penalty $\mathcal{H}_{\text{terr}}(t)$, and centripetal reentry penalty $\mathcal{H}_{\text{cent}}(t)$. For non-root nodes $\mathbf{x}_i$ ($i \ge 1$) with unit tangent $\mathbf{T}_i$, we iterate $K_{\text{disp}} = 3$ normal-plane steps coupled with normal-plane probability gradient $\mathbf{g}_{\perp, i}$, curvature gain $g_{\text{curv}, s} = 1 + 0.20\,\kappa_{s, i}$, and chord damping $s_{\text{chord}} = \min(1.0, \Delta s / 1.8)$:
\begin{align}
    \mathbf{x}_i^{\text{cand}} &= \mathbf{x}_i + \tilde{\eta}_{\text{disp}, i} \left(\mathbf{I} - \mathbf{T}_i \mathbf{T}_i^\top\right) \Delta\hat{\mathbf{u}}(\mathbf{x}_i) \nonumber\\
    &\quad + \tilde{\eta}_{g, i} \tanh\!\left(2.5\,\|\mathbf{g}_{\perp, i}\|_2\right) \frac{\mathbf{g}_{\perp, i}}{\|\mathbf{g}_{\perp, i}\|_2 + 10^{-6}},
\end{align}
where $\eta_{\text{disp}} = 0.65$, followed by normal-plane Laplacian smoothing ($w_{\text{smooth}} = 0.06$) and a four-stage midpoint subdivision and ridge cascade ($m = 1, \dots, 4$ down to $\Delta s \ge 1.06\text{ mm}$). Geometrically, interleaving normal-plane pullback with 4-stage midpoint subdivision suppresses secant chord shortcuts across curved segments while avoiding tangent-direction vertex bunching; empirically, this multi-stage cascade reduces centerline error and turning-angle roughness $\bar{\theta}_{\text{turn}}$ (Tables~\ref{tab:ablation_4arm} and~\ref{tab:stat_and_sensitivity} Part B), with residual localization error governed by local probability-ridge and displacement-field estimation rather than chord sagitta.

\textbf{Stage $M_5$: Coverage-Guarded Bilateral Collision Reduction and Subtree Handover}: Given pre-$M_5$ trees $(\mathcal{T}_{\text{LCA}}, \mathcal{T}_{\text{RCA}})$ ($N_{\min} = 15$ nodes), $\mathbf{P}_{\text{lumen}}(\mathbf{x})$, and ipsilateral ratio $\rho_{\text{ipsi}, c}(\mathbf{x}) = \mathbf{P}_c(\mathbf{x}) / (\mathbf{P}_{\text{LCA}}(\mathbf{x}) + \mathbf{P}_{\text{RCA}}(\mathbf{x}) + 10^{-6})$, Stage $M_5$ (Supplementary Algorithm~A5 and Table~S2) is a finite-pass (up to 4 passes) coverage-guarded collision-reduction heuristic that tracks modification flags $(\text{mod}_{\text{LCA}}, \text{mod}_{\text{RCA}})$ and transactionally applies five rules (committing updates only if post-prune $|\mathcal{V}_c| \ge N_{\min} = 15$): (1) \textit{Distal Void-Bridge Pruning} removes distal subtrees across jump edges $> 22.0\text{ mm}$ at root distance $> 100.0\text{ mm}$ lacking midpoint lumen and ipsilateral support; and over up to four passes while inter-tree distance $\min d(\mathcal{T}_{\text{LCA}}, \mathcal{T}_{\text{RCA}}) < 2.5\text{ mm}$, (2a) \textit{Cross-Cavity Jump Handover} transfers colliding subtrees reached via $> 15.0\text{ mm}$ cross-cavity jumps with low ipsilateral support ($\bar{\rho}_{\text{ipsi}, A} \le 0.65$) onto the contralateral tree; (2b) \textit{Inbound Tip-Capture Pruning} prunes distal subtrees whose leaves are $100\%$ captured within $2.5\text{ mm}$ of the dominant contralateral tree; and (2c--2d) \textit{Coverage-Guarded Over-Extension and Residual Handover} graft uncovered high-confidence distal branches ($|\mathcal{C}_{\{u\}}| \ge 5$, $\bar{\mathbf{P}}_{\text{lumen}} \ge 0.60$, $\bar{\rho}_{\text{ipsi}, A} \ge 0.55$, bridge $\le 8.0\text{ mm}$) onto the true host tree $\mathcal{T}_A$ before pruning colliding over-extended ($d_B > d_A + 35.0\text{ mm}$) or residual ($d_B > 90.0\text{ mm}$) subtrees from $\mathcal{T}_B$. Because Stage~$M_5$ intentionally preserves high-confidence branches when neither jump nor low-territory gates trigger, post-$M_5$ separation satisfies $\min d \ge 2.5\text{ mm}$ in $13/14$ \textbf{Dev-14} cases ($1/14$ $<2.5\text{ mm}$: Case 135 at $0.8463\text{ mm}$; min/med/mean $0.85/27.83/25.58\text{ mm}$) and $48/50$ \textbf{OOS-50} cases ($2/50$ $<2.5\text{ mm}$: Case 270 at $2.3300\text{ mm}$, Case 295 at $1.5434\text{ mm}$; min/med/mean $1.54/21.50/23.74\text{ mm}$).

\vspace{-4pt}
\section{Experiments and Results}
\label{sec:experiments}

\begin{figure*}[!t]
\centering
\includegraphics[width=0.84\textwidth]{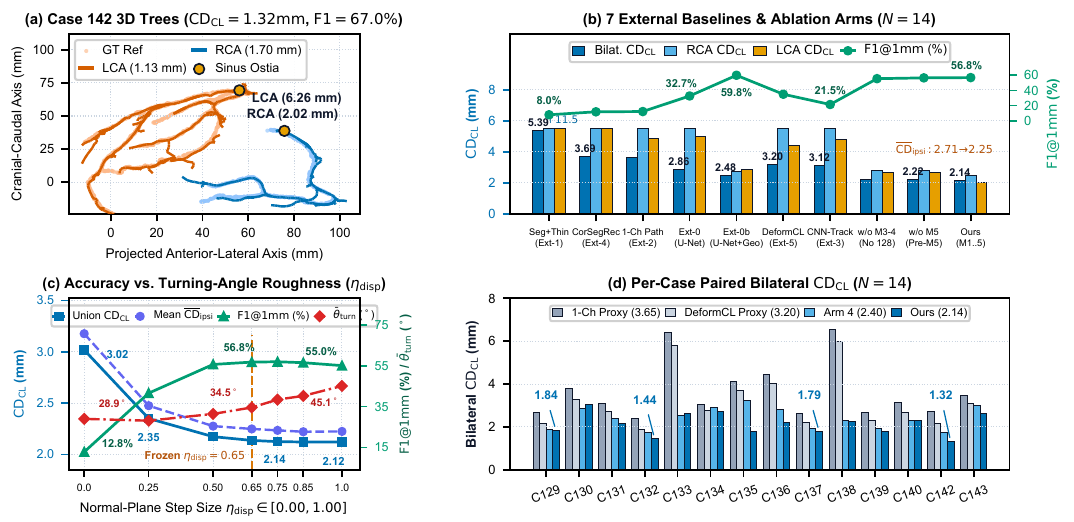}
\vspace{-9pt}
\caption{\fontsize{7.2}{7.8}\selectfont Qualitative and quantitative evaluation of the canonical $(Z_{\text{IS}}, Y_{\text{PA}}, X_{\text{LR}})$ world-axis bilateral coronary arborescence framework on Dev-14 ($N=14$ unseen volumes, $28$ trees): (a) 3D bilateral reconstruction on held-out Case 142 ($\text{CD}_{\text{CL}}^{\text{union}} = 1.32\text{ mm}$, $\overline{\text{CD}}_{\text{ipsi}} = 1.42\text{ mm}$, $\text{F1}@1\text{mm} = 67.0\%$, LCA/RCA $\text{CD}_{\text{CL}} = 1.13 / 1.70\text{ mm}$, 3D inter-tree separation $8.98\text{ mm}$, $\beta_0=1, \beta_1=0$); (b) comparison against standalone single-task 3D U-Net baselines (Ext-0 and Ext-0b), five controlled same-backbone proxies (Ext-1..5), and ablation progression; (c) 7-point sensitivity analysis across $\eta_{\text{disp}} \in \{0.00, 0.25, 0.50, 0.65, 0.75, 0.85, 1.00\}$ showing the frozen point $\eta_{\text{disp}} = 0.65$ and turning-angle roughness $\bar{\theta}_{\text{turn}}$; and (d) per-case paired $\text{CD}_{\text{CL}}^{\text{union}}$ across all 14 unseen cases.}
\label{fig:bilateral_benchmark}
\end{figure*}
\subsection{Disjoint Four-Cohort Split, Reference Provenance, Freeze Protocol, and Multi-Tier Tree-Topology Metrics}
On the public 3D CCTA ImageCAS cohort~\cite{zeng2023imagecas} within standardized GT-foreground-conditioned $128^3$ coronary ROIs ($\pm 15$ raw-voxel margin, $\approx 9.8\text{--}12.0\text{ mm}$; seed-free inference without ground-truth ostium or centerline seeds inside the ROI), we enforce a disjoint four-cohort partition and freeze protocol: (i) \textbf{Training Cohort (Train-35)}: $35$ volumes (Cases $1, 10\text{--}14, 100\text{--}128$) used to train all network weights (conditional 30-epoch fine-tuning stability across $\text{seeds} \in \{20261012, 20261013, 20261014\}$ from the shared Stage-0 anchor); (ii) \textbf{Held-Out Development Cohort (Dev-14, Stratified Blocks 0..4)}: $14$ unseen volumes (Cases 129--140, 142, 143; $28$ trees) on which all downstream solver and baseline hyperparameters ($\eta_{\text{disp}} = 0.65$, $\tau_{\text{snr}} = 0.25$, $M_3$--$M_5$ thresholds) are selected and frozen (consequently, \textbf{Dev-14} carries development-set hyperparameter-selection bias and serves for ablation/calibration rather than unbiased generalization claims); (iii) \textbf{Locked-Weight Out-of-Sample Cohort (OOS-50)}: $50$ unseen volumes (Cases 251--300; $100$ trees; plus descriptive pooled $N=64$ summary across $128$ unseen trees) serving as the primary generalization benchmark evaluated with locked neural weights and frozen thresholds (locked-weight reused out-of-sample evaluation mode, binding separate prediction-generation and evaluation timestamps and SHA-256 source-code digests in the evaluation receipt); and (iv) \textbf{Quarantined Vault (Vault-50)}: Cases 201--250 remain strictly isolated ($0$ reads, $0$ writes). Reference centerlines and radii $r_{\text{mm}}$ are derived deterministically from expert binary lumen masks via 3D morphological skeletonization and Euclidean distance transform (EDT). Beyond constructive shortest-path/MST invariants ($\beta_0=1, \beta_1=0$) and co-primary \textbf{Bilateral Union $\text{CD}_{\text{CL}}^{\text{union}}$} and \textbf{Mean Ipsilateral $\overline{\text{CD}}_{\text{ipsi}} = \frac{1}{2}(\text{CD}_{\text{LCA}} + \text{CD}_{\text{RCA}})$} ($\Delta_{\text{ipsi}-\text{union}}$, $\text{P2G}$, $\text{G2P}$, $\text{HD}_{95}$, $\text{F1}@1/2\text{mm}$, Ostium Med/Mean Error), our bilateral scaffold evaluator (Supplementary Table~S2) implements four deterministic tiers: (1) \textbf{Tier 1 (Spatial Corridor Skeletal Tree)}: reference graphs use $2.2\text{ mm}$ Poisson-disk clustering, $\text{MST}$ adjacency with $\times 12.0$ soft penalty for chords $>6.5\text{ mm}$ and $25.0\text{ mm}$ hard cap, and 4-pass $<5.5\text{ mm}$ terminal twig pruning ($\ell_{\min}^{\text{twig}} = 5.5\text{ mm}$; deterministic cutoff sensitivity across $2.0 / 3.5 / 5.5\text{ mm}$ yields $78.43\% / 78.56\% / 79.95\%$ junction-corridor F1 and $83.17\% / 83.28\% / 83.80\%$ branch-corridor F1 on \textbf{Dev-14}, and $74.22\% / 75.86\% / 76.33\%$ junction-corridor F1 and $78.84\% / 79.09\% / 79.65\%$ branch-corridor F1 on \textbf{OOS-50}), while predicted directed graphs apply 4-pass $<5.5\text{ mm}$ twig pruning, $4.0\text{ mm}$ minimum branch length, and $3.5\text{ mm}$ bifurcation deduplication to score \textbf{Junction-Corridor Rec/Pr/F1} ($3.0\text{ mm}$ tolerance), \textbf{Branch-Corridor F1 [$\text{FP}\%$]} ($2.5\text{ mm}$ tolerance, $\ge 50\%$ point coverage), \textbf{Leaf-to-Curve Distance}, \textbf{Length Ratio}, and \textbf{Prox/Mid/Distal $\text{Rec}@2\text{mm}$}; (2) \textbf{Tier 2 (1-to-1 Simplified Skeletal Matching on Unpruned Graphs)}: on unpruned reference and predicted trees ($\mathcal{T}_{\text{unpruned}}, \mathcal{T}_{\text{actual,unpruned}}$), 4-pass $<3.0\text{ mm}$ terminal spur pruning and $2.0\text{ mm}$ connected-component junction/leaf merging are followed by $5.0\text{ mm}$ 1-to-1 Hungarian bipartite matching for \textbf{1-to-1 Bifurcation F1 / Loc Error} and \textbf{1-to-1 Leaf Distance}; (3) \textbf{Tier 3 (Continuous-Edge $1.0\text{ mm}$ Resampling)}: $\text{CD}_{\text{CL}}^{\text{arc1mm}}$ and $\overline{\text{CD}}_{\text{ipsi}}^{\text{arc1mm}}$ resample every non-branching predicted polyline at $1.0\text{ mm}$ arc-length steps ($0.15\text{ mm}$ endpoint deduplication) against reference trees clustered at $0.8\text{ mm}$ ($\times 12.0$ penalty $>5.0\text{ mm}$, $25.0\text{ mm}$ cap, 3 passes of $<1.8\text{ mm}$ terminal spur pruning); and (4) \textbf{Tier 4 (Reference Thick-Stem Ostia)}: extracted in the canonical $(Z_{\text{IS}}, Y_{\text{PA}}, X_{\text{LR}})$ cache builder via $r_{\text{mm}} \ge \max(1.10\text{ mm}, P_{75}(r_{\text{mm}}))$ and $Z_{\text{IS}} \ge 0.45$, within the top $0.035$ $Z_{\text{IS}}$ band for LCA and $Z_{\text{IS}} \ge 0.52$ for RCA ($3.5 / 3.0\text{ mm}$ sinus-wall band). Given $128^3$ spacing ($\sim 1.35\text{ mm}$) and 3-seed SD ($0.0369 / 0.0415\text{ mm}$), we reference a heuristic sub-voxel margin ($\delta_{\text{tol}} = 0.15\text{ mm}$) alongside paired Wilcoxon signed-rank tests (pre-specified primary full-cohort $p_{\text{zsplit}}$ and tie-handling sensitivity non-zero pair $p_{\text{wilcox}}$ with $N_{\Delta \neq 0}$) under pre-specified multiplicity families: \textit{external\_7} ($M=7$) and \textit{loo\_9} ($M=9$) on \textbf{Dev-14}, plus \textit{external\_7} ($M=7$), \textit{core\_mechanism\_ablation\_4} ($M=4$), \textit{ablation\_4} ($M=4$), and combined 8-arm ablation (\textit{ablation\_combined\_8}, $M=8$) on \textbf{OOS-50}.

\begin{table*}[!t]
\centering
\caption{Held-Out Development Cohort (Dev-14, $N=14$ Unseen Volumes, $28$ Coronary Trees) Bilateral Union, Continuous-Edge 1.0~mm Arc-Length Resampled ($\text{CD}_{\text{CL}}^{\text{arc1mm}}$), Ipsilateral, and Actual-Graph Topology Benchmark in Canonical $(Z_{\text{IS}}, Y_{\text{PA}}, X_{\text{LR}})$ Space Against Standalone Single-Task 3D U-Net Baselines (Ext-0, Ext-0b) and Five Controlled Same-Backbone Downstream Proxies (Ext-1..5).}
\label{tab:main_benchmark}
\vspace{-7pt}
{\fontsize{6.5}{6.8}\selectfont\thinmuskip=0mu\medmuskip=0mu\thickmuskip=0mu
\renewcommand{\arraystretch}{0.80}
\setlength{\tabcolsep}{1.1pt}
\begin{tabular}{@{}lccccccc@{}}
\toprule
Paradigm & Bilat $\text{CD}_{\text{CL}}^{\text{union}}$ & $\text{CD}_{\text{CL}}^{\text{arc1mm}}$ & Ipsi $\overline{\text{CD}}_{\text{ipsi}}$ & P2G/G2P & $\text{F1}@1/2$[Prox/ & Junc R/P/F1 & Br F1[FP]/End \\
 & [Med](mm) & [Ost Med/Mn] & [$\Delta_{\text{ipsi-u}}$/$\overline{\text{CD}}_{\text{ipsi}}^{\text{arc}}$] & [$\text{HD}_{95}$](mm) & Mid/Dist](\%) & [Loc/1-1 F1](\%) & [Len](mm) \\
\midrule
Hist.\ 1-Tree & $16.18 \pm 1.50$[$16.2$] & ---[$14.3\!/\!16.4$] & --- & $17.5\!/\!14.8$[$37.5$] & --- & --- & --- \\
Ext-0 U-Net~\cite{ronneberger2015u} & $2.8561 \pm 0.4344$[$2.74$] & $3.4785$[$13.7\!/\!14.1$] & $5.6028$[$+2.75\!/\!6.83$] & $2.93\!/\!2.78$[$15.9$] & $32.7\!/\!71.5$[$81\!/\!72\!/\!55$] & $77\!/\!46\!/\!52.4$[$5.4\!/\!\mathbf{38.0}$] & $68.6$[$37$]\!/\!$12.4$[$1.1$] \\
Ext-0b +$M_{3..5}$ & $2.4773 \pm 0.6818$[$2.25$] & $2.7398$[$13.7\!/\!14.1$] & $2.8155$[$+0.34\!/\!3.14$] & $\mathbf{1.77}\!/\!3.19$[$15.5$] & $\mathbf{59.8}\!/\!76.7$[$\mathbf{85}\!/\!76\!/\!56$] & $78\!/\!\mathbf{84}\!/\!77.3$[$\mathbf{1.5}\!/\!36.4$] & $81.2$[$17$]\!/\!$7.88$[$0.9$] \\
Ext-1 Seg+Sk.~\cite{shit2021cldice} & $5.3926 \pm 1.5744$[$5.01$] & $6.1892$[$6.22\!/\!7.91$] & $9.5487$[$+4.16\!/\!10.4$] & $7.32\!/\!3.47$[$25.6$] & $8.0\!/\!34.6$[$36\!/\!41\!/\!34$] & $48\!/\!17\!/\!20.7$[$8.4\!/\!11.7$] & $38.0$[$69$]\!/\!$16.5$[$1.9$] \\
Ext-2 1Ch~\cite{wolterink2019coronary} & $3.6539 \pm 1.3351$[$3.13$] & $3.9841$[$6.22\!/\!7.91$] & $5.7136$[$+2.06\!/\!5.94$] & $3.32\!/\!3.99$[$19.5$] & $12.3\!/\!51.8$[$51\!/\!58\!/\!48$] & $76\!/\!58\!/\!58.8$[$4.0\!/\!19.4$] & $66.9$[$36$]\!/\!$9.68$[$1.2$] \\
Ext-3 Tracker~\cite{wolterink2019coronary} & $3.1151 \pm 0.6172$[$3.05$] & $3.2494$[$\mathbf{4.69}\!/\!4.99$] & $6.7790$[$+3.66\!/\!6.87$] & $3.49\!/\!2.74$[$15.1$] & $21.5\!/\!60.4$[$67\!/\!71\!/\!55$] & $76\!/\!51\!/\!54.3$[$4.4\!/\!30.3$] & $67.8$[$37$]\!/\!$10.9$[$1.3$] \\
Ext-4 CorSeg.~\cite{qiu2025topology} & $3.6919 \pm 0.6773$[$3.69$] & $4.0366$[$\mathbf{4.69}\!/\!\mathbf{4.93}$] & $6.1225$[$+2.43\!/\!6.34$] & $3.93\!/\!3.45$[$17.0$] & $11.9\!/\!44.2$[$42\!/\!48\!/\!39$] & $48\!/\!69\!/\!50.6$[$3.8\!/\!25.0$] & $57.7$[$40$]\!/\!$11.1$[$1.1$] \\
Ext-5 DefCL~\cite{zhao2025deformcl} & $3.2024 \pm 1.2929$[$2.75$] & $3.5628$[$\mathbf{4.69}\!/\!\mathbf{4.93}$] & $5.0817$[$+1.88\!/\!5.34$] & $2.82\!/\!3.58$[$19.4$] & $35.0\!/\!68.1$[$72\!/\!75\!/\!59$] & $76\!/\!68\!/\!65.6$[$3.2\!/\!22.1$] & $75.8$[$24$]\!/\!$9.37$[$1.2$] \\
\midrule
1-Ch Geodesic & $3.6295 \pm 1.2898$[$3.13$] & $3.9567$[$4.94\!/\!5.26$] & $5.5402$[$+1.91\!/\!5.77$] & $3.31\!/\!3.94$[$19.5$] & $12.3\!/\!51.9$[$51\!/\!58\!/\!48$] & $76\!/\!59\!/\!59.6$[$3.8\!/\!20.5$] & $67.1$[$35$]\!/\!$9.55$[$1.2$] \\
Dual-Ch Surv. & $3.0999 \pm 0.4420$[$3.11$] & $3.3311$[$4.94\!/\!5.26$] & $3.7464$[$+0.65\!/\!3.93$] & $3.34\!/\!2.86$[$13.4$] & $12.6\!/\!52.8$[$51\!/\!58\!/\!50$] & $82\!/\!62\!/\!66.2$[$2.6\!/\!20.2$] & $71.2$[$32$]\!/\!$7.77$[$1.2$] \\
Sinus-Centrip. & $2.3986 \pm 0.5026$[$2.36$] & $2.6594$[$4.94\!/\!5.26$] & $3.0383$[$+0.64\!/\!3.24$] & $2.58\!/\!2.22$[$13.4$] & $56.1\!/\!75.6$[$82\!/\!82\!/\!65$] & $\mathbf{85}\!/\!70\!/\!72.7$[$1.9\!/\!21.9$] & $79.5$[$23$]\!/\!$7.34$[$1.2$] \\
Sub-Vox Ost+Br & $2.3227 \pm 0.4757$[$2.32$] & $2.6288$[$\mathbf{4.69}\!/\!5.12$] & $2.9099$[$+0.59\!/\!3.20$] & $2.45\!/\!2.19$[$12.8$] & $57.4\!/\!76.1$[$83\!/\!82\!/\!65$] & $84\!/\!72\!/\!73.7$[$1.9\!/\!20.9$] & $80.4$[$22$]\!/\!$7.24$[$1.2$] \\
Curv.\ 3-Stage & $2.2719 \pm 0.4312$[$2.32$] & $2.6144$[$4.79\!/\!5.20$] & $2.7920$[$+0.52\!/\!3.16$] & $2.37\!/\!\mathbf{2.18}$[$11.9$] & $55.3\!/\!76.6$[$83\!/\!83\!/\!67$] & $84\!/\!76\!/\!76.0$[$1.9\!/\!20.8$] & $81.9$[$20$]\!/\!$6.82$[$1.2$] \\
w/o Stage-$M_5$ & $2.2234 \pm 0.4668$[$2.29$] & $2.5680$[$\mathbf{4.69}\!/\!\mathbf{4.93}$] & $2.7145$[$+0.49\!/\!3.10$] & $2.27\!/\!\mathbf{2.18}$[$11.7$] & $56.4\!/\!77.3$[$83\!/\!\mathbf{83}\!/\!\mathbf{68}$] & $83\!/\!78\!/\!78.2$[$1.9\!/\!20.8$] & $82.3$[$18$]\!/\!$6.73$[$1.2$] \\
\textbf{Prop.\ ($M_{1..5}$)} & $\mathbf{2.1362 \pm 0.5041}$[$\mathbf{2.17}$] & $\mathbf{2.3798}$[$\mathbf{4.69}\!/\!\mathbf{4.93}$] & $\mathbf{2.2480}$[$\mathbf{+0.11}\!/\!\mathbf{2.50}$] & $2.07\!/\!2.20$[$\mathbf{10.8}$] & $56.8\!/\!\mathbf{77.7}$[$83\!/\!\mathbf{83}\!/\!\mathbf{68}$] & $83\!/\!81\!/\!\mathbf{80.0}$[$\mathbf{1.5}\!/\!21.1$] & $\mathbf{83.8}$[$\mathbf{15.8}$]\!/\!$\mathbf{6.09}$[$\mathbf{1.1}$] \\
\bottomrule
\multicolumn{8}{@{}p{0.96\textwidth}@{}}
{\fontsize{6.5}{7.1}\selectfont \textit{Notes}: Columns report Mean $\pm$ SD [Median] $\text{CD}_{\text{CL}}^{\text{union}}$, continuous-edge $\text{CD}_{\text{CL}}^{\text{arc1mm}}$ [Ostium Med/Mean], $\overline{\text{CD}}_{\text{ipsi}}$ [$\Delta_{\text{ipsi-u}} / \overline{\text{CD}}_{\text{ipsi}}^{\text{arc1mm}}$], P2G/G2P [$\text{HD}_{95}$], $\text{F1}@1/2\text{mm}$ [Prox/Mid/Distal $\text{Rec}@2\text{mm}$], $3.0\text{ mm}$ Junction Rec/Pr/F1 [Loc.\ Err / 1-to-1 Bifurcation F1 on $\mathcal{G}_{\text{actual}}$], and Branch F1 [FP\%] / Leaf-to-Curve [Length Ratio]. Ext-0 and Ext-0b use an independently trained $481,681$-parameter 3D U-Net on $64^3$ average-pooled CCTA inputs supervised by the same derived tubular targets (centerline-radius-dilated binary masks for Dice + Gaussian soft targets for BCE) rather than raw full-resolution voxel masks; Ext-1..5 are controlled same-backbone downstream proxies. Published DeformCL~\cite{zhao2025deformcl} extracts image features and coarse vessel segmentation from 3D medical images to generate and deform an adaptive centerline template graph without autonomous ostium detection or explicit LCA/RCA territory decoupling, while RefTr~\cite{naeem2025reftr} and VesselTok~\cite{prabhakar2026vesseltok} address recurrent trajectory refinement and generative vessel synthesis.}
\end{tabular}}
\vspace{-2pt}
\end{table*}

\begin{table*}[!t]
\centering
\caption{Component Ablation Study Across Held-Out Dev-14 ($28$ Trees) in Canonical $(Z_{\text{IS}}, Y_{\text{PA}}, X_{\text{LR}})$ Space: Part I Summarizes Key Forward Build Milestones (Full 17-Arm Progression in Supplementary Table~S4) and Part II Reports All Nine True Leave-One-Out (w/o Single Module) and $2\times 2$ Pullback Factorial Arms (LOO-1, LOO-1a, LOO-1b, LOO-2..7) From the Full Proposed Model ($M_{1..5}$).}
\label{tab:ablation_4arm}
\vspace{-7pt}
{\fontsize{6.5}{6.8}\selectfont\thinmuskip=0mu\medmuskip=0mu\thickmuskip=0mu
\renewcommand{\arraystretch}{0.80}
\setlength{\tabcolsep}{1.2pt}
\begin{tabular}{@{}lcccccccc@{}}
\toprule
Ablation Arm & Active Modules & Bilat $\text{CD}_{\text{CL}}^{\text{union}}$ & Arc & Ipsi $\overline{\text{CD}}_{\text{ipsi}}$ & P2G/G2P & $\text{HD}_{95}$ & $\text{F1}@1/2$(\%) & Junc/Br F1 \\
 & ($\Delta\hat{\mathbf{u}}$, Ipsi, Sinus, $M_{3\text{--}5}$) & (mm) & $\text{CD}_{\text{CL}}^{\text{arc1mm}}$ & [$\Delta_{\text{ipsi-u}}$] & (mm) & (mm) & [LCA/RCA/Ost] & [FP/Endpt] \\
\midrule
\multicolumn{9}{@{}l@{}}{\textbf{Part I: Primary Forward Incremental Build Milestones (See Supplementary Table~S4 for All 17 Cumulative Build Steps)}} \\
\midrule
1. Recomb. 1-Ch & Base 1-Ch Geodesic & $3.6295 \pm 1.2898$ & $3.9567$ & $5.5402$[$+1.91$] & $3.31/3.94$ & $19.45$ & $12.3/51.9$[$4.84/6.24/4.94$] & $59.6/67.1$[$35.4/9.55$] \\
2. + Ipsi Decoup. & + Dual-Ch Ipsi + Surv & $3.0999 \pm 0.4420$ & $3.3311$ & $3.7464$[$+0.65$] & $3.34/2.86$ & $13.42$ & $12.6/52.8$[$3.74/3.75/4.94$] & $66.2/71.2$[$32.4/7.77$] \\
3. Sub-Voxel Only & + $\Delta\hat{\mathbf{u}}$ (w/o Dual-Ch) & $2.9407 \pm 1.3351$ & $3.3053$ & $4.8196$[$+1.88$] & $2.56/3.32$ & $19.45$ & $55.2/73.5$[$4.16/5.48/4.94$] & $68.4/75.8$[$23.4/9.11$] \\
4. + Centrip. Surv. & + $\Delta\hat{\mathbf{u}}$ + Ipsi + Sinus & $2.3986 \pm 0.5026$ & $2.6594$ & $3.0383$[$+0.64$] & $2.58/2.22$ & $13.38$ & $56.1/75.6$[$3.07/3.01/4.94$] & $72.7/79.5$[$23.2/7.34$] \\
7. + Ost.+Bridge & + Sub-Vox Ost + Brg & $2.3227 \pm 0.4757$ & $2.6288$ & $2.9099$[$+0.59$] & $2.45/2.19$ & $12.78$ & $57.4/76.1$[$2.89/2.93/\mathbf{4.69}$] & $73.7/80.4$[$21.7/7.24$] \\
11. + Curv. 3-Stage & + 3-Stg Curv Pullback & $2.2719 \pm 0.4312$ & $2.6144$ & $2.7920$[$+0.52$] & $2.37/2.18$ & $11.93$ & $55.3/76.6$[$2.74/2.85/4.79$] & $76.0/81.9$[$19.6/6.82$] \\
\midrule
\multicolumn{9}{@{}l@{}}{\textbf{Part II: Leave-One-Out (w/o Single Module) Orthogonal Ablation \& $2\times 2$ Pullback Factorial From Full Model ($M_{1..5}$)}} \\
\midrule
LOO-1. w/o $\Delta\hat{\mathbf{u}}$ (Neither) & w/o $\Delta\hat{\mathbf{u}}$, w/o Ridge & $3.0174 \pm 0.4533$ & $3.1931$ & $3.1755$[$+0.16$] & $3.16/2.88$ & $12.67$ & $12.8/53.6$[$2.95/3.40/\mathbf{4.69}$] & $67.7/73.4$[$29.6/7.00$] \\
LOO-1a. Ridge Only ($\eta\!=\!0$) & Ridge Only ($\Delta\hat{\mathbf{u}}\!=\!\mathbf{0}$) & $2.7709 \pm 0.4574$ & $3.0398$ & $2.9285$[$+0.16$] & $2.77/2.77$ & $\mathbf{10.70}$ & $15.0/57.6$[$2.69/3.16/\mathbf{4.69}$] & $74.7/75.9$[$26.8/6.52$] \\
LOO-1b. $\Delta\hat{\mathbf{u}}$ Only & $\Delta\hat{\mathbf{u}}$ Only ($\mathbf{g}_{\perp}\!=\!\mathbf{0}$) & $\mathbf{2.1033 \pm 0.4953}$ & $\mathbf{2.3442}$ & $\mathbf{2.2029}$[$\mathbf{+0.10}$] & $\mathbf{2.04}/\mathbf{2.17}$ & $10.71$ & $\mathbf{58.9}/\mathbf{77.7}$[$2.02/\mathbf{2.39}/\mathbf{4.69}$] & $79.9/83.3$[$16.4/6.10$] \\
LOO-2. w/o Dual-Ch Ipsi & 1-Ch $M_{4,5}$ ($M_3$ Fix.) & $2.5397 \pm 1.0365$ & $2.9282$ & $3.8941$[$+1.35$] & $2.30/2.78$ & $15.34$ & $55.7/75.6$[$3.02/4.76/\mathbf{4.69}$] & $75.0/79.1$[$18.4/8.13$] \\
LOO-3. w/o Centrip. Prun. & $\mathcal{H}_{\text{cent}}(t) = 0$ & $2.1388 \pm 0.4971$ & $2.3981$ & $2.2533$[$+0.11$] & $2.10/2.18$ & $10.81$ & $56.8/\mathbf{77.7}$[$\mathbf{2.01}/2.49/\mathbf{4.69}$] & $\mathbf{80.1}/83.7$[$16.1/6.10$] \\
LOO-4. w/o Cyl.\ Sinus & Raw Detector Ostia & $2.1563 \pm 0.5282$ & $2.3984$ & $2.2690$[$+0.11$] & $2.08/2.24$ & $10.96$ & $56.6/77.6$[$2.02/2.52/6.18$] & $77.6/82.8$[$17.1/6.10$] \\
LOO-5. w/o $M_3$ Orif. & +$M_{4,5}$ (w/o $M_3$) & $2.1356 \pm 0.5054$ & $2.3795$ & $2.2480$[$+0.11$] & $2.07/2.20$ & $10.75$ & $56.7/\mathbf{77.7}$[$2.02/2.48/4.79$] & $80.0/\mathbf{83.8}$[$\mathbf{15.8}/\mathbf{6.09}$] \\
LOO-6. w/o $M_4$ $128^3$ & +$M_{3,5}$ (w/o $M_4$) & $2.1703 \pm 0.4462$ & $2.4233$ & $2.2806$[$+0.11$] & $2.14/2.20$ & $10.96$ & $55.7/77.1$[$2.05/2.51/\mathbf{4.69}$] & $79.3/\mathbf{83.8}$[$17.1/6.19$] \\
LOO-7. w/o $M_5$ Excl. & +$M_{3,4}$ (Pre-$M_5$) & $2.2234 \pm 0.4668$ & $2.5680$ & $2.7145$[$+0.49$] & $2.27/2.18$ & $11.69$ & $56.4/77.3$[$2.64/2.79/\mathbf{4.69}$] & $78.2/82.3$[$18.2/6.73$] \\
\textbf{Full Proposed ($M_{1..5}$)} & \textbf{All Modules Active} & $2.1362 \pm 0.5041$ & $2.3798$ & $2.2480$[$+0.11$] & $2.07/2.20$ & $10.79$ & $56.8/\mathbf{77.7}$[$2.02/2.48/\mathbf{4.69}$] & $80.0/\mathbf{83.8}$[$\mathbf{15.8}/\mathbf{6.09}$] \\
\bottomrule
\multicolumn{9}{@{}p{0.97\textwidth}@{}}
{\fontsize{6.5}{7.1}\selectfont \textit{Notes}: $\Delta_{\text{ipsi-u}} = \overline{\text{CD}}_{\text{ipsi}} - \text{CD}_{\text{CL}}^{\text{union}}$; $[\text{LCA/RCA/Ost}]$ reports mean LCA $\text{CD}_{\text{CL}}$, RCA $\text{CD}_{\text{CL}}$, and Ostium median error (mm); $[\text{FP/Endpt}]$ reports Branch-Corridor FP rate (\%) and Leaf-to-Curve distance (mm). Full 17-step build metrics (including intermediate sub-steps 5, 6, 8--10, and 12--16) are detailed in Supplementary Table~S4.}
\end{tabular}}
\vspace{-2pt}
\end{table*}

\begin{table*}[!t]
\centering
\caption{Paired Non-Parametric Statistical Significance With Holm--Bonferroni ($p_{\text{H}}$) and Benjamini--Hochberg FDR ($q_{\text{BH}}$) Multiplicity Control (Dev-14), Extended 7-Point Sub-Voxel Step-Size ($\eta_{\text{disp}} \in [0.00, 1.00]$) Sensitivity With Turning-Angle Roughness ($\bar{\theta}_{\text{turn}}$), and Locked-Weight Out-of-Sample Generalization Across OOS-50 (Cases 251--300, $100$ Trees) and Descriptive Pooled $N=64$ Unseen Cohort ($128$ Trees).}
\label{tab:stat_and_sensitivity}
\vspace{-7pt}
{\fontsize{6.5}{6.7}\selectfont\thinmuskip=0mu\medmuskip=0mu\thickmuskip=0mu
\renewcommand{\arraystretch}{0.80}
\setlength{\tabcolsep}{1.2pt}
\begin{tabular}{@{}lccccccc@{}}
\toprule
\multicolumn{8}{@{}l@{}}{\textbf{Part A: Paired Statistical Significance \& Multiplicity Control vs.\ Baselines ($M=7$) \& LOO Arms ($M=9$) (Dev-14, $N=14$)}} \\
\midrule
Comparison & $\Delta\text{CD}_{\text{CL}}^{\text{union}}$[95\%CI] & Wilc.\ $p$($p_{\text{H}}/q_{\text{BH}}$) & $\Delta\overline{\text{CD}}_{\text{ipsi}}$($p/p_{\text{H}}$) & $\Delta\text{F1}@1$[CI] & Wilc.\ $p$ & $\Delta\text{L}/\text{R}$($p_{\text{L}}/p_{\text{R}}$) & Fam. \\
\midrule
vs.\!Ext-0~\cite{ronneberger2015u} & $-0.720$[$-0.95,\!-0.47$] & $\mathbf{6.1\text{e-}4}$\!($\mathbf{.0012}/\mathbf{7.1\text{e-}4}$) & $-3.36$\!($\mathbf{1.2\text{e-}4}/\mathbf{8.5\text{e-}4}$) & $+24.1$[$+22,\!+26$] & $\mathbf{1.2\text{e-}4}$ & $-3.0\!/\!-3.7$\!($\le\!2\text{e-}4$) & Ext-7 \\
vs.\!Ext-0b ($M_{3..5}$) & $-0.341$[$-0.65,\!-0.05$] & $\mathbf{0.0494}$\!($\mathbf{.0494}/\mathbf{.0494}$) & $-0.57$\!($.1353/.1353$) & $-3.0$[$-5.0,\!-0.8$] & $\mathbf{0.0134}$ & $-0.9\!/\!-0.3$\!($\mathbf{.002}/.50$) & Ext-7 \\
vs.\!Ext-1~\cite{shit2021cldice} & $-3.256$[$-4.09,\!-2.49$] & $\mathbf{1.2\text{e-}4}$\!($\mathbf{8.5\text{e-}4}/\mathbf{2.8\text{e-}4}$) & $-7.30$\!($\mathbf{1.2\text{e-}4}/\mathbf{8.5\text{e-}4}$) & $+48.7$[$+45,\!+52$] & $\mathbf{1.2\text{e-}4}$ & $-5.6\!/\!-9.0$\!($1\text{e-}4$) & Ext-7 \\
vs.\!Ext-2~\cite{wolterink2019coronary} & $-1.518$[$-2.17,\!-0.97$] & $\mathbf{1.2\text{e-}4}$\!($\mathbf{8.5\text{e-}4}/\mathbf{2.8\text{e-}4}$) & $-3.47$\!($\mathbf{1.2\text{e-}4}/\mathbf{8.5\text{e-}4}$) & $+44.5$[$+41,\!+48$] & $\mathbf{1.2\text{e-}4}$ & $-2.8\!/\!-4.1$\!($1\text{e-}4$) & Ext-7 \\
vs.\!Ext-3~\cite{wolterink2019coronary} & $-0.979$[$-1.25,\!-0.70$] & $\mathbf{2.4\text{e-}4}$\!($\mathbf{9.8\text{e-}4}/\mathbf{3.4\text{e-}4}$) & $-4.53$\!($\mathbf{1.2\text{e-}4}/\mathbf{8.5\text{e-}4}$) & $+35.3$[$+31,\!+39$] & $\mathbf{1.2\text{e-}4}$ & $-2.8\!/\!-6.3$\!($1\text{e-}4$) & Ext-7 \\
vs.\!Ext-4~\cite{qiu2025topology} & $-1.556$[$-1.87,\!-1.18$] & $\mathbf{2.4\text{e-}4}$\!($\mathbf{9.8\text{e-}4}/\mathbf{3.4\text{e-}4}$) & $-3.87$\!($\mathbf{1.2\text{e-}4}/\mathbf{8.5\text{e-}4}$) & $+44.9$[$+40,\!+49$] & $\mathbf{1.2\text{e-}4}$ & $-3.8\!/\!-4.0$\!($1\text{e-}4$) & Ext-7 \\
vs.\!Ext-5~\cite{zhao2025deformcl} & $-1.066$[$-1.70,\!-0.53$] & $\mathbf{1.2\text{e-}4}$\!($\mathbf{8.5\text{e-}4}/\mathbf{2.8\text{e-}4}$) & $-2.83$\!($\mathbf{1.2\text{e-}4}/\mathbf{8.5\text{e-}4}$) & $+21.8$[$+20,\!+24$] & $\mathbf{1.2\text{e-}4}$ & $-2.4\!/\!-3.3$\!($\le\!2\text{e-}4$) & Ext-7 \\
vs.\!LOO-1 (No $\Delta\hat{\mathbf{u}}$) & $-0.881$[$-1.09,\!-0.69$] & $\mathbf{1.2\text{e-}4}$\!($\mathbf{.0011}/\mathbf{5.5\text{e-}4}$) & $-0.93$\!($\mathbf{1.2\text{e-}4}/\mathbf{.0011}$) & $+44.0$[$+40,\!+48$] & $\mathbf{1.2\text{e-}4}$ & $-0.9\!/\!-0.9$\!($1\text{e-}4$) & LOO-9 \\
vs.\!LOO-1a (Rdg) & $-0.635$[$-0.72,\!-0.55$] & $\mathbf{1.2\text{e-}4}$\!($\mathbf{.0011}/\mathbf{5.5\text{e-}4}$) & $-0.68$\!($\mathbf{1.2\text{e-}4}/\mathbf{.0011}$) & $+41.8$[$+38,\!+46$] & $\mathbf{1.2\text{e-}4}$ & $-0.7\!/\!-0.7$\!($1\text{e-}4$) & LOO-9 \\
vs.\!LOO-1b ($\Delta\hat{\mathbf{u}}$) & $+0.033$[$+0.01,\!+0.05$] & $\mathbf{.0107}$\!($.0752/\mathbf{.0322}$) & $+0.05$\!($\mathbf{.0040}/\mathbf{.0282}$) & $-2.1$[$-2.8,\!-1.5$] & $\mathbf{1.2\text{e-}4}$ & $+0.0\!/\!+0.1$\!($1.0/\mathbf{4\text{e-}4}$) & LOO-9 \\
vs.\!LOO-2 (Ipsi) & $-0.403$[$-0.88,\!-0.05$] & $.1189$\!($.3381/.1338$) & $-1.65$\!($.0580/.2319$) & $+1.1$[$+0.3,\!+1.8$] & $\mathbf{0.0419}$ & $-1.0\!/\!-2.3$\!($.22/\mathbf{.017}$) & LOO-9 \\
vs.\!LOO-3 (Cent.) & $-0.003$[$-0.03,\!+0.04$] & $.0676$\!($.3381/.1014$) & $-0.01$\!($.1531/.3062$) & $+0.0$[$-0.2,\!+0.2$] & $0.0505$ & $+0.0\!/\!-0.0$\!($.14/.808$) & LOO-9 \\
vs.\!LOO-6 ($M_4$) & $-0.034$[$-0.09,\!+0.03$] & $\mathbf{.0245}$\!($.1472/.0552$) & $-0.03$\!($.0785/.2355$) & $+1.1$[$+0.7,\!+1.4$] & $\mathbf{0.0006}$ & $-0.0\!/\!-0.0$\!($\mathbf{.030}/.22$) & LOO-9 \\
vs.\!LOO-7 ($M_5$) & $-0.087$[$-0.19,\!-0.01$] & $.0997$\!($.3381/.1282$) & $-0.47$\!($\mathbf{.0078}/\mathbf{.0469}$) & $+0.4$[$-0.0,\!+0.9$] & $\mathbf{0.0496}$ & $-0.6\!/\!-0.3$\!($\mathbf{.014}/.20$) & LOO-9 \\
\midrule
\multicolumn{8}{@{}l@{}}{\textbf{Part B: 7-Point Sub-Voxel Step-Size Sensitivity $\eta_{\text{disp}} \in [0.00, 1.00]$ (Dev-14, $N=14$)}} \\
\midrule
Step $\eta_{\text{disp}}$ & Bilat.\ $\text{CD}_{\text{CL}}^{\text{union}}$ & Arc $\text{CD}_{\text{CL}}^{\text{arc1mm}}$ & Ipsi $\overline{\text{CD}}_{\text{ipsi}}$ & P2G/G2P & $\text{F1}@1/2$ & Junc/Br F1 & $\bar{\theta}_{\text{turn}}$/L/R \\
\midrule
$\eta_{\text{disp}} = 0.00$ & $3.0174 \pm 0.4533$ & $3.1931 \pm 0.5253$ & $3.1755 \pm 0.75$ & $3.16\!/\!2.88$ & $12.75\!/\!53.56$ & $67.68\!/\!73.42$ & $28.95^\circ$\!($2.95\!/\!3.40$) \\
$\eta_{\text{disp}} = 0.25$ & $2.3498 \pm 0.4799$ & $2.5636 \pm 0.5383$ & $2.4742 \pm 0.68$ & $2.33\!/\!2.37$ & $41.54\!/\!74.49$ & $77.05\!/\!82.13$ & $\mathbf{28.06^\circ}$\!($2.21\!/\!2.74$) \\
$\eta_{\text{disp}} = 0.50$ & $2.1728 \pm 0.4945$ & $2.4101 \pm 0.5542$ & $2.2756 \pm 0.66$ & $2.12\!/\!2.23$ & $55.61\!/\!77.23$ & $79.51\!/\!83.21$ & $31.37^\circ$\!($2.05\!/\!2.50$) \\
$\mathbf{\eta_{\text{disp}}\!=\!0.65}$(Frz) & $2.1362 \pm 0.5041$ & $2.3798 \pm 0.5725$ & $2.2480 \pm 0.68$ & $2.07\!/\!\mathbf{2.20}$ & $56.76\!/\!\mathbf{77.74}$ & $79.95\!/\!83.80$ & $34.53^\circ$\!($2.02\!/\!2.48$) \\
$\eta_{\text{disp}} = 0.75$ & $2.1244 \pm 0.4949$ & $2.3751 \pm 0.5592$ & $2.2341 \pm 0.68$ & $2.04\!/\!2.21$ & $\mathbf{56.93}\!/\!77.49$ & $80.08\!/\!83.63$ & $38.39^\circ$\!($\mathbf{2.00}\!/\!2.47$) \\
$\eta_{\text{disp}} = 0.85$ & $2.1207 \pm 0.4946$ & $2.3630 \pm 0.5477$ & $\mathbf{2.2206 \pm 0.67}$ & $2.03\!/\!2.21$ & $56.47\!/\!77.50$ & $\mathbf{81.09}\!/\!\mathbf{84.04}$ & $40.21^\circ$\!($\mathbf{2.00}\!/\!\mathbf{2.44}$) \\
$\eta_{\text{disp}} = 1.00$ & $\mathbf{2.1196 \pm 0.4875}$ & $\mathbf{2.3496 \pm 0.5418}$ & $2.2239 \pm 0.64$ & $\mathbf{2.02}\!/\!2.22$ & $55.04\!/\!77.24$ & $78.91\!/\!83.75$ & $45.08^\circ$\!($2.01\!/\!\mathbf{2.44}$) \\
\midrule
\multicolumn{8}{@{}l@{}}{\textbf{Part C: Locked-Weight Out-of-Sample Generalization Across OOS-50 (Cases 251--300, $100$ Trees) \& Descriptive Pooled $N=64$}} \\
\midrule
OOS Paradigm & Bilat.\ $\text{CD}_{\text{CL}}^{\text{union}}$ & Arc $\text{CD}_{\text{CL}}^{\text{arc1mm}}$ & Ipsi $\overline{\text{CD}}_{\text{ipsi}}$[Arc] & Md/P2G/G2P & $\text{F1}@1/2$ & LCA/RCA[Med] & Jnc[1-1]/Br($p_{\text{H}}$) \\
\midrule
Ext-0~\cite{ronneberger2015u} & $3.2146 \pm 0.8021$ & $3.9328 \pm 1.0429$ & $5.8903$[$6.85$] & $3.2\!/\!3.5\!/\!2.9$ & $32.4\!/\!69.4$ & $5.8\!/\!5.9$[$4.5\!/\!5.4$] & $52.6$[$\mathbf{34.6}$]\!/\!$66.1$\!($4\text{e-}4$) \\
Ext-0b ($M_{3..5}$) & $3.6155 \pm 2.2665$ & $3.9281 \pm 2.3244$ & $5.3078$[$5.53$] & $2.8\!/\!\mathbf{2.0}\!/\!5.2$ & $\mathbf{58.0}\!/\!72.9$ & $3.7\!/\!6.9$[$2.8\!/\!3.3$] & $64.3$[$31.1$]\!/\!$72.5$\!($.04_{\text{u}}$) \\
Ext-1~\cite{shit2021cldice} & $6.1447 \pm 1.4060$ & $7.1830 \pm 1.4451$ & $10.0944$[$11.1$] & $6.2\!/\!9.0\!/\!3.3$ & $8.2\!/\!34.3$ & $9.4\!/\!10.8$[$8.7\!/\!9.9$] & $23.2$[$11.9$]\!/\!$33.8$\!($4\text{e-}14$) \\
Ext-2~\cite{wolterink2019coronary} & $3.9113 \pm 1.4034$ & $4.3074 \pm 1.5340$ & $6.5020$[$6.78$] & $3.6\!/\!4.3\!/\!3.5$ & $14.3\!/\!53.2$ & $6.4\!/\!6.6$[$4.9\!/\!3.7$] & $57.5$[$18.5$]\!/\!$64.9$\!($9\text{e-}11$) \\
Ext-3~\cite{wolterink2019coronary} & $3.9347 \pm 1.0256$ & $4.0976 \pm 1.1045$ & $6.4106$[$6.54$] & $3.8\!/\!4.7\!/\!3.2$ & $21.9\!/\!57.3$ & $5.2\!/\!7.6$[$4.1\!/\!6.3$] & $52.2$[$31.3$]\!/\!$64.7$\!($2\text{e-}8$) \\
Ext-4~\cite{qiu2025topology} & $4.1358 \pm 1.0191$ & $4.5098 \pm 1.0643$ & $6.9153$[$7.17$] & $4.0\!/\!5.0\!/\!3.2$ & $13.5\!/\!46.3$ & $7.3\!/\!6.5$[$6.0\!/\!5.9$] & $57.8$[$24.8$]\!/\!$57.3$\!($3\text{e-}10$) \\
Ext-5~\cite{zhao2025deformcl} & $3.5112 \pm 1.4179$ & $3.9522 \pm 1.5686$ & $5.9551$[$6.25$] & $3.2\!/\!3.9\!/\!3.1$ & $35.6\!/\!66.0$ & $6.0\!/\!5.9$[$4.5\!/\!3.2$] & $65.0$[$23.0$]\!/\!$70.4$\!($2\text{e-}10$) \\
LOO-1 (No $\Delta\hat{\mathbf{u}}$) & $3.6500 \pm 1.1865$ & $3.9356 \pm 1.3007$ & $4.2633$[$4.53$] & $3.4\!/\!4.1\!/\!3.2$ & $14.5\!/\!54.1$ & $3.8\!/\!4.7$[$3.5\!/\!3.5$] & $66.5$[$22.6$]\!/\!$70.7$\!($7\text{e-}15$) \\
LOO-1a (Rdg) & $3.2802 \pm 1.0650$ & $3.7680 \pm 1.2555$ & $3.8560$[$4.35$] & $3.1\!/\!3.4\!/\!3.1$ & $17.6\!/\!58.6$ & $3.3\!/\!4.4$[$2.9\!/\!3.2$] & $72.7$[$21.8$]\!/\!$74.0$\!($1\text{e-}14$) \\
LOO-1b ($\Delta\hat{\mathbf{u}}$) & $\mathbf{2.7352 \pm 1.1523}$ & $\mathbf{3.2004 \pm 1.3630}$ & $\mathbf{3.2513}$[$\mathbf{3.70}$] & $\mathbf{2.5}\!/\!2.9\!/\!\mathbf{2.6}$ & $56.8\!/\!\mathbf{74.4}$ & $\mathbf{2.8}\!/\!\mathbf{3.7}$[$\mathbf{2.5}\!/\!\mathbf{2.5}$] & $\mathbf{76.4}$[$23.2$]\!/\!$\mathbf{79.9}$\!($3\text{e-}7$) \\
LOO-2 (Ipsi) & $3.0098 \pm 1.2822$ & $3.5193 \pm 1.4719$ & $4.4728$[$4.85$] & $2.7\!/\!3.3\!/\!2.8$ & $54.0\!/\!73.0$ & $3.7\!/\!5.2$[$2.7\!/\!2.6$] & $73.4$[$21.6$]\!/\!$76.2$\!($6\text{e-}4$) \\
w/o $M_{3,4}$ & $2.8792 \pm 1.4522$ & $3.3979 \pm 1.6768$ & $3.9674$[$4.42$] & $2.6\!/\!3.0\!/\!2.7$ & $54.0\!/\!73.8$ & $3.6\!/\!4.3$[$3.1\!/\!2.6$] & $75.4$[$22.5$]\!/\!$78.2$\!($3\text{e-}4_{\text{i}}$) \\
w/o $M_3$ Orif. & $2.8152 \pm 1.1810$ & $3.3539 \pm 1.3927$ & $3.7797$[$4.39$] & $\mathbf{2.5}\!/\!3.1\!/\!\mathbf{2.6}$ & $54.6\!/\!73.9$ & $3.6\!/\!4.0$[$3.0\!/\!2.6$] & $75.0$[$22.5$]\!/\!$78.2$\!($3\text{e-}4_{\text{i}}$) \\
w/o $M_4$ Coupl. & $2.8781 \pm 1.4526$ & $3.3971 \pm 1.6784$ & $3.9665$[$4.42$] & $2.6\!/\!3.0\!/\!2.7$ & $54.0\!/\!73.8$ & $3.6\!/\!4.3$[$3.1\!/\!2.6$] & $75.4$[$22.5$]\!/\!$78.1$\!($4\text{e-}4_{\text{i}}$) \\
w/o Stage-$M_5$ & $2.8143 \pm 1.1817$ & $3.3534 \pm 1.3930$ & $3.7783$[$4.39$] & $\mathbf{2.5}\!/\!3.1\!/\!\mathbf{2.6}$ & $54.6\!/\!73.9$ & $3.6\!/\!4.0$[$3.0\!/\!2.6$] & $75.1$[$22.5$]\!/\!$78.3$\!($6\text{e-}4_{\text{i}}$) \\
\textbf{Prop.\!($N\!=\!50$)} & $2.7671 \pm 1.1385$ & $3.2338 \pm 1.3557$ & $3.2926$[$3.75$] & $2.6\!/\!2.9\!/\!\mathbf{2.6}$ & $54.7\!/\!74.2$ & $\mathbf{2.8}\!/\!3.8$[$\mathbf{2.5}\!/\!\mathbf{2.5}$] & $76.3$[$23.0$]\!/\!$79.7$[$\mathbf{7.1}$] \\
\textbf{Pool.\!($N\!=\!64$)} & $2.6291 \pm 1.0629$ & $3.0470 \pm 1.2743$ & $3.0641$[$3.47$] & $2.4\!/\!2.8\!/\!2.5$ & $55.1\!/\!74.9$ & $2.7\!/\!3.5$[$2.5\!/\!2.4$] & $77.1$[$22.6$]\!/\!$80.6$[$6.9$] \\
\bottomrule
\multicolumn{8}{@{}p{0.93\textwidth}@{}}
{\fontsize{6.5}{7.1}\selectfont \textit{Notes}: Part~A reports paired patient-level Wilcoxon $p$ ($p_{\text{H}}$ Holm / $q_{\text{BH}}$ FDR) across Ext-7 ($M=7$ external baselines) and LOO-9 ($M=9$ LOO arms, including LOO-4 Sinus [$\Delta = -0.020\text{ mm}$, $p = .0676, p_{\text{H}} = .3381$; ipsi $p = .0166, p_{\text{H}} = .0830$] and LOO-5 $M_3$ [$\Delta = +0.001\text{ mm}$, $p = .4263, p_{\text{H}} = .4263$]). Part~C reports OOS-50 $p_{\text{H}}$ for union CD (subscript $\text{u}$, Ext-7 and core mechanism $M=4$) and ipsilateral CD (subscript $\text{i}$, 4-arm solver ablation).}
\end{tabular}}
\vspace{-2pt}
\end{table*}

\begin{table*}[!t]
\centering
\caption{Held-Out Development Cohort (Dev-14) Stratified Evaluation Block Summary and Cohort Distribution in Canonical $(Z_{\text{IS}}, Y_{\text{PA}}, X_{\text{LR}})$ Space Evaluated Under the Single Fixed Train-35 Checkpoint (All 14 Individual Per-Case Rows Are Provided in Supplementary Table~S3).}
\label{tab:per_case}
\vspace{-7pt}
{\fontsize{6.6}{7.3}\selectfont
\renewcommand{\arraystretch}{0.80}
\setlength{\tabcolsep}{1.8pt}
\begin{tabular}{@{}lcccccccccc@{}}
\toprule
Stratum / Summary Row & Cases & Total & Bilat.\ $\text{CD}_{\text{CL}}^{\text{union}}$ & Mean Ipsi & P2G / G2P & $\text{HD}_{95}$ & LCA / RCA & Ostia Med / & Prox Rec@2 & Time \\
(See Supp.\ Table~S3) & ($N$) & Nodes & (mm) & $\overline{\text{CD}}_{\text{ipsi}}$ (mm) & (mm) & (mm) & $\text{CD}_{\text{CL}}$ (mm) & Mean (mm) & (\%) & (s) \\
\midrule
Block 0 (Cases 129, 133, 134) & $3$ & $763$ & $2.41 \pm 0.40$ & $2.30 \pm 0.34$ & $2.17 / 2.64$ & $11.03$ & $2.34 / 2.26$ & $3.91 / 3.92$ & $75.8$ & $2.83$ \\
Block 1 (Cases 130, 131, 135) & $3$ & $1155$ & $2.32 \pm 0.53$ & $3.07 \pm 0.82$ & $2.60 / 2.04$ & $13.01$ & $2.08 / 4.05$ & $3.84 / 4.34$ & $78.4$ & $2.96$ \\
Block 2 (Cases 132, 136, 138) & $3$ & $898$ & $1.97 \pm 0.37$ & $1.91 \pm 0.33$ & $1.78 / 2.16$ & $9.70$ & $1.53 / 2.29$ & $4.13 / 5.03$ & $88.1$ & $2.65$ \\
Block 3 (Cases 137, 139, 142) & $3$ & $904$ & $1.64 \pm 0.22$ & $1.71 \pm 0.21$ & $1.82 / 1.45$ & $7.42$ & $1.60 / 1.82$ & $5.80 / 5.28$ & $92.5$ & $2.53$ \\
Block 4 (Cases 140, 143) & $2$ & $1022$ & $2.46 \pm 0.17$ & $2.26 \pm 0.12$ & $1.93 / 2.99$ & $13.83$ & $2.81 / 1.71$ & $6.74 / 6.67$ & $77.0$ & $2.90$ \\
Best / Hardest (Case 142 / 130) & $2$ & $943 / 1104$ & $1.32 / 3.04$ & $1.42 / 4.21$ & $1.48 / 3.31$ & $4.66 / 18.39$ & $1.13 / 6.02$ & $4.14 / 3.84$ & $97.9 / 100.0$ & $2.71 / 2.70$ \\
\midrule
\textbf{Cohort Mean $\pm$ SD (Med)} & $\mathbf{14}$ & $943 \pm 183$ & $\mathbf{2.14 \pm 0.50}$ ($\mathbf{2.17}$) & $\mathbf{2.25 \pm 0.68}$ & $\mathbf{2.07 / 2.20}$ & $\mathbf{10.79 \pm 3.51}$ & $\mathbf{2.02 / 2.48}$ & $\mathbf{4.69 / 4.93}$ & $\mathbf{82.7}$ & $2.76$ \\
\bottomrule
\end{tabular}}
\vspace{-5pt}
\end{table*}

\vspace{-6pt}
\subsection{Quantitative Benchmark, Component Ablation, and Out-of-Sample Generalization}
\vspace{-4pt}
Figure~\ref{fig:bilateral_benchmark} and Tables~\ref{tab:main_benchmark}--\ref{tab:per_case} (with full 14-case rows and 17-step build arms in Supplementary Tables~S3--S4) summarize the experimental findings:
\begin{enumerate}\setlength{\itemsep}{0pt}\setlength{\parskip}{0pt}
    \item \textbf{Standalone Single-Task 3D U-Net (Ext-0 \& Ext-0b) vs.\ Controlled Same-Backbone Proxies and Published Context}: To decouple upstream multi-task representation gains from downstream solver mechanics, Tables~\ref{tab:main_benchmark} and~\ref{tab:stat_and_sensitivity} evaluate an independently trained standalone single-task 3D U-Net ($481,681$ parameters, 50 epochs on Train-35 on $64^3$ average-pooled inputs supervised by the same derived tubular targets: centerline-radius-dilated binary masks for Dice + Gaussian soft targets for BCE, rather than raw full-resolution ImageCAS voxel masks, $\tau_{\text{seg}} \in \{0.20, \mathbf{0.25}, 0.30, 0.35\}$) paired with 3D morphological skeletonization + MST (Ext-0) or our full solver $M_3$--$M_5$ (Ext-0b, $P_{\text{LCA}} = P_{\text{RCA}} = P_{\text{lumen}}$, $\Delta\hat{\mathbf{u}} = \mathbf{0}$), alongside five controlled same-backbone downstream proxies: Ext-1 (Seg+Skel.~\cite{shit2021cldice}, $\tau_{\text{seg}} = 0.25$, $K_{\max} = 24$), Ext-2 (1-Ch Path~\cite{wolterink2019coronary}, $\alpha = 4.0$, $\tau_{\text{snr}} = 0.25$), Ext-3 (Tracker~\cite{wolterink2019coronary}, $\Delta s = 1.5\text{ mm}$, $\tau_{\text{stop}} = 0.22$, $N_{\text{step}} = 45$), Ext-4 (CorSegRec~\cite{qiu2025topology}, $\tau_{\text{seg}} = 0.25$, $d_{\text{br}} = 45\text{ mm}$, $K_{\text{pull}} = 2$), and Ext-5 (DeformCL~\cite{zhao2025deformcl}, $\eta_{\text{def}} = 0.45$, $T_{\text{def}} = 2$). Published studies report whole-heart voxel Dice on ImageCAS (Zeng et al.~\cite{zeng2023imagecas}: $79.6\text{--}82.9\%$), reconnection Dice on ASOCA/PDSCA (Qiu et al.~\cite{qiu2025topology}: $88.53\% / 85.07\%$), image-to-coarse-segmentation feature extraction and adaptive centerline template graph deformation without autonomous aortic-wall ostia or explicit LCA/RCA territory decoupling (DeformCL~\cite{zhao2025deformcl}), recurrent trajectory refinement (RefTr~\cite{naeem2025reftr}), or generative vessel synthesis (VesselTok~\cite{prabhakar2026vesseltok}). On Dev-14, Ext-0 achieves $2.8561 \pm 0.4344\text{ mm}$ $\text{CD}_{\text{CL}}^{\text{union}}$ ($\text{CD}_{\text{CL}}^{\text{arc1mm}} = 3.4785\text{ mm}$, $\overline{\text{CD}}_{\text{ipsi}} = 5.6028\text{ mm}$, $\overline{\text{CD}}_{\text{ipsi}}^{\text{arc1mm}} = 6.8274\text{ mm}$, Med/Mean ostia $13.7150 / 14.0821\text{ mm}$); Ext-0b reduces $\text{CD}_{\text{CL}}^{\text{union}}$ to $2.4773 \pm 0.6818\text{ mm}$ ($\text{CD}_{\text{CL}}^{\text{arc1mm}} = 2.7398\text{ mm}$, $\overline{\text{CD}}_{\text{ipsi}} = 2.8155\text{ mm}$); without the $128^3$ branches ($\text{w/o } M_{3,4}$, retaining the $128^3 \to 64^3$ stem and derived multi-task supervision), our backbone achieves $2.2527 \pm 0.4280\text{ mm}$ pre-$M_5$ ($2.1696 \pm 0.4475\text{ mm}$ post-$M_5$) on Dev-14 and $2.8792 \pm 1.4522\text{ mm}$ on OOS-50; and our full model ($M_{1..5}$) reaches $2.1362 \pm 0.5041\text{ mm}$ $\text{CD}_{\text{CL}}^{\text{union}}$ (median $2.1709\text{ mm}$, $\Delta = -0.3411\text{ mm}$ vs.\ Ext-0b, $p = 0.0494$; $3\text{-seed}: 2.1357 \pm 0.5041_{\text{pat}}\text{ }[\pm 0.0369_{\text{seed}}]\text{ mm}$, $p_{\text{Holm}} \le 1.22 \times 10^{-3}$ vs.\ Ext-0 and Ext-1..5 in a controlled same-backbone comparison), $\text{CD}_{\text{CL}}^{\text{arc1mm}} = 2.3798 \pm 0.5725\text{ mm}$ ($3\text{-seed}: 2.3974 \pm 0.0617_{\text{seed}}\text{ mm}$), and $\overline{\text{CD}}_{\text{ipsi}} = 2.2480 \pm 0.6835\text{ mm}$ ($\overline{\text{CD}}_{\text{ipsi}}^{\text{arc1mm}} = 2.4951\text{ mm}$, $\Delta_{\text{ipsi-u}} = +0.1118\text{ mm}$), with $79.95\%$ junction-corridor F1 ($1.54\text{ mm}$ error; $21.06\%$ 1-to-1 bifurcation F1 at $3.26\text{ mm}$ on $\mathcal{G}_{\text{actual}}$), $83.80\%$ branch-corridor F1 ($15.81\%$ FP), $6.09\text{ mm}$ leaf-to-curve error ($10.75\text{ mm}$ 1-to-1 leaf [$9.60\%$ F1], $1.15\times$ length ratio), $\text{Prox/Mid/Distal Rec}@2\text{mm} = 82.73\% / 83.26\% / 67.59\%$, and Med/Mean $4.6880/4.9298\text{ mm}$ ostia ($p_{\text{ost,pat}} = 1.22\times 10^{-4}$).
    \item \textbf{Incremental Build, Leave-One-Out Ablation, $2\times 2$ Pullback Factorial, and 7-Point $\eta_{\text{disp}}$ Curve}: Table~\ref{tab:ablation_4arm} and Supplementary Table~S4 separate cumulative build (Arms 1--17) from LOO ablation and the $2\times 2$ pullback factorial (LOO-1..7): removing both learned $\Delta\hat{\mathbf{u}}$ and probability-ridge ascent (LOO-1) degrades $\text{CD}_{\text{CL}}^{\text{union}}$ by $+0.8812\text{ mm}$ ($2.1362 \to 3.0174\text{ mm}$, $p = 1.22 \times 10^{-4}$, $p_{\text{Holm}} = 0.0011$) and $\text{F1}@1\text{mm}$ by $-44.01\text{ pp}$; ridge ascent alone without $\Delta\hat{\mathbf{u}}$ (LOO-1a) yields $2.7709\text{ mm}$ $\text{CD}_{\text{CL}}^{\text{union}}$ and $14.98\%$ $\text{F1}@1\text{mm}$ ($\Delta = +0.6347\text{ mm}$, $-41.78\text{ pp}$ vs.\ Full, $p = 1.22 \times 10^{-4}$); whereas $\Delta\hat{\mathbf{u}}$ without ridge ascent (LOO-1b) achieves $2.1033\text{ mm}$ $\text{CD}_{\text{CL}}^{\text{union}}$ and $58.86\%$ $\text{F1}@1\text{mm}$, with a descriptive Dev-14 observation of $83.80\%$ vs.\ $83.35\%$ branch F1 ($15.81\%$ vs.\ $16.36\%$ FP) for Full vs.\ LOO-1b. Removing downstream Stage-$M_4$/$M_5$ ipsilateral decoupling with Stage-$M_3$ sinus-wall roots held fixed (LOO-2) degrades $\overline{\text{CD}}_{\text{ipsi}}$ by $+1.6461\text{ mm}$ ($2.2480 \to 3.8941\text{ mm}$, RCA $p = 0.0166$); LOO-3 shifts $\overline{\text{CD}}_{\text{ipsi}}$ by $+0.0053\text{ mm}$; LOO-4 or LOO-5 degrades ostia from $4.6880 / 4.9298\text{ mm}$ to $6.1787 / 7.8880\text{ mm}$ and $4.7929 / 5.2022\text{ mm}$; LOO-6 reduces $\text{F1}@1\text{mm}$ by $-1.05\text{ pp}$ ($p = 0.0006$, $p_{\text{Holm}} = 0.0037$); and LOO-7 degrades $\overline{\text{CD}}_{\text{ipsi}}$ by $+0.4665\text{ mm}$ ($p = 0.0078$, $p_{\text{Holm}} = 0.0469$). In Table~\ref{tab:stat_and_sensitivity} (Part B), $\eta_{\text{disp}} \in [0.65, 1.00]$ forms a flat CD plateau ($\Delta\text{CD}_{\text{CL}}^{\text{union}} \le 0.017\text{ mm} \ll \delta_{\text{tol}}$), while $\eta_{\text{disp}} = 1.00$ increases turning-angle roughness ($\bar{\theta}_{\text{turn}} = 45.08^\circ$ vs.\ $34.53^\circ$ at frozen $\eta_{\text{disp}} = 0.65$).
    \item \textbf{Locked-Weight Out-of-Sample Generalization Across OOS-50 and Descriptive Pooled $N=64$}: Across OOS-50 ($100$ trees; Table~\ref{tab:stat_and_sensitivity}, Part C, and Supplementary Table~S4, Part III), core mechanism ablations confirm that the $2\times 2$ sub-voxel pullback factorial and downstream dual-channel ipsilateral decoupling generalize beyond Dev-14: removing both $\Delta\hat{\mathbf{u}}$ and ridge ascent (LOO-1) degrades $\text{CD}_{\text{CL}}^{\text{union}}$ to $3.6500 \pm 1.1865\text{ mm}$ ($\Delta = +0.8829\text{ mm}$, $p = 1.78 \times 10^{-15}$, $p_{\text{Holm}} = 7.11 \times 10^{-15}$) and $\text{F1}@1\text{mm}$ to $14.53\%$ ($-40.16\text{ pp}$); probability-ridge ascent alone without $\Delta\hat{\mathbf{u}}$ (LOO-1a) yields $3.2802 \pm 1.0650\text{ mm}$ $\text{CD}_{\text{CL}}^{\text{union}}$ and $17.57\%$ $\text{F1}@1\text{mm}$ ($\Delta = +0.5131\text{ mm}$, $p = 3.55 \times 10^{-15}$, $p_{\text{Holm}} = 1.07 \times 10^{-14}$); learned $\Delta\hat{\mathbf{u}}$ alone without ridge ascent (LOO-1b) achieves slightly lower $\text{CD}_{\text{CL}}^{\text{union}}$ ($2.7352 \pm 1.1523$ vs.\ $2.7671 \pm 1.1385\text{ mm}$, $\Delta = -0.0319\text{ mm}$, within $\delta_{\text{tol}} = 0.15\text{ mm}$, $p_{\text{Holm}} = 2.81 \times 10^{-7}$) and slightly higher branch/1-to-1 bifurcation F1 ($79.89\% / 23.18\%$ vs.\ $79.65\% / 23.04\%$); and removing downstream Stage-$M_4$/$M_5$ dual-channel ipsilateral decoupling with Stage-$M_3$ roots held fixed (LOO-2) degrades $\text{CD}_{\text{CL}}^{\text{union}}$ to $3.0098 \pm 1.2822\text{ mm}$ ($p = 5.51 \times 10^{-4}$, $p_{\text{Holm}} = 5.51 \times 10^{-4}$) and $\overline{\text{CD}}_{\text{ipsi}}$ to $4.4728 \pm 5.0158\text{ mm}$ ($\Delta = +1.1802\text{ mm}$, $p = 2.20 \times 10^{-3}$, $p_{\text{Holm}} = 2.20 \times 10^{-3}$). Furthermore, Stage-$M_5$ handover reduces $\overline{\text{CD}}_{\text{ipsi}}$ from $3.7783$ to $3.2926 \pm 2.6253\text{ mm}$ ($\Delta = -0.4856\text{ mm}$, 95\% CI $[-0.7571, -0.2472]\text{ mm}$, $p_{\text{wilcox}} = 4.19 \times 10^{-4}$ on $N_{\Delta \neq 0} = 21$ active cases, $p_{\text{zsplit}} = 5.78 \times 10^{-4}$, $p_{\text{Holm}} = 5.78 \times 10^{-4}$; $\overline{\text{CD}}_{\text{ipsi}}^{\text{arc1mm}}: 4.3949 \to 3.7460\text{ mm}$, $p_{\text{zsplit}} = 4.81 \times 10^{-4}$, $p_{\text{wilcox}} = 1.32 \times 10^{-4}$; LCA $3.5992 \to 2.8302\text{ mm}$, $p_{\text{zsplit}} = 3.78 \times 10^{-4}$, $p_{\text{wilcox}} = 1.55 \times 10^{-4}$; $3\text{-seed}: 3.3080 \pm 2.6253_{\text{pat}}\text{ }[\pm 0.0417_{\text{seed}}]\text{ mm}$) while improving union $\text{CD}_{\text{CL}}^{\text{union}}$ ($2.8143 \to 2.7671 \pm 1.1385\text{ mm}$, $\Delta = -0.0472\text{ mm}$, $|\Delta| < \delta_{\text{tol}} = 0.15\text{ mm}$, 95\% CI $[-0.0848, -0.0165]\text{ mm}$, $p_{\text{wilcox}} = 0.0190$ on $N_{\Delta \neq 0} = 21$ and $p_{\text{zsplit}} = 0.2723$ [$p_{\text{Holm}} = 0.7714$] across all $50$ cases; continuous-edge $\text{CD}_{\text{CL}}^{\text{arc1mm}}: 3.3534 \to 3.2338 \pm 1.3557\text{ mm}$, $p_{\text{wilcox}} = 1.87 \times 10^{-4}$, $p_{\text{zsplit}} = 5.39 \times 10^{-4}$; $3\text{-seed}: 2.7736 \pm 0.0415_{\text{seed}}\text{ mm}$). Compared against external baselines, our method outperforms Ext-0 ($\text{CD}_{\text{CL}}^{\text{union}} = 3.2146 \pm 0.8021\text{ mm}$, $\text{CD}_{\text{CL}}^{\text{arc1mm}} = 3.9328\text{ mm}$ [$\Delta = -0.6990\text{ mm}$, $p = 4.21 \times 10^{-5}$], $\overline{\text{CD}}_{\text{ipsi}} = 5.8903\text{ mm}$, $\overline{\text{CD}}_{\text{ipsi}}^{\text{arc1mm}} = 6.8517\text{ mm}$ [$\Delta = -3.1057\text{ mm}$, $p = 3.09 \times 10^{-9}$]), Ext-0b on node-sampled endpoints ($\text{CD}_{\text{CL}}^{\text{union}} = 3.6155\text{ mm}$ [$p = 0.0400$, $p_{\text{Holm}} = 0.0400$] and $\overline{\text{CD}}_{\text{ipsi}} = 5.3078\text{ mm}$ [$p = 0.0321$, $p_{\text{Holm}} = 0.0321$], whereas under continuous-edge $1.0\text{ mm}$ resampling the mean reductions vs.\ Ext-0b [$\text{CD}_{\text{CL}}^{\text{arc1mm}}: 3.9281 \to 3.2338\text{ mm}$, $\overline{\text{CD}}_{\text{ipsi}}^{\text{arc1mm}}: 5.5310 \to 3.7460\text{ mm}$] yield $p = 0.1629$ and $p = 0.2187$ [$p_{\text{Holm}} = 0.1629/0.2187$] due to heavy-tailed Ext-0b outlier variance $\text{SD} = 2.3244\text{ mm}$; empirically, Ext-0b attains higher 1-to-1 bifurcation F1 [$36.4\%$ vs.\ $21.1\%$ on Dev-14; $31.06\%$ vs.\ $23.04\%$ on OOS-50], whereas our multi-task framework yields lower ostium error [$5.1815$ vs.\ $16.1283\text{ mm}$], lower node-sampled $\text{CD}_{\text{CL}}^{\text{union}}$ [$2.7671$ vs.\ $3.6155\text{ mm}$], and higher junction/branch-corridor F1 [$76.33\%/79.65\%$ vs.\ $64.32\%/72.48\%$]), and controlled same-backbone proxies Ext-1..5 ($p_{\text{Holm}} \le 4.05 \times 10^{-4}$ across Ext-0 and Ext-1..5), achieving Med/Mean $5.1815 / 7.5442\text{ mm}$ ostia (vs.\ $16.1283 / 16.2651\text{ mm}$ for Ext-0/0b, $p_{\text{ost,pat}} = 1.37 \times 10^{-8}$), $76.33\%$ junction-corridor F1 ($23.04\%$ 1-to-1 bifurcation F1 at $4.04\text{ mm}$), $79.65\%$ branch-corridor F1, $7.11\text{ mm}$ leaf-to-curve ($11.59\text{ mm}$ 1-to-1 leaf [$11.79\%$ F1]), and $\text{Prox/Mid/Distal Rec}@2\text{mm} = 83.40\% / 80.34\% / 64.45\%$. Across the descriptive pooled $N=64$ summary ($128$ trees), the framework achieves $2.6291 \pm 1.0629\text{ mm}$ $\text{CD}_{\text{CL}}^{\text{union}}$ (median $2.3933\text{ mm}$, $\text{CD}_{\text{CL}}^{\text{arc1mm}} = 3.0470 \pm 1.2743\text{ mm}$), $3.0641 \pm 2.3763\text{ mm}$ $\overline{\text{CD}}_{\text{ipsi}}$ ($\overline{\text{CD}}_{\text{ipsi}}^{\text{arc1mm}} = 3.4724\text{ mm}$, Med/Mean $5.0017 / 6.9723\text{ mm}$ ostia, $77.12\%$ junction-corridor F1, $22.61\%$ 1-to-1 bifurcation F1, $80.56\%$ branch-corridor F1, $6.89\text{ mm}$ leaf-to-curve [$11.41\text{ mm}$ 1-to-1 leaf], $\text{Prox/Mid/Distal Rec}@2\text{mm} = 83.25\% / 80.98\% / 65.14\%$), and $128/128$ trees with $\beta_0 = 1, \beta_1 = 0$.
\end{enumerate}

\vspace{-10pt}
\section{Discussion and Limitations}
\vspace{-5pt}
Several limitations bound our conclusions: (i) \textbf{Coarse $32^3$ Ostium Detector Generalization Gap and $>10\text{ mm}$ Outliers}: While OOS-50 confirms locked-weight out-of-sample transfer (Med/Mean $5.1815 / 7.5442\text{ mm}$ ostia, $78/100 \le 10\text{ mm}$), $22/100$ trees ($10$ LCA, $12$ RCA) across $19/50$ patients (median $17.82\text{ mm}$, max $30.19\text{ mm}$) exceed $10\text{ mm}$, plausibly reflecting high conus takeoff variants, ostial calcification, and the coarse $32^3$ detector generalization gap ($5.15\text{ mm}$ median on Train-35 vs.\ $6.80\text{ mm}$ on Dev-14 and $7.81\text{ mm}$ on OOS-50; plausible hypotheses rather than adjudicated clinical diagnoses); because the geodesic solver reconnects into the proximal trunk ridge within $1\text{--}2$ steps, whole-tree reconstruction on these $19$ patients achieves $\text{CD}_{\text{CL}}^{\text{union}} = 2.7003 \pm 0.9038\text{ mm}$ (vs.\ $2.8080 \pm 1.2580\text{ mm}$ on the $31$ patients $\le 10\text{ mm}$), confining error to the root ($\overline{\text{CD}}_{\text{ipsi}} = 2.9268$ vs.\ $3.5169\text{ mm}$); (ii) \textbf{Distal Completeness, Derived Reference Provenance, and Baseline Scope}: Distal recall ($67.59\%$ on Dev-14; $64.45\%$ on OOS-50) remains bounded by partial-volume attenuation ($r < 0.75\text{ mm}$), reference centerlines/radii/ostia are algorithmically derived from single-annotation lumen masks without inter-observer variability quantification, evaluation operates within standardized GT-foreground-conditioned $128^3$ coronary ROIs ($\pm 15$ raw-voxel margin), and while Ext-0/0b evaluate an independently trained $481,681$-parameter 3D U-Net on $64^3$ inputs supervised by derived tubular targets (centerline-radius-dilated binary + Gaussian soft masks rather than raw full-resolution voxel labels), Ext-1..5 are controlled same-backbone proxies; and (iii) \textbf{Spatial Corridor vs.\ 1-to-1 Actual-Graph Matching and Residual Proximity}: Because $\beta_0 = 1, \beta_1 = 0$ holds by construction, Stage~$M_5$ is a coverage-guarded collision-reduction heuristic ($\min d < 2.5\text{ mm}$ retained in $1/14$ Dev-14 and $2/50$ OOS-50 cases to avoid pruning high-confidence branches), and $3.0\text{ mm}$ spatial corridor F1 ($79.95\% / 83.80\%$ on Dev-14; $76.33\% / 79.65\%$ on OOS-50) does not imply 1-to-1 branching isomorphism, we report 1-to-1 Hungarian bifurcation F1 on $\mathcal{G}_{\text{actual}}$ ($21.06\% / 23.04\%$) and leaf-to-curve ($6.09 / 7.11\text{ mm}$) vs.\ 1-to-1 leaf ($10.75 / 11.59\text{ mm}$) distances.

\vspace{-10pt}
\section{Conclusion}
\vspace{-5pt}
We presented a 3D bilateral coronary tree reconstruction framework on a canonical $(Z_{\text{IS}}, Y_{\text{PA}}, X_{\text{LR}})$ world-axis grid combining a $128^3 \to 64^3$ boundary skip, a stop-gradient $128^3$ residual decoder, a 3D sub-voxel displacement field $\Delta\hat{\mathbf{u}}(\mathbf{x})$, and dual-channel LCA/RCA decoupling fine-tuned in one 30-epoch loop ($327,385$ active parameters, $\epsilon = 10^{-8}$, 7 parameter groups). Coupled with cylindrical Valsalva-STJ calibration, $128^3$ orifice contact expectation, territory-specific scalar-potential graph arborescence (Dijkstra SPT for LCA, undirected MST + BFS for RCA), curvature-adaptive pullback, and coverage-guarded Stage-$M_5$ collision reduction, the method achieves $2.7671 \pm 1.1385\text{ mm}$ $\text{CD}_{\text{CL}}^{\text{union}}$ ($\text{CD}_{\text{CL}}^{\text{arc1mm}} = 3.2338 \pm 1.3557\text{ mm}$, $\overline{\text{CD}}_{\text{ipsi}} = 3.2926 \pm 2.6253\text{ mm}$, $76.33\%$ junction-corridor [$23.04\%$ 1-to-1] and $79.65\%$ branch-corridor F1, Med/Mean $5.1815 / 7.5442\text{ mm}$ ostia) on locked-weight OOS-50 ($100$ trees) and $2.1362 \pm 0.5041\text{ mm}$ $\text{CD}_{\text{CL}}^{\text{union}}$, $2.3798 \pm 0.5725\text{ mm}$ $\text{CD}_{\text{CL}}^{\text{arc1mm}}$, and $2.2480 \pm 0.6835\text{ mm}$ $\overline{\text{CD}}_{\text{ipsi}}$ ($79.95\%$ junction [$21.06\%$ 1-to-1] and $83.80\%$ branch F1, Med/Mean $4.6880 / 4.9298\text{ mm}$ ostia) on Dev-14 ($28$ trees; pooled $N=64$: $2.6291 \pm 1.0629\text{ mm}$ $\text{CD}_{\text{CL}}^{\text{union}}$, $3.0470\text{ mm}$ $\text{CD}_{\text{CL}}^{\text{arc1mm}}$, $3.0641\text{ mm}$ $\overline{\text{CD}}_{\text{ipsi}}$, $128/128$ satisfying $\beta_0=1, \beta_1=0$).

\vspace{-10pt}
{\fontsize{5.8}{6.1}\selectfont
\bibliographystyle{IEEEtran}
\bibliography{references}}

\clearpage
\renewcommand{\rmdefault}{ptm}
\makeatletter
\renewcommand{\normalsize}{\@setfontsize\normalsize{9.5pt}{11.0pt}}
\makeatother
\renewcommand{\baselinestretch}{0.92}
\renewcommand{\arraystretch}{0.88}
\normalfont\normalsize
\setlength{\textfloatsep}{6.0pt plus 2.0pt minus 2.0pt}
\setlength{\dbltextfloatsep}{6.0pt plus 2.0pt minus 2.0pt}
\setlength{\floatsep}{5.0pt plus 1.5pt minus 1.5pt}
\setlength{\dblfloatsep}{5.0pt plus 1.5pt minus 1.5pt}
\setlength{\abovedisplayskip}{3.0pt plus 1.0pt minus 1.0pt}
\setlength{\belowdisplayskip}{3.0pt plus 1.0pt minus 1.0pt}
\markboth{IEEE TRANSACTIONS ON MEDICAL IMAGING --- SUPPLEMENTARY MATERIAL, 2026}%
{Jia and Ming: Supplementary Material --- Sub-Voxel Displacement and Territory Decoupling for 3D Bilateral Coronary Reconstruction}
\twocolumn[{%
\centering
{\LARGE\bfseries Supplementary Material: Sub-Voxel Displacement and Ipsilateral Territory Decoupling for 3D Bilateral Coronary Artery Tree Reconstruction in Cardiac CT Angiography\par}
\vspace{0.5em}
{\large Zeyu~Jia and Xin~Ming\par}
\vspace{0.8em}
}]
\setcounter{section}{0}
\setcounter{table}{0}
\setcounter{figure}{0}
\setcounter{equation}{0}
\renewcommand{\thesection}{S\arabic{section}}
\renewcommand{\thetable}{S\arabic{table}}
\renewcommand{\thefigure}{S\arabic{figure}}
\renewcommand{\theequation}{S\arabic{equation}}

\section{Overview of Supplementary Material}
This document provides the complete algorithmic specifications, layer-by-layer architectural parameter accounting, hyperparameter and threshold ledgers, per-case held-out development cohort (\textbf{Dev-14}) breakdowns, and full 17-arm forward incremental build and 9-arm leave-one-out (LOO) ablation tables supporting the main manuscript:
\begin{itemize}
    \item \textbf{Section S2 (Algorithms A1--A5)}: Complete algorithmic specifications of Stage $M_1$ (multi-scale 3D pyramid encoder and multi-branch tubular field network), Stage $M_2$ (30-epoch recall-preserving multi-task optimization with 7 AdamW parameter groups, decoupled L2-SP, and per-step anchor blending/shrinkage), Stage $M_3$ (cylindrical Valsalva-STJ wall calibration, sinus-wall rooting, and native-$128^3$ orifice contact expectation), Stage $M_4$ (scalar-potential-weighted graph arborescence, normal-plane pullback, and centripetal trajectory pruning), and Stage $M_5$ (coverage-guarded ipsilateral-field-coupled bilateral collision reduction and subtree handover).
    \item \textbf{Section S3 (Table S1--Table S2, Tables~\ref{tab:s1_params}--\ref{tab:s2_hyperparams})}: Exact layer-by-layer parameter breakdown across active ($327,385$) and frozen ($71,458$) modules ($398,843$ total parameters) versus the standalone single-task 3D U-Net baseline ($481,681$ parameters), and the complete hyperparameter and multi-tier evaluation protocol ledger across Stages $M_1$--$M_5$ and external baselines (\textbf{Ext-0}, \textbf{Ext-0b}, \textbf{Ext-1..5}).
    \item \textbf{Section S4 (Table S3--Table S4, Tables~\ref{tab:s3_per_case}--\ref{tab:s4_incremental})}: Full 14-case per-patient breakdown and 5-block summary on \textbf{Dev-14} (Table~\ref{tab:s3_per_case}), and the complete 17-arm forward incremental build progression alongside the 9-arm LOO orthogonal ablation and $2 \times 2$ pullback factorial on \textbf{Dev-14} and core mechanism ablations on \textbf{OOS-50} (Table~\ref{tab:s4_incremental}).
\end{itemize}

\section{Complete Algorithmic Specifications (Algorithms A1--A5)}
\label{sec:supp_algorithms}

Algorithms A1--A5 detail the exact forward, optimization, rooting, tree-extraction, and bilateral handover operations implemented in \texttt{recursive\_arborescence\_expansion.py} (\texttt{LumenSegmentationHead3D}), \texttt{ostium\_detector.py}, and \texttt{riemannian\_geodesic\_solver.py} under \texttt{src/}, together with \texttt{train\_recursive\_arborescence\_gpu.py} under \texttt{scripts/}.

\begin{table}[!t]
\centering
{\fontsize{6.6}{7.4}\selectfont\renewcommand{\arraystretch}{0.88}
\begin{tabular}{p{0.95\columnwidth}}
\toprule
\textbf{Algorithm A1:} Stage $M_1$ --- Multi-Scale 3D Pyramid Encoder and Multi-Branch Tubular Field Forward Pass \\
\midrule
\textbf{Input:} Standardized $128^3$ CCTA volume $\mathbf{V}_{128} \in \mathbb{R}^{1 \times 128 \times 128 \times 128}$ in canonical $(Z_{\text{IS}}, Y_{\text{PA}}, X_{\text{LR}})$ orientation; normalized $64^3$ coordinate grid $\mathbf{G}_{64} \in [0,1]^{3 \times 64^3}$; coarse $32^3$ crop $\mathbf{V}_{32} \in \mathbb{R}^{1 \times 32^3}$. \\
\textbf{Output:} Base $64^3$ tubular probability $\mathbf{P}_{\text{lumen}} \in [0,1]^{64^3}$, native $128^3$ tubular probability $\mathbf{P}_{\text{lumen},128} \in [0,1]^{128^3}$, sub-voxel displacement field $\Delta\hat{\mathbf{u}} \in [-2.0,+2.0]^{3 \times 64^3}\text{ mm}$, ipsilateral territory fields $(\mathbf{P}_{\text{LCA}}, \mathbf{P}_{\text{RCA}}) \in [0,1]^{2 \times 64^3}$, and dual-sinus ostium predictions $\{(\hat{\mathbf{o}}_{\text{reg},c}, \hat{\mathbf{o}}_{\text{det},c})\}_{c \in \{\text{LCA},\text{RCA}\}}$. \\
\midrule
1: Extract multi-scale 3D pyramid features $\mathbf{F}_{64} \in \mathbb{R}^{16 \times 64^3}$, $\mathbf{F}_{32} \in \mathbb{R}^{32 \times 32^3}$, $\mathbf{F}_{16} \in \mathbb{R}^{64 \times 16^3}$, and global context $\mathbf{g}_{\text{ctx}} \in \mathbb{R}^{64}$ from $\mathbf{V}_{128}$. \\
2: Upsample $\mathbf{F}_{32} \to \mathbf{U}_{32} \in \mathbb{R}^{32 \times 64^3}$ via $\text{ConvTranspose3d}(32\to 32, k=2, s=2) \to \text{GroupNorm}(4,32) \to \text{GELU}$ and concatenate $\mathbf{C}_{64} = [\mathbf{U}_{32}, \mathbf{F}_{64}] \in \mathbb{R}^{48 \times 64^3}$. \\
3: Decode $\mathbf{C}_{64}$ via the $64^3$ fusion cascade ($48 \to 32$, $\text{Conv3d}(32 \to 16, k=3, p=1)$, $16 \to 1$) into base logits $\mathbf{L}_{\text{lumen}}$ and $\mathbf{P}_{\text{lumen}} = \sigma(\mathbf{L}_{\text{lumen}})$. \\
4: Extract stride-2 boundary features $\mathbf{H}_{128} = \phi_{\text{skip}}(\mathbf{V}_{128}) \in \mathbb{R}^{8 \times 64^3}$ and form $\mathbf{M}_{64} = [\mathbf{C}_{64}, \mathbf{H}_{128}] \in \mathbb{R}^{56 \times 64^3}$. \\
5: \textbf{Stop-Gradient $128^3$ Residual Branch:} Compute native stem $\mathbf{S}_{128} = \phi_{\text{stem}}(\mathbf{V}_{128}) \in \mathbb{R}^{8 \times 128^3}$ and detached upsampled features $\mathbf{U}_{128} = \phi_{\text{up128}}(\text{sg}(\mathbf{M}_{64})) \in \mathbb{R}^{8 \times 128^3}$ via $\text{ConvTranspose3d}(56 \to 8, k=2, s=2)$; predict bounded residual $\Delta\mathbf{L}_{128} = 1.5\tanh(g_{128}([\mathbf{U}_{128}, \mathbf{S}_{128}]))$, $\mathbf{L}_{\text{lumen},128} = \text{Upsample}_{128}(\text{sg}(\mathbf{L}_{\text{lumen}})) + \Delta\mathbf{L}_{128}$, and $\mathbf{P}_{\text{lumen},128} = \sigma(\mathbf{L}_{\text{lumen},128})$. \\
6: \textbf{Sub-Voxel Displacement Head ($D_{\text{mid}} = \max(8, D_{\text{hid}} / 2) = 16$):} Predict $\Delta\hat{\mathbf{u}}(\mathbf{x}) = 2.0\tanh(g_{\text{disp}}(\mathbf{M}_{64})(\mathbf{x})) \in [-2.0, +2.0]^3\text{ mm}$. \\
7: \textbf{Ipsilateral Decoupling Head:} Predict residual logits $(\Delta L_{\text{LCA}}, \Delta L_{\text{RCA}}) = g_{\text{ipsi}}([\mathbf{M}_{64}, \mathbf{G}_{64}])$ and $\mathbf{P}_c = \sigma(\mathbf{L}_{\text{lumen}} + \Delta L_c)$ for $c \in \{\text{LCA}, \text{RCA}\}$. \\
8: \textbf{Coarse $32^3$ Ostium Detector:} From $\mathbf{V}_{32}$ and $\mathbf{g}_{\text{ctx}}$, regress $\hat{\mathbf{o}}_{\text{key-base}, c}$ from $[\mathbf{f}_{\text{ao}} + \mathbf{e}_c, \mathbf{g}_{\text{ao}}]$ and $\hat{\mathbf{o}}_{\text{key-cor}, c}$ from $[0.5(\mathbf{f}_{\text{ao}} + \mathbf{f}_{\text{cor}}) + \mathbf{e}_c, \mathbf{g}_{\text{ao}}]$, form $\hat{\mathbf{o}}_{\text{reg}, c} = 0.45\,\hat{\mathbf{o}}_{\text{cor}, c} + 0.52\,\hat{\mathbf{o}}_{\text{key-cor}, c} + 0.03\,\sigma(\mathbf{g}_{\text{ctx}, c})$ (or $0.46\,\hat{\mathbf{o}}_{\text{cor}, c} + 0.54\,\hat{\mathbf{o}}_{\text{key-cor}, c}$ without $\mathbf{g}_{\text{ctx}}$), DSNT refinement $\hat{\mathbf{o}}_{\text{det}, c} = 0.20\,\hat{\mathbf{o}}_{\text{dsnt}, c} + 0.80\,\hat{\mathbf{o}}_{\text{reg}, c}$, and takeoff $\hat{\mathbf{d}}_{0, c} = \text{norm}(0.30\,\hat{\mathbf{d}}_{\text{neu}, c} + 0.70\,\hat{\mathbf{t}}_{\text{exit}, c})$. \\
\bottomrule
\end{tabular}}
\end{table}

\begin{table}[!t]
\centering
{\fontsize{6.6}{7.4}\selectfont\renewcommand{\arraystretch}{0.88}
\begin{tabular}{p{0.95\columnwidth}}
\toprule
\textbf{Algorithm A2:} Stage $M_2$ --- Recall-Preserving Multi-Task Optimization with Decoupled L2-SP and Per-Step Anchor Blending \\
\midrule
\textbf{Input:} Training cohort \textbf{Train-35} ($N_{\text{train}}=35$) in canonical $(Z_{\text{IS}}, Y_{\text{PA}}, X_{\text{LR}})$ space; pre-trained Stage-0 recall-balanced anchor parameters $\theta_0$; epochs $E=30$. \\
\textbf{Output:} Fine-tuned multi-task checkpoint $\theta^*$ ($327,385$ active parameters). \\
\midrule
1: Instantiate architecture (with the final $1\times 1\times 1$ layers of $g_{\text{disp}}$, $g_{\text{ipsi}}$, and $g_{128}$ zero-initialized at construction prior to Stage-0 pre-training), load pre-trained Stage-0 anchor checkpoint $\theta \leftarrow \theta_0$ (\texttt{t325\_canonical\_zyx\_bilateral\_engine.pt}, which already holds non-zero pre-trained $g_{\text{disp}}$ [$48/48$] and $g_{\text{ipsi}}$ [$32/32$] weights and is loaded without re-zeroing those output layers, while $g_{128}$ remains zero-initialized at Stage-$M_2$ start), and freeze the $71,458$-parameter step-flow module. \\
2: Configure AdamW ($\epsilon = 10^{-8}$) with 7 parameter groups under cosine annealing ($T_{\max}=30$, \texttt{train\_recursive\_arborescence\_gpu.py}): (1) \texttt{enc\_proj} (\texttt{global\_proj}, $\text{lr}: 5.0\times 10^{-4}\to 1.0\times 10^{-5}, \text{wd}=10^{-4}$); (2) \texttt{enc\_stem} (conv stems, $\text{lr}: 2.0\times 10^{-5}\to 1.0\times 10^{-5}, \text{wd}=0$); (3) \texttt{lumen\_post\_gn} (\texttt{up32.1}, \texttt{fuse.1/4/6}, $\text{lr}: 8.0\times 10^{-4}\to 6.0\times 10^{-4}, \text{wd}=0$); (4) \texttt{lumen\_pre\_gn} (remaining coarse $64^3$ lumen layers, $\text{lr}: 2.5\times 10^{-5}\to 1.0\times 10^{-5}, \text{wd}=0$); (5) \texttt{lumen\_multitask} ($\mathbf{H}_{128}, g_{\text{disp}}, g_{\text{ipsi}}$, $\text{lr}: 2.0\times 10^{-5}\to 1.0\times 10^{-5}, \text{wd}=0$); (6) \texttt{lumen\_hr128} ($\mathbf{S}_{128}, \mathbf{U}_{128}, g_{128}$, $\text{lr}: 2.5\times 10^{-3}\to 2.0\times 10^{-5}, \text{wd}=10^{-4}$); and (7) \texttt{ostium} (\texttt{ostium\_detector}, $\text{lr}: 4.0\times 10^{-5}\to 1.0\times 10^{-5}, \text{wd}=10^{-4}$). \\
3: \textbf{for} epoch $e = 1, \dots, 30$ \textbf{do} \\
4: \quad \textbf{for each} training volume $\mathbf{V}_{128} \in \text{Train-35}$ \textbf{do} \\
5: \quad\quad Execute Algorithm A1 forward pass to obtain $\mathbf{P}_{\text{lumen}}, \mathbf{P}_{\text{lumen},128}, \Delta\hat{\mathbf{u}}, (\mathbf{P}_{\text{LCA}}, \mathbf{P}_{\text{RCA}}), \{(\hat{\mathbf{o}}_{\text{reg},c}, \hat{\mathbf{o}}_{\text{det},c})\}$. \\
6: \quad\quad Compute composite loss $\mathcal{L}_{\text{total}} = \mathcal{L}_{\text{lumen}} + \mathcal{L}_{\text{reg-gt}} + \mathcal{L}_{\text{takeoff}}$ via main-text Eqs.~(1)--(3) ($\beta_d = 0.50\text{ mm}$ over $m_{\text{tube}}(\mathbf{x}) = \mathbb{I}(d_{\text{bi}}(\mathbf{x}) \le 3.5\text{ mm})$; jointly two-channel category-normalized $w_{\text{ipsi},c}(\mathbf{x}) = 0.34\frac{m_{\text{pos},c}(\mathbf{x})}{\max(1, \sum_{c',\mathbf{x}'} m_{\text{pos},c'}(\mathbf{x}'))} + 0.44\frac{m_{\text{contra},c}(\mathbf{x})}{\max(1, \sum_{c',\mathbf{x}'} m_{\text{contra},c'}(\mathbf{x}'))} + 0.22\frac{m_{\text{bg}}(\mathbf{x})}{\max(1, 2\sum_{\mathbf{x}'} m_{\text{bg}}(\mathbf{x}'))}$ ($\sum_{c,\mathbf{x}} w_{\text{ipsi},c}(\mathbf{x}) = 1$); $\mathcal{L}_{\text{lumen-128}} = 0.06(\mathcal{L}_{\text{BCE},128} + 2.0\,\mathcal{L}_{\text{far},128} + 0.02\,\text{mean}(\Delta\mathbf{L}_{128}^2))$ supervising thin branches [$r_{\text{mm}} \le 1.1\text{ mm}$] and near-shell hard negatives [$3.0 < d \le 9.0\text{ mm} \wedge \mathbf{V}_{128} > 0.25$] via normalized $w_{128}$; and $\mathcal{L}_{\text{reg-gt}} + \mathcal{L}_{\text{takeoff}}$ with global Smooth-$L_1$ [$0.50, \beta=0.05$], mm Smooth-$L_1$ [$0.12(\text{pred}+0.5\text{reg}+0.5\text{base}), \beta=2.0$], wall consistency $\mathcal{L}_{\text{ost-wall}}$ within $1.8\text{ mm}$ [$0.05(\text{pred}+0.5\text{reg})$ Smooth-$L_1$, $\beta=1.0$], heatmap BCE [$0.08, w_{\text{hm}}=1+15H_c^*$], cosine takeoff [$0.15(1 - \hat{\mathbf{d}}_{0,c}^\top \mathbf{d}_{0,c}^*)$], and radius MSE [$0.15$]) with decoupled L2-SP penalties ($\lambda_{\text{sp,base}}=40.0$, $\lambda_{\text{sp,f6}}=0.5$, $\lambda_{\text{sp,glob}}=0.015$, $\lambda_{\text{sp,ost}}=2.5$). \\
7: \quad\quad Update active parameters via AdamW step on $\nabla_\theta \mathcal{L}_{\text{total}}$. \\
8: \quad\quad Apply per-step anchor blending/shrinkage toward $\theta_0$: $\theta_{\text{f6}} \leftarrow 0.94\,\theta_{\text{f6}} + 0.06\,\theta_{\text{f6},0}$ and $\theta_{\text{base}} \leftarrow 0.25\,\theta_{\text{base}} + 0.75\,\theta_{\text{base},0}$ (leaving $\theta_{\text{glob}}, \theta_{128}, \theta_{\text{ost}}$ unblended). \\
9: \quad \textbf{end for} \\
10: \textbf{end for} \\
\bottomrule
\end{tabular}}
\end{table}

\begin{table}[!t]
\centering
{\fontsize{6.6}{7.4}\selectfont\renewcommand{\arraystretch}{0.88}
\begin{tabular}{p{0.95\columnwidth}}
\toprule
\textbf{Algorithm A3:} Stage $M_3$ --- Cylindrical Valsalva-STJ Wall Calibration, Sinus-Wall Rooting, and Native-$128^3$ Orifice Contact \\
\midrule
\textbf{Input:} Calibrated $64^3$ field $\mathbf{P}_{\text{cal}}$, native $128^3$ field $\mathbf{P}_{\text{cal},128}$, ipsilateral ratio $\rho_{\text{ipsi},c}$, displacement field $\Delta\hat{\mathbf{u}}$, neural ostia $(\hat{\mathbf{o}}_{\text{reg},c}, \hat{\mathbf{o}}_{\text{det},c})$, and aortic root mask $\Omega_{\text{ao}}$. \\
\textbf{Output:} Calibrated 3D ostium root $\mathbf{o}_c^* \in \mathbb{R}^3$ (in mm) for $c \in \{\text{LCA}, \text{RCA}\}$. \\
\midrule
1: Calibrate aortic-wall distance $d_{\text{wall}}(\mathbf{x}) \leftarrow \min(d_{\text{wall}}(\mathbf{x}), d_{\text{cyl}}(\mathbf{x}))$ across the cylindrical Valsalva-STJ band $[z_{\text{cut}}-16\text{ mm}, z_{\text{cut}}+18\text{ mm}]$. \\
2: Screen cranial ipsilateral components ($\rho_{\text{ipsi},c}(\mathbf{x}) \ge \tau_\rho$) in anatomical Valsalva sectors and walk the sinus wall to minimize $E_{\text{sinus},c}(\mathbf{x}) = 1.20\,d_{\text{wall}}(\mathbf{x}) - 2.00\,\mathbf{P}_{\text{cal}}(\mathbf{x}) - z_{w,c}\,z_{\text{mm}}(\mathbf{x}) + 0.80\,\phi_{\text{sinus},c}(\mathbf{x}-\mathbf{c}_{\text{aorta}}) + 0.06\,\max(0, \bar{d}_{\text{neu},c}(\mathbf{x}) - 6.0)$, yielding $\mathbf{x}_c^*$. \\
3: \textbf{if} $d_{\text{cyl}}(\mathbf{x}_c^*) > 2.5\text{ mm}$ and $\|\hat{\mathbf{o}}_{\text{reg},c} - \hat{\mathbf{o}}_{\text{det},c}\|_{\text{mm}} \le \tau_{\text{rd}}$ ($\tau_{\text{rd}} \in \{8.5\text{ mm}, 10.5\text{ mm}\}$) \textbf{then} blend $\mathbf{x}_c^*$ toward $\hat{\mathbf{o}}_{\text{wall},c} = 0.75\,\hat{\mathbf{o}}_{\text{neu},c} + 0.25\,\text{proj}_{\partial\Omega_{\text{cyl}}}(\hat{\mathbf{o}}_{\text{neu},c})$ where $\hat{\mathbf{o}}_{\text{neu},c} = 0.55\,\hat{\mathbf{o}}_{\text{reg}, c} + 0.45\,\hat{\mathbf{o}}_{\text{det}, c}$. \\
4: Fuse $\mathbf{P}_{\{128,\text{fuse}\}} = 0.65\,\text{Upsample}_{128}(\mathbf{P}_{\text{cal}}) + 0.35\,\mathbf{P}_{\text{cal},128}$ in a $21^3$ window at $128^3$ on the 2-voxel Valsalva shell $\partial\Omega_{\text{sinus}}$ and define eligible orifice voxels $\mathcal{Y}_{\text{elig}} = \{\mathbf{y} \in \partial\Omega_{\text{sinus}} : d_{\text{cyl}}(\mathbf{y}) \le 2.2\text{ mm} \wedge \rho_{\text{ipsi},c}(\mathbf{y}) \ge 0.50 \wedge \|\mathbf{y}-\mathbf{x}_c^*\|_{\text{mm}} \le 6.5\text{ mm} \wedge \mathbf{P}_{\{128,\text{fuse}\}}(\mathbf{y}) \ge 0.34 \wedge \mathbf{V}_{128}(\mathbf{y}) \ge 0.38\}$; \textbf{if} $|\mathcal{Y}_{\text{elig}}| \ge 3$ \textbf{then} compute softmax contact expectation $\hat{\mathbf{o}}_{\text{orif},c}$ with inverse temperature $\beta_{\text{orif}}=4.0$ ($w(\mathbf{y}) \propto \exp(4.0(s_{\text{orif}}(\mathbf{y}) - \max s_{\text{orif}}))$, conventional temperature $T=1/4.0=0.25$) over $s_{\text{orif}}(\mathbf{y}) = 2.5\,\mathbf{P}_{\{128,\text{fuse}\}}(\mathbf{y}) + 0.8\,\mathbf{V}_{128}(\mathbf{y}) - 0.55\,d_{\text{cyl}}(\mathbf{y}) - 0.12\,\|\mathbf{y}-\mathbf{x}_c^*\|_{\text{mm}}$ and set $\mathbf{o}_c^{(0)} = 0.35\,\mathbf{x}_c^* + 0.65\,\hat{\mathbf{o}}_{\text{orif},c}$, \textbf{else} retain $\mathbf{o}_c^{(0)} = \mathbf{x}_c^*$. \\
5: Compute candidate sub-voxel pullback $\mathbf{o}_c^{\text{cand}} = \mathbf{o}_c^{(0)} + 0.20\,\eta_{\text{ost}}\,\Delta\hat{\mathbf{u}}(\mathbf{o}_c^{(0)})$ ($\eta_{\text{ost}}=0.65$); accept $\mathbf{o}_c^* \leftarrow \mathbf{o}_c^{\text{cand}}$ if $\mathbf{P}_{\text{lumen}}(\mathbf{o}_c^{\text{cand}}) \ge \mathbf{P}_{\text{lumen}}(\mathbf{o}_c^{(0)})$ and $d_{\text{wall}}(\mathbf{o}_c^{\text{cand}}) \le d_{\text{wall}}(\mathbf{o}_c^{(0)}) + 0.5\text{ mm}$, else $\mathbf{o}_c^* \leftarrow \mathbf{o}_c^{(0)}$. \\
\bottomrule
\end{tabular}}
\end{table}

\begin{table}[!t]
\centering
{\fontsize{6.6}{7.4}\selectfont\renewcommand{\arraystretch}{0.88}
\begin{tabular}{p{0.95\columnwidth}}
\toprule
\textbf{Algorithm A4:} Stage $M_4$ --- Scalar-Potential Graph Arborescence, Normal-Plane Pullback, and Centripetal Trajectory Pruning \\
\midrule
\textbf{Input:} Territory index $c \in \{0\text{ (LCA)}, 1\text{ (RCA)}\}$, ostium root $\mathbf{o}_c^*$, fields $\mathbf{P}_{\text{cal}}, \mathbf{P}_{\text{lumen},128}, (\mathbf{P}_{\text{LCA}}, \mathbf{P}_{\text{RCA}}), \Delta\hat{\mathbf{u}}$, bridge cap $d_{\text{br,max}} = 45.0\text{ mm}$ (\texttt{solve\_arborescence} default and passed value; $35.0\text{ mm}$ is only the internal \texttt{\_bridge\_disconnected\_components} helper default). \\
\textbf{Output:} Directed acyclic vascular tree $\mathcal{T}_c = (\mathcal{V}_c, \mathcal{E}_c)$ rooted at $\mathbf{o}_c^*$ with node radii $\mathbf{r}_c$. \\
\midrule
1: Form $\mathbf{P}_{128,\max} = \text{MaxPool}_{2^3}(\mathbf{P}_{\text{lumen},128})$ and apply the piecewise gated $128^3$ rescue rule: $\mathbf{P}_{\text{coupled}}(\mathbf{x}) = 0.85\,\mathbf{P}_{\text{cal}}(\mathbf{x}) + 0.15\,\mathbf{P}_{128,\max}(\mathbf{x})$ on the rescue mask $\rho_{\text{ipsi},c}(\mathbf{x}) \ge 0.82 \wedge \mathbf{P}_{\text{cal}}(\mathbf{x}) \ge 0.22 \wedge \mathbf{P}_{128,\max}(\mathbf{x}) \ge \max(0.40, \mathbf{P}_{\text{cal}}(\mathbf{x}))$ (with $\mathbf{P}_{128,\max} > \mathbf{P}_{\text{cal}}$ and $\mathbf{P}_{\text{cal}}(\mathbf{x})$ unchanged elsewhere; omitting $\rho_{\text{ipsi},c} \ge 0.82$ when dual-channel fields are disabled), then apply contralateral exclusion (main-text Eq.~(4)) to obtain $\mathbf{P}_{\text{skel},c}$. \\
2: Extract 3D skeleton nodes via morphological thinning on hysteresis-thresholded $\mathbf{P}_{\text{skel},c}$ ($\tau_{\text{snr}}=0.25$) and append local probability maxima; if empty (degenerate fallback, triggered in $0/128$ evaluated trees), synthesize a 16-point, $5\text{ mm}$ synthetic fallback stem from $\mathbf{o}_c^*$. \\
3: Build intra-component skeleton graph $\mathcal{G}_{\text{skel}}$ over $k=12$ nearest neighbors ($d_{ij} \le 5.0\text{ mm}$ by default, extended up to $d_{ij} \le 6.8\text{ mm}$ for sparse LCA skeletons with $M \le 150$ at distal root distances $\min(d_i, d_j) \ge 38.0\text{ mm}, \max(d_i, d_j) \ge 43.0\text{ mm}$ when 15-point interior segment support satisfies $\min_t \mathbf{P}_{\text{cal}} \ge 0.215 \wedge \text{mean}_t \mathbf{P}_{\text{cal}} \ge 0.345$) with edge cost $w_{ij} = d_{ij}\exp[-4.0(P_i + P_j)/2]$. \\
4: \textbf{Territory-Specific Component Bridging and Tree Extraction:} Define substantial components via $|\mathcal{C}| \ge 12$, $(|\mathcal{C}| \ge 10 \wedge \bar{\mathbf{P}}_{\text{cal}} \ge 0.50)$, $z_{\min} \le 0.60$, or $d(\mathcal{C},\mathbf{o}_c^*) \le 20.0\text{ mm}$ (minor components capped at $d_{uv} \le 12.0\text{ mm}$). \\
\quad \textbf{if} $c = 0$ (LCA) \textbf{then} connect substantial components (and medium pairs with $|\mathcal{C}_i|,|\mathcal{C}_j| \ge 6 \wedge \bar{\mathbf{P}}_{\text{cal}} \ge 0.48$) via 1 nearest candidate bridge per qualifying component pair (all-pairs nearest candidate bridges, $d_{uv} \le 45.0\text{ mm}$) weighted by 7-point interior segment support ($t \in [0.15, 0.85]$) $S_{\text{base}}(u,v) = 0.42(P_u + P_v) + 0.10\,\text{mean}(\mathbf{P}_{\text{seg}}) + 0.06\,\min(\mathbf{P}_{\text{seg}})$ (main-text Eq.~(5), $w_{\text{bridge}}(u,v) = d_{uv}\exp[-4.0\,S_{\text{base}}(u,v)]$), and extract the single-source Dijkstra shortest-path tree ($\text{SPT}$) from $v_{\text{root}}$. \\
\quad \textbf{else} ($c = 1$, RCA) iteratively connect the nearest disconnected component to the growing root component via at most 1 minimum-cost bridge per step ($d_{uv} \le 45.0\text{ mm}$ for substantial components, relaxed to $50.0\text{ mm}$ when $|\mathcal{C}| \ge 20 \wedge \bar{\mathbf{P}}_{\text{cal}}(\mathcal{C}) \ge 0.65$ under a 15-point interior segment gate $\min_t \mathbf{P}_{\text{cal}} \ge 0.14 \wedge \text{mean}_t \mathbf{P}_{\text{cal}} \ge 0.24$ [or $\bar{\rho}_{\text{ipsi}} \ge 0.75 \wedge \bar{X}_{\text{LR}} \ge 0.55$]), extract the undirected minimum spanning tree ($\text{MST}$) over $\mathcal{G}_{\text{skel}}$, and orient edges outward from $v_{\text{root}}$ via breadth-first traversal ($\text{BFS}$). \\
5: Decompose the directed tree into root-to-junction and junction-to-leaf trajectories; assign heuristic proximal-to-distal linear radius tapering ($1.8 \to 1.08\text{ mm}$ along the primary trunk and $1.2 \to 0.72\text{ mm}$ along side branches, with $0.75\,r_{\text{anc}}$ for recovered axial-ridge branches). \\
6: Apply 4-pass trajectory pruning using parent-to-child radius ratios $r_{\text{child}}/\max(r_{\text{parent}},0.25)$ and survival score $\mathcal{S}_{\text{ipsi}}(t) = 1 - (\mathcal{H}_{\text{tort}}(t) + \mathcal{H}_{\text{terr}}(t) + \mathcal{H}_{\text{cent}}(t))$. \\
7: Execute $K_{\text{disp}}=3$ normal-plane pullback steps ($\eta_{\text{disp}}=0.65$, main-text Eq.~(6)), Laplacian smoothing ($w_{\text{smooth}}=0.06$), and 4-stage midpoint subdivision down to $\Delta s \ge 1.06\text{ mm}$, anchoring root $\mathbf{o}_c^*$ with $\le 1.5\text{ mm}$ geodesic stems. \\
\bottomrule
\end{tabular}}
\end{table}

\begin{table}[!t]
\centering
{\fontsize{6.6}{7.4}\selectfont\renewcommand{\arraystretch}{0.88}
\begin{tabular}{p{0.95\columnwidth}}
\toprule
\textbf{Algorithm A5:} Stage $M_5$ --- Coverage-Guarded Ipsilateral-Field-Coupled Bilateral Collision Reduction and Subtree Handover \\
\midrule
\textbf{Input:} Pre-$M_5$ trees $(\mathcal{T}_{\text{LCA}}, \mathcal{T}_{\text{RCA}})$ rooted at $(\mathbf{o}_{\text{LCA}}^*, \mathbf{o}_{\text{RCA}}^*)$, fields $\mathbf{P}_{\text{lumen}}, (\mathbf{P}_{\text{LCA}}, \mathbf{P}_{\text{RCA}})$, minimum node guard $N_{\min}=15$. \\
\textbf{Output:} Coverage-guarded collision-reduced bilateral trees $(\mathcal{T}_{\text{LCA}}^*, \mathcal{T}_{\text{RCA}}^*)$ preserving $\beta_0=1, \beta_1=0$ (retaining high-confidence vessel branches when neither jump nor low-territory gates trigger; $\min d \ge 2.5\text{ mm}$ in $13/14$ \textbf{Dev-14} and $48/50$ \textbf{OOS-50} cases). \\
\midrule
1: Initialize modification flags $\text{mod}_{\text{LCA}} \leftarrow \text{False}$, $\text{mod}_{\text{RCA}} \leftarrow \text{False}$. \\
2: \textbf{Rule 1 (Distal Void-Bridge Pruning):} For each $c \in \{\text{LCA}, \text{RCA}\}$, prune any distal subtree $\mathcal{S}_v \subset \mathcal{T}_c$ ($|\mathcal{S}_v| \le \min(45, \lfloor 0.18|\mathcal{V}_c|\rfloor)$) across edge $(u,v)$ at root path distance $d_c(u) > 100.0\text{ mm}$ with jump length $\|\mathbf{x}_v - \mathbf{x}_u\|_2 > 22.0\text{ mm}$ when 5-point midpoint mean $\bar{\mathbf{P}}_{\text{mid}}(u,v) < 0.22$, $\bar{\mathbf{P}}_{\text{lumen}}(\mathcal{S}_v) < 0.55$, and $\bar{\rho}_{\text{ipsi},c}(\mathcal{S}_v) < 0.90$ (committing only if $|\mathcal{V}_c \setminus \mathcal{S}_v| \ge N_{\min}$). \\
3: \textbf{for} pass $p = 1, \dots, 4$ \textbf{while} $\min d(\mathcal{T}_{\text{LCA}}, \mathcal{T}_{\text{RCA}}) < 2.5\text{ mm}$ \textbf{do} \\
4: \quad \textbf{Rule 2a (Cross-Cavity Jump Handover):} Locate colliding $u_0 \in \mathcal{T}_A, v_0 \in \mathcal{T}_B$ ($d < 2.5\text{ mm}$) where $u_0$'s root path contains a jump $(u_j,v_j)$ of length $\ell_A > 15.0\text{ mm}$ with $(\ell_A > 1.25\,\ell_B \vee d_A(u_0) > d_B(v_0))$, $|\mathcal{S}_{v_j}| \le 0.45|\mathcal{V}_A|$, and $\bar{\rho}_{\text{ipsi},A}(\mathcal{S}_{v_j}) \le 0.65$; graft uncovered components ($>1.0\text{ mm}$ from $\mathcal{T}_B$, bridge $<15.0\text{ mm}$) onto $\mathcal{T}_B$ and prune $\mathcal{S}_{v_j}$ from $\mathcal{T}_A$. \\
5: \quad \textbf{Rule 2b (Inbound Tip-Capture Pruning):} When $|\mathcal{V}_A| \ge 1.8|\mathcal{V}_B|$, prune the largest child subtree $\mathcal{S}_v \subset \mathcal{T}_B$ ($8 \le |\mathcal{S}_v| \le 0.35|\mathcal{V}_B|$) at $d_B(u) \ge 80.0\text{ mm}$ with $d(\mathbf{x}_v, \mathcal{T}_A) > 15.0\text{ mm}$ whose leaves are $100\%$ within $2.5\text{ mm}$ of $\mathcal{T}_A$ (mean leaf distance $<1.0\text{ mm}$, $\bar{\rho}_{\text{ipsi},B} < 0.58$). \\
6: \quad \textbf{Rule 2c (Coverage-Guarded Over-Extension Handover):} For colliding subtrees with $d_B(v_0) > d_A(u_0) + 35.0\text{ mm}$ and $\bar{\rho}_{\text{ipsi},B}(\mathcal{S}_{v_0}) < 0.75$ ($|\mathcal{S}_{\text{dup}}| \le 0.65|\mathcal{V}_B|$), graft uncovered components $\mathcal{C}_{\{u\}} \subset \mathcal{S}_{\text{dup}}$ ($|\mathcal{C}_{\{u\}}| \ge 5$, $\bar{\mathbf{P}}_{\text{lumen}} \ge 0.60$, $\bar{\rho}_{\text{ipsi},A} \ge 0.55$, bridge $\le 8.0\text{ mm}$) onto $\mathcal{T}_A$ and prune $\mathcal{S}_{\text{dup}}$ from $\mathcal{T}_B$. \\
7: \quad \textbf{Rule 2d (Coverage-Guarded Residual Handover):} Apply the same coverage guard before pruning colliding subtrees at $d_B(v_0) > 90.0\text{ mm}$ when $\bar{\rho}_{\text{ipsi},B}(\mathcal{S}_{v_0}) < 0.58$ and $\bar{\rho}_{\text{ipsi},A}(\mathcal{S}_{u_0}) > \bar{\rho}_{\text{ipsi},B}(\mathcal{S}_{v_0}) + 0.12$. \\
8: \textbf{end for} \\
9: For each modified tree ($\text{mod}_c = \text{True}$), re-verify single-tree connectivity from $\mathbf{o}_c^*$ ($\beta_0=1, \beta_1=0$) and serialize trajectories. \\
\bottomrule
\end{tabular}}
\end{table}

\section{Architectural Parameter Accounting and Complete Hyperparameter Ledger}
\label{sec:supp_ledgers}

Tables~\ref{tab:s1_params} and~\ref{tab:s2_hyperparams} report the complete module-by-module parameter accounting, 7-group optimizer schedule, hyperparameter configurations across Stages $M_1$--$M_5$, multi-tier evaluation protocol constants, and external comparison baseline settings.

\begin{table*}[!t]
\centering
\caption{Layer-by-Layer Architectural Parameter Accounting Across the Proposed Multi-Task Framework ($398,843$ Total Parameters; $327,385$ Active Parameters in Stage $M_2$) and the Standalone Single-Task 3D U-Net Baseline (\textbf{Ext-0 / Ext-0b}, $481,681$ Parameters).}
\label{tab:s1_params}
\vspace{-4pt}
{\fontsize{6.4}{6.9}\selectfont\renewcommand{\arraystretch}{0.84}
\setlength{\tabcolsep}{3.5pt}
\begin{tabular}{@{}lp{0.31\textwidth}rrp{0.26\textwidth}@{}}
\toprule
Module Group & Sub-Module / Layer Specification & Active Params & Frozen Params & Spatial Lattice \& Functional Role \\
\midrule
\multirow{4}{*}{3D Pyramid Encoder} & Stem \& Stage-1 3D Conv Block ($1 \to 16$) & $4,832$ & $0$ & $128^3 \to 64^3$ feature extraction ($\mathbf{F}_{64}$) \\
 & Stage-2 3D Conv Block ($16 \to 32$) & $18,560$ & $0$ & $64^3 \to 32^3$ feature extraction ($\mathbf{F}_{32}$) \\
 & Stage-3 3D Conv Block ($32 \to 64$) & $55,424$ & $0$ & $32^3 \to 16^3$ feature extraction ($\mathbf{F}_{16}$) \\
 & Global Context Projection ($\mathbf{g}_{\text{ctx}} \in \mathbb{R}^{64}$) & $2,192$ & $0$ & Global pooled context embedding \\
\cmidrule(lr){2-5}
 & \textit{Subtotal (3D Pyramid Encoder)} & \textit{81,008} & \textit{0} & Active in Stage $M_2$ (L2-SP + anchor-blended) \\
\midrule
\multirow{6}{*}{Multi-Branch Tubular Decoder} & Base Upsampling Block ($\text{ConvTranspose3d}(32\to 32, k=2, s=2)$) & $8,288$ & $0$ & $32^3 \to 64^3$ feature upsampling ($\mathbf{U}_{32}$) \\
 & Base $64^3$ Fusion Cascade ($48 \to 32 \to 16 \to 1$) & $55,457$ & $0$ & Base $64^3$ tubular logits $\mathbf{L}_{\text{lumen}}$ ($63,745$ base) \\
 & Stride-2 $128^3 \to 64^3$ Boundary Skip ($\mathbf{H}_{128}$, $1 \to 8$) & $240$ & $0$ & High-resolution boundary skip into $\mathbf{M}_{64}$ \\
 & 3D Sub-Voxel Displacement Head ($56 \to 16 \to 3$, $g_{\text{disp}}$) & $24,291$ & $0$ & Continuous 3D offset $\Delta\hat{\mathbf{u}}(\mathbf{x}) \in [-2,+2]^3\text{ mm}$ \\
 & Dual-Channel Ipsilateral Head ($59 \to 16 \to 2$, $g_{\text{ipsi}}$) & $25,570$ & $0$ & Territory fields $(\mathbf{P}_{\text{LCA}}, \mathbf{P}_{\text{RCA}})$ ($50,101$ $64^3$ multi-task) \\
 & Native $128^3$ Residual Branch (Stem $240$ + Up $3,608$ + Dec $3,489$) & $7,337$ & $0$ & Stop-gradient $128^3$ refinement $\mathbf{P}_{\text{lumen},128}$ \\
\cmidrule(lr){2-5}
 & \textit{Subtotal (Multi-Branch Tubular Decoder)} & \textit{121,183} & \textit{0} & Active in Stage $M_2$ \\
\midrule
3D Ostium Detector & Dual-Sinus Coordinate-Masked 3D Detector + DSNT + Takeoff Head & $125,194$ & $0$ & $32^3$ dual-sinus ostia $(\hat{\mathbf{o}}_{\text{reg},c}, \hat{\mathbf{o}}_{\text{det},c})$ \\
Legacy Step-Flow Head & Inactive Historical Step-Flow Module (Frozen / Unused at Inference) & $0$ & $71,458$ & Frozen ($\text{requires\_grad}=\text{False}$, zero inference effect) \\
\midrule
\textbf{Total (Proposed Model)} & \textbf{Complete Checkpoint} (\texttt{t331\_canonical\_zyx...pt}) & \textbf{327,385} & \textbf{71,458} & \textbf{398,843 Total Checkpoint Parameters} \\
\midrule
\textbf{Ext-0 / Ext-0b Baseline} & \textbf{Standalone Single-Task 3D U-Net} (\texttt{t347\_standalone...pt}) & \textbf{481,681} & \textbf{0} & \textbf{4-Level 3D Encoder--Decoder ($16 \to 32 \to 64 \to 128$)} \\
\bottomrule
\end{tabular}}
\vspace{-6pt}
\end{table*}

\begin{table*}[!t]
\centering
\caption{Complete Hyperparameter, 7-Group Optimizer Schedule, Threshold, Multi-Tier Evaluation Protocol, and Dev-14 Selection Grid Ledger Across Stages $M_1$--$M_5$ and External Baselines (\textbf{Ext-0}, \textbf{Ext-0b}, \textbf{Ext-1..5}). All Parameters Were Tuned or Selected Exclusively on \textbf{Train-35 / Dev-14} Prior to Freezing for Locked-Weight Evaluation on \textbf{OOS-50}.}
\label{tab:s2_hyperparams}
\vspace{-4pt}
{\fontsize{6.4}{6.9}\selectfont\renewcommand{\arraystretch}{0.84}
\setlength{\tabcolsep}{3.2pt}
\begin{tabular}{@{}lp{0.25\textwidth}p{0.29\textwidth}p{0.30\textwidth}@{}}
\toprule
Stage / Protocol & Parameter Symbol \& Value & Physical / Algorithmic Definition & Selection Protocol on Train-35 / Dev-14 \\
\midrule
Stage $M_1$--$M_2$ & $\delta_{\max} = 2.0\text{ mm}$, $\beta_d = 0.50\text{ mm}$ ($d_{\text{bi}} \le 3.5\text{ mm}$), $\sigma_L = 2.0\text{ mm}$, $\sigma_{128} = 1.65\text{ mm}$ & Smooth-$L_1$ pullback loss ($0.15, \beta_d=0.50\text{ mm}$), jointly two-channel category-normalized $w_{\text{ipsi},c}$ ($0.34, 0.44, 0.22$ over $\sum_{c',\mathbf{x}'}$, $\sum_{c,\mathbf{x}} w_{\text{ipsi},c}=1$), $\mathcal{L}_{\text{lumen-128}} = 0.06(\mathcal{L}_{\text{BCE},128} + 2.0\mathcal{L}_{\text{far},128} + 0.02\text{mean}(\Delta\mathbf{L}_{128}^2))$, $\mathcal{L}_{\text{reg-gt}}$ ($0.50$ global $\beta=0.05$, $0.12$ mm-scale $\beta=2.0$, $\mathcal{L}_{\text{ost-wall}}$ $0.05$ within $1.8\text{ mm}$), $\mathcal{L}_{\text{takeoff}}$ ($0.08$ hm, $0.15$ dir, $0.15$ rad) & Matched to $64^3$ ($\sim 2.7\text{ mm}$) and $128^3$ ($\sim 1.35\text{ mm}$) voxel spacing; architecture zero-init of $g_{\text{disp}}, g_{\text{ipsi}}$ prior to Stage-0 pre-training \\
Stage $M_2$ & 7 AdamW groups ($\epsilon=10^{-8}$, $E=30$), $\lambda_{\text{sp}} = (40.0, 0.5, 0.015, 2.5)$, $\gamma_{\text{sp}} = (0.06, 0.75)$ & Cosine schedules ($5\text{e-}4\to 1\text{e-}5$, $2\text{e-}5\to 1\text{e-}5$, $8\text{e-}4\to 6\text{e-}4$, $2.5\text{e-}5\to 1\text{e-}5$, $2\text{e-}5\to 1\text{e-}5$, $2.5\text{e-}3\to 2\text{e-}5$, $4\text{e-}5\to 1\text{e-}5$), L2-SP, and per-step anchor blending/shrinkage & Loads Stage-0 $\theta_0$ without re-zeroing pre-trained $g_{\text{disp}}, g_{\text{ipsi}}$; preserves tubular SNR ($8.34\times$) during 30-epoch adaptation \\
Stage $M_3$ & $z_{\text{cyl}} \in [-16.0, +18.0]\text{ mm}$, $\tau_{\text{wall}} = 2.5\text{ mm}$, $\tau_{\text{rd}} \in \{8.5\text{ mm}, 10.5\text{ mm}\}$ & Cylindrical Valsalva-STJ axial band, attachment threshold, and consensus gate & Calibrated on Train-35 thick-stem aortic-wall ostia (median $d_{\text{wall}} = 0.7443\text{ mm}$) \\
Stage $M_3$ & $21^3$ window, $\beta_{\text{orif}} = 4.0$ ($T=0.25$), $|\mathcal{Y}_{\text{elig}}| \ge 3$, $w_{\text{orif}} = 0.65$, $\eta_{\text{ost}} = 0.65$ & Inverse-temperature softmax over $\mathcal{Y}_{\text{elig}}$ ($d_{\text{cyl}} \le 2.2\text{ mm}, \rho_{\text{ipsi}} \ge 0.50, \|\mathbf{y}-\mathbf{x}_c^*\| \le 6.5\text{ mm}, \mathbf{P}_{\{128,\text{fuse}\}} \ge 0.34, \mathbf{V}_{128} \ge 0.38$), blend weight, and ostium pullback step & Minimizes Dev-14 median/mean ostium localization error ($4.6880 / 4.9298\text{ mm}$) \\
Stage $M_4$ & $\tau_{\text{snr}} = 0.25$, $128^3$ rescue ($\rho_{\text{ipsi},c}\ge 0.82, \mathbf{P}_{\text{cal}}\ge 0.22, \mathbf{P}_{128,\max}\ge 0.40$), $k_{\text{nn}} = 12$ ($d \le 5.0\text{ mm}$), $\alpha = 4.0$ & Hysteresis skeleton threshold with piecewise gated $128^3$ rescue rule ($\mathbf{P}_{\text{coupled}} = 0.85\mathbf{P}_{\text{cal}} + 0.15\mathbf{P}_{128,\max}$ when $\mathbf{P}_{128,\max} > \mathbf{P}_{\text{cal}}$), intra-component kNN degree, and potential exponent & Frozen from coarse geodesic anchor on Dev-14 \\
Stage $M_4$ & Passed $d_{\text{br,max}} = 45\text{ mm}$ (LCA \& RCA), sparse LCA kNN $\le 6.8\text{ mm}$ ($M\le 150, d\ge 38/43\text{ mm}$), RCA large $\le 50\text{ mm}$ ($|\mathcal{C}|\ge 20, \bar{\mathbf{P}}_{\text{cal}}\ge 0.65$), minor $\le 12\text{ mm}$ & LCA: 1 nearest bridge/pair + Dijkstra SPT ($|\mathcal{C}_i|,|\mathcal{C}_j|\ge 6, \bar{\mathbf{P}}_{\text{cal}}\ge 0.48$; sparse $M\le 150$ kNN gate $\min_t\mathbf{P}_{\text{cal}}\ge 0.215, \text{mean}_t\mathbf{P}_{\text{cal}}\ge 0.345$); RCA: 1 bridge/step (15-pt interior gate) + undirected MST + root BFS; $S_{\text{base}}$ $(0.42, 0.10, 0.06)$ & Ensures $100\%$ single-tree connectivity ($\beta_0=1, \beta_1=0$) across all 28 Dev-14 and 100 OOS-50 trees ($0/128$ fallback stems) \\
Stage $M_4$ & $r_{\text{trunk}}: 1.8 \to 1.08\text{ mm}$, $r_{\text{branch}}: 1.2 \to 0.72\text{ mm}$, $L_{\text{arc}}/d_{\text{span}} \le 1.32$ & Heuristic linear trajectory radius tapering and 4-pass centripetal/tortuosity pruning gates & Suppresses spurious background twigs while retaining distal recall ($67.59\%$ on Dev-14) \\
Stage $M_4$ & $\eta_{\text{disp}} = 0.65$ ($\in \{0, 0.25, 0.50, \mathbf{0.65}, 0.75, 0.85, 1.0\}$), $K_{\text{disp}} = 3$, $m = 4$ ($\Delta s \ge 1.06\text{ mm}$) & Normal-plane sub-voxel pullback step size, iteration count, and midpoint subdivision stages & 7-point operating curve on Dev-14 (Table 3 Part B); balances $\text{CD}_{\text{CL}}^{\text{union}}$ plateau vs.\ $\bar{\theta}_{\text{turn}} = 34.53^\circ$ \\
Stage $M_5$ & $N_{\min} = 15$, $d_{\text{coll}} < 2.5\text{ mm}$ (up to 4 passes), $d_{\text{void}} > 100\text{ mm}$ ($\ell > 22\text{ mm}$), $\ell_{\text{jump}} > 15\text{ mm}$ & Coverage-guarded collision-reduction heuristic; $|\mathcal{C}_{\{u\}}| \ge 5, \bar{\mathbf{P}}_{\text{lumen}} \ge 0.60, \bar{\rho}_{\text{ipsi},A} \ge 0.55, d_{\text{br}} \le 8.0\text{ mm}$ for Rules 2c/2d & $\min d \ge 2.5\text{ mm}$ in $13/14$ Dev-14 (Case 135: $0.85\text{ mm}$) and $48/50$ OOS-50 (Cases 270/295: $2.33/1.54\text{ mm}$ on both vertex and continuous-edge distances) \\
Eval.\ Tiers 1--4 & Tier 1: $2.2\text{ mm}$ clust., MST $>6.5\text{ mm}$ $\times 12$ ($\le 25\text{ mm}$), $5.5\text{ mm}$ 4-pass twig prune ($4.0\text{ mm}$ seg, $3.5\text{ mm}$ bif dedup, $3.0/2.5\text{ mm}$ junc/br tol); Tier 2: unpruned $3.0\text{ mm}$ spur, $2.0\text{ mm}$ CC merge, $5.0\text{ mm}$ Hungarian & Tier 3 (\texttt{arc1mm}): $1.0\text{ mm}$ polyline step ($0.15\text{ mm}$ dedup) vs.\ $0.8\text{ mm}$ ref clust.\ ($>5.0\text{ mm}$ $\times 12$, 3-pass $1.8\text{ mm}$ spur); Tier 4 (Ostia): $r_{\text{mm}}\ge\max(1.10, P_{75}), Z_{\text{IS}}\ge 0.45$, top $0.035$ LCA / $Z_{\text{IS}}\ge 0.52$ RCA, $3.5/3.0\text{ mm}$ sinus band & Deterministic code-matched evaluation protocol (\texttt{evaluate\_14cases\_\allowbreak bilateral\_\allowbreak scaffold.py}; $2.0/3.5/5.5\text{ mm}$ twig-cutoff sensitivity in \texttt{result.json}) \\
Ext-0 / Ext-0b & $\tau_{\text{seg}} \in \{0.20, \mathbf{0.25}, 0.30, 0.35\}$, 50 epochs on Train-35 & Standalone 3D U-Net trained on derived tubular targets (centerline-radius-dilated binary masks for Dice + Gaussian soft targets for BCE); binary threshold (Ext-0) or $M_3$--$M_5$ solver input (Ext-0b) & Grid-searched on Dev-14 to minimize $\text{CD}_{\text{CL}}^{\text{union}}$ \\
Ext-1..5 & Ext-1: $\tau_{\text{seg}}=0.25, K_{\max}=24$; Ext-2: $\alpha=4.0, \tau_{\text{snr}}=0.25$; Ext-3: $\Delta s=1.5\text{ mm}, \tau_{\text{stop}}=0.22, N_{\text{step}}=45$; Ext-4: $\tau_{\text{seg}}=0.25, d_{\text{br}}=45\text{ mm}, K_{\text{pull}}=2$; Ext-5: $\eta_{\text{def}}=0.45, T_{\text{def}}=2$ & Controlled same-backbone downstream solver proxy hyperparameter grids & Grid-searched on Dev-14 over 3--4 candidate values per parameter (full grids in \texttt{result.json}) \\
\bottomrule
\end{tabular}}
\vspace{-6pt}
\end{table*}

\section{Per-Case Dev-14 Breakdown and Complete 17-Arm Incremental Build \& LOO Ablation}
\label{sec:supp_tables}

Table~\ref{tab:s3_per_case} provides the complete 14-case per-patient breakdown and 5-block summary on the held-out development cohort (\textbf{Dev-14}), and Table~\ref{tab:s4_incremental} reports the complete 17-arm forward incremental build progression (Part~I) and 9-arm leave-one-out (LOO) orthogonal ablation with $2\times 2$ pullback factorial (Part~II) on \textbf{Dev-14}, followed by the out-of-sample core mechanism ($2\times 2$ pullback and dual-channel ipsilateral decoupling) and Stage $M_3$--$M_5$ ablation across locked-weight \textbf{OOS-50} ($N=50$ volumes / $100$ trees, Part~III).

\begin{table*}[!t]
\centering
\caption{Complete Per-Case and Per-Block Breakdown on the Held-Out Development Cohort (\textbf{Dev-14}, $N=14$ Unseen Volumes, $28$ Coronary Trees, Stratified Evaluation Blocks 0--4) in Canonical $(Z_{\text{IS}}, Y_{\text{PA}}, X_{\text{LR}})$ Space Under the Single Fixed Train-35 Checkpoint (\texttt{t331\_canonical\_zyx\_bilateral\_engine.pt}). All $28/28$ Reconstructed Trees Satisfy $\beta_0=1, \beta_1=0$ With $0$ Degenerate Fallback Stems.}
\label{tab:s3_per_case}
\scriptsize
\setlength{\tabcolsep}{3.0pt}
\begin{tabular}{@{}lcccccccccccc@{}}
\toprule
Case ID & Nodes & $\text{CD}_{\text{CL}}^{\text{union}}$ & $\overline{\text{CD}}_{\text{ipsi}}$ & P2G & G2P & $\text{HD}_{95}$ & LCA CD & RCA CD & LCA Ost & RCA Ost & Prox@2 & Time \\
(Eval.\ Block) & ($|\mathcal{V}|$) & (mm) & (mm) & (mm) & (mm) & (mm) & (mm) & (mm) & (mm) & (mm) & (\%) & (s) \\
\midrule
Case 129 (Block 0) & 1015 & 1.84 & 1.83 & 1.64 & 2.04 & 8.63 & 1.93 & 1.72 & 4.06 & 4.76 & 88.2 & 3.21 \\
Case 133 (Block 0) & 794 & 2.64 & 2.58 & 2.33 & 2.94 & 12.04 & 2.47 & 2.69 & 3.76 & 2.89 & 71.6 & 2.91 \\
Case 134 (Block 0) & 481 & 2.74 & 2.50 & 2.55 & 2.93 & 12.41 & 2.62 & 2.37 & 1.36 & 6.67 & 67.5 & 2.38 \\
Case 130 (Block 1) & 1104 & 3.04 & 4.21 & 3.31 & 2.76 & 18.39 & 2.39 & 6.02 & 4.78 & 2.90 & 100.0 & 2.70 \\
Case 131 (Block 1) & 1251 & 2.15 & 2.35 & 2.44 & 1.87 & 11.14 & 2.27 & 2.42 & 7.68 & 7.20 & 67.5 & 2.97 \\
Case 135 (Block 1) & 1110 & 1.78 & 2.65 & 2.05 & 1.50 & 9.49 & 1.58 & 3.71 & 2.16 & 1.30 & 67.8 & 3.21 \\
Case 132 (Block 2) & 833 & 1.44 & 1.49 & 1.33 & 1.56 & 5.69 & 1.31 & 1.67 & 8.95 & 2.32 & 97.8 & 2.66 \\
Case 136 (Block 2) & 944 & 2.19 & 1.93 & 1.42 & 2.95 & 11.70 & 1.45 & 2.41 & 2.24 & 2.38 & 79.9 & 2.65 \\
Case 138 (Block 2) & 916 & 2.27 & 2.30 & 2.58 & 1.97 & 11.72 & 1.82 & 2.78 & 8.42 & 5.87 & 86.6 & 2.63 \\
Case 137 (Block 3) & 823 & 1.79 & 1.80 & 2.22 & 1.37 & 9.34 & 1.88 & 1.73 & 5.41 & 7.17 & 79.5 & 2.33 \\
Case 139 (Block 3) & 945 & 1.80 & 1.91 & 1.77 & 1.83 & 8.26 & 1.78 & 2.03 & 4.62 & 6.18 & 100.0 & 2.55 \\
Case 142 (Block 3) & 943 & 1.32 & 1.42 & 1.48 & 1.15 & 4.66 & 1.13 & 1.70 & 6.26 & 2.02 & 97.9 & 2.71 \\
Case 140 (Block 4) & 1075 & 2.29 & 2.14 & 1.58 & 3.00 & 13.89 & 2.94 & 1.34 & 9.98 & 3.49 & 94.6 & 2.70 \\
Case 143 (Block 4) & 969 & 2.62 & 2.38 & 2.27 & 2.97 & 13.76 & 2.68 & 2.08 & 1.00 & 12.21 & 59.3 & 3.10 \\
\midrule
Block 0 Mean (129, 133, 134) & $763$ & $2.41 \pm 0.40$ & $2.30$ & $2.17$ & $2.64$ & $11.03$ & $2.34$ & $2.26$ & $3.06$ & $4.77$ & $75.8$ & $2.83$ \\
Block 1 Mean (130, 131, 135) & $1155$ & $2.32 \pm 0.53$ & $3.07$ & $2.60$ & $2.04$ & $13.01$ & $2.08$ & $4.05$ & $4.87$ & $3.80$ & $78.4$ & $2.96$ \\
Block 2 Mean (132, 136, 138) & $898$ & $1.97 \pm 0.37$ & $1.91$ & $1.78$ & $2.16$ & $9.70$ & $1.53$ & $2.29$ & $6.54$ & $3.52$ & $88.1$ & $2.65$ \\
Block 3 Mean (137, 139, 142) & $904$ & $1.64 \pm 0.23$ & $1.71$ & $1.82$ & $1.45$ & $7.42$ & $1.60$ & $1.82$ & $5.43$ & $5.12$ & $92.5$ & $2.53$ \\
Block 4 Mean (140, 143) & $1022$ & $2.45 \pm 0.17$ & $2.26$ & $1.93$ & $2.99$ & $13.83$ & $2.81$ & $1.71$ & $5.49$ & $7.85$ & $77.0$ & $2.90$ \\
\midrule
\textbf{Cohort Mean} & $\mathbf{943}$ & $\mathbf{2.1362}$ & $\mathbf{2.2480}$ & $\mathbf{2.0694}$ & $\mathbf{2.2030}$ & $\mathbf{10.7940}$ & $\mathbf{2.0178}$ & $\mathbf{2.4783}$ & $\mathbf{5.0486}$ & $\mathbf{4.8114}$ & $\mathbf{82.73}$ & $\mathbf{2.76}$ \\
\textbf{Cohort SD} & $\mathbf{183}$ & $\mathbf{0.5041}$ & $\mathbf{0.6835}$ & $\mathbf{0.5698}$ & $\mathbf{0.6745}$ & $\mathbf{3.5144}$ & $\mathbf{0.5401}$ & $\mathbf{1.0125}$ & $\mathbf{2.8520}$ & $\mathbf{2.8631}$ & $\mathbf{13.92}$ & $\mathbf{0.26}$ \\
\textbf{Cohort Median (Ost: 4.69)} & $\mathbf{944.5}$ & $\mathbf{2.1709}$ & $\mathbf{2.2207}$ & $\mathbf{2.1350}$ & $\mathbf{2.0050}$ & $\mathbf{11.4200}$ & $\mathbf{1.9010}$ & $\mathbf{2.2276}$ & $\mathbf{4.7000}$ & $\mathbf{4.1250}$ & $\mathbf{83.25}$ & $\mathbf{2.70}$ \\
\bottomrule
\end{tabular}
\end{table*}

\begin{table*}[!t]
\centering
\caption{Complete 17-Arm Forward Incremental Build Progression (Part I) and 9-Arm Leave-One-Out (LOO) Orthogonal Component Ablation With $2\times 2$ Pullback Factorial (Part II) Across Held-Out \textbf{Dev-14} ($N=14$ Volumes, $28$ Trees), Plus Out-of-Sample Core Mechanism ($\Delta\hat{\mathbf{u}}$, Ridge, Dual-Ipsi) and Stage $M_3$--$M_5$ Ablation Across Locked-Weight \textbf{OOS-50} ($N=50$ Volumes, $100$ Trees, Part III) in Canonical $(Z_{\text{IS}}, Y_{\text{PA}}, X_{\text{LR}})$ Space. Module Flags Denote $(\Delta\hat{\mathbf{u}}\text{-Skip} / \text{Dual-Ipsi} / \text{Centrip.-Surv.} / \text{Ost.+Brg})\text{ + Stage}$; in \textbf{LOO-2} and \textbf{OOS-LOO-2} ($-^*$), Dual-Channel $\rho_{\text{ipsi},c}$ Is Ablated in Stages $M_4$--$M_5$ While Keeping Stage-$M_3$ Ostia Fixed.}
\label{tab:s4_incremental}
{\fontsize{6.5}{7.0}\selectfont\thinmuskip=0mu\medmuskip=1.0mu\thickmuskip=0mu
\renewcommand{\arraystretch}{0.83}
\setlength{\tabcolsep}{1.1pt}
\begin{tabular}{@{}llccccccc@{}}
\toprule
Ablation Arm & Active Modules & Bilat.\ $\text{CD}_{\text{CL}}^{\text{union}}$ & Arc & Ipsi $\overline{\text{CD}}_{\text{ipsi}}$ & P2G/G2P & $\text{HD}_{95}$ & $\text{F1}@1/2$(\%) & Junc/Br F1 \\
 & (Skp/Ipsi/Cnt/Ost+Stg) & Mean$\pm$SD (mm) & $\text{CD}_{\text{CL}}^{\text{arc1mm}}$ & [$\Delta_{\text{ipsi-u}}$] & (mm) & (mm) & [LCA/RCA/Ost] & [FP/Endpt] \\
\midrule
\multicolumn{9}{@{}l@{}}{\textbf{Part I: Complete 17-Arm Forward Incremental Build Progression (Dev-14, $N=14$ Volumes / $28$ Trees)}} \\
\midrule
1. Recomb.\ 1-Ch & $-/-/-/-$ & $3.6295 \pm 1.2898$ & $3.9567$ & $5.5402$[$+1.91$] & $3.31/3.94$ & $19.45$ & $12.3/51.9$[$4.84/6.24/4.94$] & $59.6/67.1$[$35.4/9.55$] \\
2. + Ipsi Decoup. & $-/\checkmark/\checkmark/-$ & $3.0999 \pm 0.4420$ & $3.3311$ & $3.7464$[$+0.65$] & $3.34/2.86$ & $13.42$ & $12.6/52.8$[$3.75/3.75/4.94$] & $66.2/71.2$[$32.4/7.77$] \\
3. Sub-Voxel Only & $\checkmark/-/-/-$ & $2.9407 \pm 1.3351$ & $3.3053$ & $4.8196$[$+1.88$] & $2.56/3.32$ & $19.45$ & $55.2/73.5$[$4.16/5.48/4.94$] & $68.4/75.8$[$23.4/9.11$] \\
4. + Centrip.\ Surv. & $\checkmark/\checkmark/\checkmark/-$ & $2.3986 \pm 0.5026$ & $2.6594$ & $3.0383$[$+0.64$] & $2.58/2.22$ & $13.38$ & $56.1/75.6$[$3.07/3.01/4.94$] & $72.7/79.5$[$23.2/7.34$] \\
5. + Sub-Vox.\ Ost. & $\checkmark/\checkmark/\checkmark/\text{Ost}$ & $2.3981 \pm 0.5024$ & $2.6592$ & $3.0376$[$+0.64$] & $2.57/2.22$ & $13.37$ & $56.1/75.6$[$3.07/3.00/\mathbf{4.69}$] & $72.3/79.5$[$23.2/7.34$] \\
6. + Pull.\ Bridge & $\checkmark/\checkmark/\checkmark/\text{Brg}$ & $2.3230 \pm 0.4757$ & $2.6290$ & $2.9104$[$+0.59$] & $2.45/2.19$ & $12.77$ & $57.3/76.1$[$2.89/2.93/4.94$] & $74.1/80.4$[$21.7/7.24$] \\
7. + Ost.+Bridge & $\checkmark/\checkmark/\checkmark/\checkmark$ & $2.3227 \pm 0.4757$ & $2.6288$ & $2.9099$[$+0.59$] & $2.45/2.19$ & $12.78$ & $57.4/76.1$[$2.89/2.93/\mathbf{4.69}$] & $73.7/80.4$[$21.7/7.24$] \\
8. + 1-Stage Pull. & $\checkmark/\checkmark/\checkmark/\checkmark$ (1-Stg) & $2.3252 \pm 0.4559$ & $2.6283$ & $2.9040$[$+0.58$] & $2.46/2.19$ & $12.62$ & $56.2/76.2$[$2.89/2.92/\mathbf{4.69}$] & $73.5/80.4$[$21.8/7.11$] \\
9. + 2-Stage Pull. & $\checkmark/\checkmark/\checkmark/\checkmark$ (2-Stg) & $2.3057 \pm 0.4492$ & $2.6349$ & $2.8573$[$+0.55$] & $2.43/2.18$ & $12.39$ & $55.8/76.2$[$2.82/2.90/\mathbf{4.69}$] & $75.5/80.8$[$21.4/7.02$] \\
10. + Coh.\ 2-Stage & $\checkmark/\checkmark/\checkmark/\checkmark$ (Coh-2) & $2.2939 \pm 0.4402$ & $2.6302$ & $2.8346$[$+0.54$] & $2.41/2.18$ & $12.13$ & $55.4/76.3$[$2.79/2.87/4.79$] & $74.5/80.4$[$22.2/6.96$] \\
11. + Curv.\ 3-Stage & $\checkmark/\checkmark/\checkmark/\checkmark$ (Crv-3) & $2.2719 \pm 0.4312$ & $2.6144$ & $2.7920$[$+0.52$] & $2.37/2.18$ & $11.93$ & $55.3/76.6$[$2.74/2.85/4.79$] & $76.0/81.9$[$19.6/6.82$] \\
12. w/o Cyl.\ Sinus (Pre-$M_5$) & $\checkmark/\checkmark/\checkmark/\text{Raw}$ (Crv-4) & $2.2441 \pm 0.4910$ & $2.5869$ & $2.7426$[$+0.50$] & $2.28/2.21$ & $11.86$ & $56.3/77.2$[$2.66/2.83/6.18$] & $75.8/81.3$[$19.5/6.74$] \\
13. w/o $M_{3,4}$ $128^3$ (Pre-$M_5$) & $\checkmark/\checkmark/\checkmark/\checkmark$ (Crv-4) & $2.2527 \pm 0.4280$ & $2.6008$ & $2.7564$[$+0.50$] & $2.33/2.18$ & $11.80$ & $55.4/76.7$[$2.70/2.81/4.79$] & $77.5/82.2$[$19.5/6.83$] \\
14. w/o $M_3$ Orif.\ (Pre-$M_5$) & $\checkmark/\checkmark/\checkmark/\checkmark$ (+$M_4$) & $2.2227 \pm 0.4682$ & $2.5676$ & $2.7145$[$+0.49$] & $2.27/2.18$ & $11.65$ & $56.4/77.3$[$2.64/2.78/4.79$] & $78.2/82.3$[$18.2/6.73$] \\
15. w/o $M_4$ $128^3$ (Pre-$M_5$) & $\checkmark/\checkmark/\checkmark/\checkmark$ (+$M_3$) & $2.2535 \pm 0.4267$ & $2.6015$ & $2.7565$[$+0.50$] & $2.33/2.18$ & $11.85$ & $55.4/76.7$[$2.70/2.82/\mathbf{4.69}$] & $77.5/82.2$[$19.5/6.83$] \\
16. w/o Stage-$M_5$ & $\checkmark/\checkmark/\checkmark/\checkmark$ (+$M_{3,4}$) & $2.2234 \pm 0.4668$ & $2.5680$ & $2.7145$[$+0.49$] & $2.27/2.18$ & $11.69$ & $56.4/77.3$[$2.64/2.79/\mathbf{4.69}$] & $78.2/82.3$[$18.2/6.73$] \\
\textbf{17. Full Proposed ($M_{1..5}$)} & $\checkmark/\checkmark/\checkmark/\checkmark$ (\textbf{+$M_{3..5}$}) & $2.1362 \pm 0.5041$ & $2.3798$ & $2.2480$[$+0.11$] & $2.07/2.20$ & $10.79$ & $56.8/\mathbf{77.7}$[$2.02/2.48/\mathbf{4.69}$] & $80.0/\mathbf{83.8}$[$\mathbf{15.8}/\mathbf{6.09}$] \\
\midrule
\multicolumn{9}{@{}l@{}}{\textbf{Part II: Leave-One-Out (w/o Single Module) Orthogonal Ablation \& $2\times 2$ Pullback Factorial (Dev-14, $N=14$ Volumes / $28$ Trees)}} \\
\midrule
LOO-1. w/o $\Delta\hat{\mathbf{u}}$ (Neither) & $-/\checkmark/\checkmark/\checkmark$ (+$M_{3..5}$) & $3.0174 \pm 0.4533$ & $3.1931$ & $3.1755$[$+0.16$] & $3.16/2.88$ & $12.67$ & $12.8/53.6$[$2.95/3.40/\mathbf{4.69}$] & $67.7/73.4$[$29.6/7.00$] \\
LOO-1a. Ridge Only ($\eta\!=\!0$) & $-/\checkmark/\checkmark/\checkmark$ (+$M_{3..5}$) & $2.7709 \pm 0.4574$ & $3.0398$ & $2.9285$[$+0.16$] & $2.77/2.77$ & $\mathbf{10.70}$ & $15.0/57.6$[$2.69/3.16/\mathbf{4.69}$] & $74.7/75.9$[$26.8/6.52$] \\
LOO-1b. $\Delta\hat{\mathbf{u}}$ Only & $\checkmark/\checkmark/\checkmark/\checkmark$ (+$M_{3..5}$) & $\mathbf{2.1033 \pm 0.4953}$ & $\mathbf{2.3442}$ & $\mathbf{2.2029}$[$\mathbf{+0.10}$] & $\mathbf{2.04}/\mathbf{2.17}$ & $10.71$ & $\mathbf{58.9}/\mathbf{77.7}$[$2.02/\mathbf{2.39}/\mathbf{4.69}$] & $79.9/83.3$[$16.4/6.10$] \\
LOO-2. w/o Ipsi ($M_{4,5}$) & $\checkmark/-^*/\checkmark/\checkmark$ (+$M_{3..5}$) & $2.5397 \pm 1.0365$ & $2.9282$ & $3.8941$[$+1.35$] & $2.30/2.78$ & $15.34$ & $55.7/75.6$[$3.02/4.76/\mathbf{4.69}$] & $75.0/79.1$[$18.4/8.13$] \\
LOO-3. w/o Centrip.\ Prun. & $\checkmark/\checkmark/-/\checkmark$ (+$M_{3..5}$) & $2.1388 \pm 0.4971$ & $2.3981$ & $2.2533$[$+0.11$] & $2.10/2.18$ & $10.81$ & $56.7/\mathbf{77.7}$[$\mathbf{2.01}/2.49/\mathbf{4.69}$] & $\mathbf{80.1}/83.7$[$16.1/6.10$] \\
LOO-4. w/o Cyl.\ Sinus & $\checkmark/\checkmark/\checkmark/\text{Raw}$ (+$M_{3..5}$) & $2.1563 \pm 0.5282$ & $2.3984$ & $2.2690$[$+0.11$] & $2.08/2.24$ & $10.96$ & $56.6/77.6$[$2.02/2.52/6.18$] & $77.6/82.8$[$17.1/6.10$] \\
LOO-5. w/o $M_3$ Orif. & $\checkmark/\checkmark/\checkmark/\checkmark$ (+$M_{4,5}$) & $2.1356 \pm 0.5054$ & $2.3795$ & $2.2480$[$+0.11$] & $2.07/2.20$ & $10.75$ & $56.7/\mathbf{77.7}$[$2.02/2.48/4.79$] & $80.0/\mathbf{83.8}$[$\mathbf{15.8}/\mathbf{6.09}$] \\
LOO-6. w/o $M_4$ $128^3$ & $\checkmark/\checkmark/\checkmark/\checkmark$ (+$M_{3,5}$) & $2.1703 \pm 0.4462$ & $2.4233$ & $2.2806$[$+0.11$] & $2.14/2.20$ & $10.96$ & $55.7/77.1$[$2.05/2.51/\mathbf{4.69}$] & $79.3/\mathbf{83.8}$[$17.1/6.19$] \\
LOO-7. w/o $M_5$ Excl. & $\checkmark/\checkmark/\checkmark/\checkmark$ (+$M_{3,4}$) & $2.2234 \pm 0.4668$ & $2.5680$ & $2.7145$[$+0.49$] & $2.27/2.18$ & $11.69$ & $56.4/77.3$[$2.64/2.79/\mathbf{4.69}$] & $78.2/82.3$[$18.2/6.73$] \\
\textbf{Full Proposed ($M_{1..5}$)} & $\checkmark/\checkmark/\checkmark/\checkmark$ (\textbf{+$M_{3..5}$}) & $2.1362 \pm 0.5041$ & $2.3798$ & $2.2480$[$+0.11$] & $2.07/2.20$ & $10.79$ & $56.8/\mathbf{77.7}$[$2.02/2.48/\mathbf{4.69}$] & $80.0/\mathbf{83.8}$[$\mathbf{15.8}/\mathbf{6.09}$] \\
\midrule
\multicolumn{9}{@{}l@{}}{\textbf{Part III: Out-of-Sample Core Mechanism ($2\times 2$ Pullback \& Dual-Ipsi) \& Stage $M_3$--$M_5$ Ablation (OOS-50, $N=50$ Volumes / $100$ Trees)}} \\
\midrule
OOS-LOO-1. w/o $\Delta\hat{\mathbf{u}}$ (Neither) & $-/\checkmark/\checkmark/\checkmark$ (+$M_{3..5}$) & $3.6500 \pm 1.1865$ & $3.9356$ & $4.2633$[$+0.61$] & $4.06/3.24$ & $17.50$ & $14.5/54.1$[$3.80/4.73/5.30$] & $66.5/70.7$[$31.9/8.02$] \\
OOS-LOO-1a. Ridge Only ($\eta\!=\!0$) & $-/\checkmark/\checkmark/\checkmark$ (+$M_{3..5}$) & $3.2802 \pm 1.0650$ & $3.7680$ & $3.8560$[$+0.58$] & $3.43/3.13$ & $\mathbf{15.55}$ & $17.6/58.6$[$3.27/4.44/5.30$] & $72.7/74.0$[$27.5/7.51$] \\
OOS-LOO-1b. $\Delta\hat{\mathbf{u}}$ Only & $\checkmark/\checkmark/\checkmark/\checkmark$ (+$M_{3..5}$) & $\mathbf{2.7352 \pm 1.1523}$ & $\mathbf{3.2004}$ & $\mathbf{3.2513}$[$\mathbf{+0.52}$] & $\mathbf{2.91}/\mathbf{2.56}$ & $15.82$ & $\mathbf{56.8}/\mathbf{74.4}$[$\mathbf{2.83}/\mathbf{3.68}/\mathbf{5.18}$] & $\mathbf{76.4}/\mathbf{79.9}$[$\mathbf{20.6}/7.13$] \\
OOS-LOO-2. w/o Ipsi ($M_{4,5}$) & $\checkmark/-^*/\checkmark/\checkmark$ (+$M_{3..5}$) & $3.0098 \pm 1.2822$ & $3.5193$ & $4.4728$[$+1.46$] & $3.27/2.75$ & $18.58$ & $54.0/73.0$[$3.71/5.24/\mathbf{5.18}$] & $73.4/76.2$[$23.7/8.23$] \\
OOS w/o $M_{3,4}$ $128^3$ & $\checkmark/\checkmark/\checkmark/\checkmark$ (+$M_5$) & $2.8792 \pm 1.4522$ & $3.3979$ & $3.9674$[$+1.09$] & $3.03/2.73$ & $16.24$ & $54.0/73.8$[$3.62/4.32/5.73$] & $75.4/78.2$[$23.2/7.85$] \\
OOS w/o $M_3$ Orif. & $\checkmark/\checkmark/\checkmark/\checkmark$ (+$M_{4,5}$) & $2.8152 \pm 1.1810$ & $3.3539$ & $3.7797$[$+0.96$] & $3.06/2.57$ & $15.94$ & $54.6/73.9$[$3.60/3.96/5.73$] & $75.0/78.2$[$22.9/7.69$] \\
OOS w/o $M_4$ $128^3$ & $\checkmark/\checkmark/\checkmark/\checkmark$ (+$M_{3,5}$) & $2.8781 \pm 1.4526$ & $3.3971$ & $3.9665$[$+1.09$] & $3.03/2.73$ & $16.23$ & $54.0/73.8$[$3.62/4.31/\mathbf{5.18}$] & $75.4/78.1$[$23.3/7.85$] \\
OOS w/o $M_5$ Excl. & $\checkmark/\checkmark/\checkmark/\checkmark$ (+$M_{3,4}$) & $2.8143 \pm 1.1817$ & $3.3534$ & $3.7783$[$+0.96$] & $3.06/2.57$ & $15.94$ & $54.6/73.9$[$3.60/3.96/\mathbf{5.18}$] & $75.1/78.3$[$22.8/7.68$] \\
\textbf{OOS Full Proposed ($M_{1..5}$)} & $\checkmark/\checkmark/\checkmark/\checkmark$ (\textbf{+$M_{3..5}$}) & $2.7671 \pm 1.1385$ & $3.2338$ & $3.2926$[$+0.53$] & $2.94/2.59$ & $15.85$ & $54.7/74.2$[$2.83/3.76/\mathbf{5.18}$] & $76.3/79.7$[$20.8/\mathbf{7.11}$] \\
\bottomrule
\end{tabular}}
\end{table*}

\end{document}